\documentclass{new_tlp}

\usepackage[utf8]{inputenc}
\usepackage{amsmath}
\usepackage{amssymb}
\usepackage{enumitem}
\usepackage{microtype}
\usepackage{url}

\usepackage{bm}
\usepackage{textcomp}
\newcommand{\eqdef}{\ensuremath{\mathbin{\raisebox{-1pt}[-3pt][0pt]{$\stackrel{\mathit{def}}{=}$}}}}

\newcommand{\HT}{\ensuremath{\mathrm{HT}}}
\newcommand{\EL}{\ensuremath{\mathrm{EL}}}
\newcommand{\LTL}{\ensuremath{\mathrm{LTL}}}

\newcommand{\THT}{\ensuremath{\mathrm{THT}}}
\newcommand{\THTf}{\ensuremath{\mathrm{THT}_{\!f}}}

\newcommand{\TEL}{\ensuremath{\mathrm{TEL}}}
\newcommand{\TELf}{\ensuremath{{\TEL}_{\!f}}}

\newcommand{\DL}{\ensuremath{\mathrm{DL}}}

\newcommand{\DEL}{\ensuremath{\mathrm{DEL}}}

\newcommand{\MTL}{\ensuremath{\mathrm{MTL}}}
\newcommand{\MHT}{\ensuremath{\mathrm{MHT}}}
\newcommand{\MHTf}{\ensuremath{\mathrm{MHT}_{\!f}}}
\newcommand{\MHTo}{\ensuremath{\mathrm{MHT}_{\!\omega}}}
\newcommand{\MEL}{\ensuremath{\mathrm{MEL}}}
\newcommand{\MELf}{\ensuremath{{\MEL}_{\!f}}}
\newcommand{\MELo}{\ensuremath{{\MEL}_{\omega}}}

\newcommand{\last}{\ensuremath{\boldsymbol\ell}}
\newcommand{\numeral}[1]{\overline{#1}}

\renewcommand{\H}{\ensuremath{\mathbf{H}}} \newcommand{\T}{\ensuremath{\mathbf{T}}} \newcommand{\M}{\ensuremath{\mathbf{M}}}

\newcommand{\tuple}[1]{\langle #1 \rangle}

\newcommand{\next}{\raisebox{-.5pt}{\Large\textopenbullet}}  \newcommand{\previous}{\raisebox{-.5pt}{\Large\textbullet}} \newcommand{\wnext}{{\ensuremath{\widehat{\next}}}}
\newcommand{\wprevious}{{\ensuremath{\widehat{\previous}}}}
\newcommand{\alwaysF}{\ensuremath{\square}}
\newcommand{\alwaysP}{\ensuremath{\blacksquare}}
\newcommand{\eventuallyF}{\ensuremath{\Diamond}}
\newcommand{\eventuallyP}{\ensuremath{\blacklozenge}}
\newcommand{\until}{\ensuremath{\mathbin{\bm{\mathsf{U}}}}}
\newcommand{\release}{\ensuremath{\mathbin{\bm{\mathsf{R}}}}}
\newcommand{\since}{\ensuremath{\mathbin{\bm{\mathsf{S}}}}}
\newcommand{\trigger}{\ensuremath{\mathbin{\bm{\mathsf{T}}}}}
\newcommand{\finally}{\ensuremath{\bm{\mathsf{F}}}}
\newcommand{\initially}{\ensuremath{\bm{\mathsf{I}}}}

\newcommand{\metric}[2]{\ensuremath{{#1}_{#2}}}

\newcommand{\cl}{\ensuremath{\mathrm{cl}}}
\newcommand{\Lab}[1]{\ensuremath{{\mathbf{L}_{#1}}}}
\newcommand{\Tra}[1]{\ensuremath{{\textit{df\/}(#1)}}}

\newcommand{\intervc}[2]{[#1..#2]}
\newcommand{\intervo}[2]{[#1..#2)}
\newcommand{\ointerv}[2]{(#1..#2]}
\newcommand{\rangec}[3]{#1 \in \intervc{#2}{#3}}
\newcommand{\rangeo}[3]{#1 \in \intervo{#2}{#3}}
\newcommand{\orange}[3]{#1 \in \ointerv{#2}{#3}}

\newcommand{\trivaluation}[3]{\ensuremath{\bm{#3}(#1,#2)}}
\newcommand{\trival}[2]{\trivaluation{#1}{#2}{m}} \newcommand{\trivalp}[2]{{\bm{m'}}(#1,#2)}  \newcommand{\qed}{QED}

 \usepackage{xcolor}
\makeatletter

\def\starttrans{\xdef\endtrans{\catcode`\noexpand\@=\the\catcode`\@
  \catcode`\noexpand\:=\the\catcode`\:
 }\catcode`\@=11 \catcode`\:=11 }

\starttrans

\newbox\transbox

\newcount\transfactor
\newdimen\trans:dim
\newdimen\trans:dim:a
\newdimen\trans:dim:b
\newdimen\trans:dim:c
\newdimen\trans:dim:d

\newcount\trans:count

\def\trans:def{}
\def\trans:def:a{}
\def\trans:def:b{}
\def\trans:def:c{}
\def\trans:def:d{}

\def\transboxini{\afterassignment\transboxcheck
 \setbox\transbox}

\def\transboxcheck{\ifvoid\transbox
  \expandafter\aftergroup
 \fi\transboxtodo}

\def\transhboxini{\afterassignment\transhboxcheck
 \setbox\transbox}

\def\transhboxcheck{\ifvoid\transbox
  \expandafter\aftergroup\expandafter\transhboxwrap
 \else
  \expandafter\transhboxwrap
 \fi}

\def\transhboxwrap{\ifvbox\transbox
  \setbox\transbox\hbox{\box\transbox}\fi
 \transboxtodo}

\long\def\transboxdef#1\transboxend{\bgroup\def\transboxtodo{#1\egroup}\transboxini}

\long\def\transhboxdef#1\transboxend{\bgroup\def\transboxtodo{#1\egroup}\transhboxini}

\newif\iftransbbox
\newif\iftranscbox
\transbboxtrue
\transcboxtrue

\long\def\transbcboxdef#1\transbboxdef#2\transcboxdef#3\transboxend{\bgroup\edef\transboxtodo{\unexpanded{#1}\iftransbbox\unexpanded{#2}\fi
  \iftranscbox\unexpanded{#3}\else\box\transbox\fi
 \egroup}\transhboxini}

\def\bboxtranson{\hbox\bgroup\transbboxtrue
  \def\transboxtodo{\box\transbox\egroup}\transboxini}

\def\bboxtransoff{\hbox\bgroup\transbboxfalse
  \def\transboxtodo{\box\transbox\egroup}\transboxini}

\def\cboxtranson{\hbox\bgroup\transcboxtrue
  \def\transboxtodo{\box\transbox\egroup}\transboxini}

\def\cboxtransoff{\hbox\bgroup\transcboxfalse
  \def\transboxtodo{\box\transbox\egroup}\transboxini}

\def\boxresizeto#1#2#3{\hbox\transboxdef
  \ifx\relax#1\relax\else \wd\transbox=\dimexpr#1\relax \fi
  \ifx\relax#2\relax\else \ht\transbox=\dimexpr#2\relax \fi
  \ifx\relax#3\relax\else \dp\transbox=\dimexpr#3\relax \fi
  \box\transbox
 \transboxend}

\def\boxresize#1#2#3{\hbox\transboxdef
  \ifx\relax#1\relax\else \wd\transbox=\dimexpr\wd\transbox+#1\relax \fi
  \ifx\relax#2\relax\else \ht\transbox=\dimexpr\ht\transbox+#2\relax \fi
  \ifx\relax#3\relax\else \dp\transbox=\dimexpr\dp\transbox+#3\relax \fi
  \box\transbox
 \transboxend}

\def\boxextents#1#2#3#4{\hbox\transboxdef
  \kern\dimexpr#1\relax
   \vbox{\kern\dimexpr#3\relax
    \box\transbox
    \kern\dimexpr#4\relax
   }\kern\dimexpr#2\relax
 \transboxend}

\def\boxhextent#1#2{\boxextents{#1}{#2}\z@\z@}
\def\boxvextent#1#2{\boxextents\z@\z@{#1}{#2}}

\def\boxexts#1#2#3#4{\hbox\transboxdef
  \trans:dim:a=\wd\transbox
  \trans:dim:b=\dimexpr\trans:dim:a+#1+#2\relax
  \trans:dim:c=\dimexpr\ht\transbox+\dp\transbox\relax
  \trans:dim:d=\dimexpr\trans:dim:c+#3+#4\relax
  \edef\trans:def:a{\fdivide\trans:dim:b\trans:dim:a}\edef\trans:def:b{\fdivide\trans:dim:d\trans:dim:c}\savebp\trans:def:c-\dimexpr#1\relax
  \savebp\trans:def:d\dimexpr-#4+(#3+#4)*\dp\transbox/\trans:dim:c\relax
  \pdfliteral{q \trans:def:a\space 0 0 \trans:def:b\space
                \trans:def:c\space \trans:def:d\space cm}\savebp\trans:def\wd\transbox
  \box\transbox
  \pdfliteral{Q 1 0 0 1 \trans:def\space 0 cm}\transboxend}

\def\boxhext#1#2{\boxexts{#1}{#2}\z@\z@}
\def\boxvext#1#2{\boxexts\z@\z@{#1}{#2}}

\def\boxrevolveleft{\hbox\transbcboxdef
  \trans:dim:a=\wd\transbox
  \trans:dim:b=\ht\transbox
 \transbboxdef
  \wd\transbox=\dimexpr\ht\transbox+\dp\transbox\relax
  \ht\transbox=\trans:dim:a
  \dp\transbox=\z@
 \transcboxdef
  \pdfliteral{q 0 1 -1 0 \tobp{\trans:dim:b} 0 cm}\savebp\trans:def\wd\transbox
  \box\transbox
  \pdfliteral{Q 1 0 0 1 \trans:def\space 0 cm}\transboxend}

\def\boxrevolveright{\hbox\transbcboxdef
  \trans:dim:a=\wd\transbox
  \trans:dim:b=\dp\transbox
 \transbboxdef
  \wd\transbox=\dimexpr\ht\transbox+\dp\transbox\relax
  \ht\transbox=\trans:dim:a
  \dp\transbox=\z@
 \transcboxdef
  \pdfliteral{q 0 -1 1 0 \tobp{\trans:dim:b} \tobp{\trans:dim:a} cm}\savebp\trans:def\wd\transbox
  \box\transbox
  \pdfliteral{Q 1 0 0 1 \trans:def\space 0 cm}\transboxend}

\def\boxrotatexy#1#2#3{\hbox\transboxdef
  \floatsincos\trans:def:a\trans:def:b{#1}\savebp\trans:def:c\dimexpr#2\relax
  \savebp\trans:def:d\dimexpr#3\relax
  \pdfliteral{q \trans:def:b\space
                \negbp\trans:def:a\space
                \trans:def:a\space
                \trans:def:b\space
                \trans:def:c\space
                \trans:def:d\space cm
                1 0 0 1 \negbp\trans:def:c\space
                        \negbp\trans:def:d\space cm}\savebp\trans:def\wd\transbox
  \box\transbox
  \pdfliteral{Q 1 0 0 1 \trans:def\space 0 cm}\transboxend}

\def\boxrotatell#1{\boxrotatexy{#1}\z@{-\dp\transbox}}

\def\boxrotateul#1{\boxrotatexy{#1}\z@{\ht\transbox}}

\newif\ifbboxright

\def\box:rotate:bb#1{\trans:dim:a=\wd\transbox
 \trans:dim:b=\ht\transbox
 \trans:dim:c=\dp\transbox
 \trans:dim:d=\dimexpr\ht\transbox+\dp\transbox\relax
 \trans:count=\reducetrigangle{#1}\fractperiod\relax
 \ifcase\fracttrigfourth\trans:count\relax
  \fr@ct:sin:cos:i\trans:def:a\trans:def:b\trans:count
  \wd\transbox=\dimexpr\fr@ct:mul\trans:dim:a\trans:def:b
                      +\fr@ct:mul\trans:dim:d\trans:def:a\relax
  \ht\transbox=\dimexpr\fr@ct:mul\trans:dim:b\trans:def:b\relax
  \dp\transbox=\dimexpr\fr@ct:mul\trans:dim:a\trans:def:a
                      +\fr@ct:mul\trans:dim:c\trans:def:b\relax
  \savebp\trans:def:c=\dimexpr\fr@ct:mul\trans:dim:c\trans:def:a\relax
 \or
  \fr@ct:sin:cos:ii\trans:def:a\trans:def:b\trans:count
  \wd\transbox=\dimexpr-\fr@ct:mul\trans:dim:a\trans:def:b
                       +\fr@ct:mul\trans:dim:d\trans:def:a\relax
  \ht\transbox=\dimexpr-\fr@ct:mul\trans:dim:c\trans:def:b\relax
  \dp\transbox=\dimexpr-\fr@ct:mul\trans:dim:b\trans:def:b
                        +\fr@ct:mul\trans:dim:a\trans:def:a\relax
  \savebp\trans:def:c=\dimexpr-\fr@ct:mul\trans:dim:a\trans:def:b
                              +\fr@ct:mul\trans:dim:c\trans:def:a\relax
 \or
  \fr@ct:sin:cos:iii\trans:def:a\trans:def:b\trans:count
  \wd\transbox=\dimexpr-\fr@ct:mul\trans:dim:a\trans:def:b
                       -\fr@ct:mul\trans:dim:d\trans:def:a\relax
  \ht\transbox=\dimexpr-\fr@ct:mul\trans:dim:a\trans:def:a
                       -\fr@ct:mul\trans:dim:c\trans:def:b\relax
  \dp\transbox=\dimexpr-\fr@ct:mul\trans:dim:b\trans:def:b\relax
  \savebp\trans:def:c=\dimexpr-\fr@ct:mul\trans:dim:a\trans:def:b
                              -\fr@ct:mul\trans:dim:b\trans:def:a\relax
 \or
  \fr@ct:sin:cos:iv\trans:def:a\trans:def:b\trans:count
  \wd\transbox=\dimexpr\fr@ct:mul\trans:dim:a\trans:def:b
                      -\fr@ct:mul\trans:dim:d\trans:def:a\relax
  \ht\transbox=\dimexpr\fr@ct:mul\trans:dim:b\trans:def:b
                      -\fr@ct:mul\trans:dim:a\trans:def:a\relax
  \dp\transbox=\dimexpr\fr@ct:mul\trans:dim:c\trans:def:b\relax
  \savebp\trans:def:c=\dimexpr-\fr@ct:mul\trans:dim:b\trans:def:a\relax
 \fi
 \ifbboxright
  \trans:dim:d=\dimexpr\fr@ct:mul\trans:dim:a\trans:def:a\relax
  \ht\transbox=\dimexpr\ht\transbox+\trans:dim:d\relax
  \dp\transbox=\dimexpr\dp\transbox-\trans:dim:d\relax
  \savebp\trans:def:d\trans:dim:d
 \else
  \def\trans:def:d{0}\fi
 \edef\trans:def:a{\fr@ct:div\trans:def:a}\edef\trans:def:b{\fr@ct:div\trans:def:b}\pdfliteral{q \trans:def:b\space
               \negbp\trans:def:a\space
               \trans:def:a\space
               \trans:def:b\space
               \trans:def:c\space
               \trans:def:d\space cm}\savebp\trans:def=\wd\transbox
 \box\transbox
 \pdfliteral{Q 1 0 0 1 \trans:def\space 0 cm}}

\bboxrightfalse

\def\box:slant:bb#1#2{\trans:dim:a=\wd\transbox
 \trans:dim:b=\ht\transbox
 \trans:dim:c=\dp\transbox
 \trans:dim:d=\dimexpr\ht\transbox+\dp\transbox\relax
 \trans:count=\reducetrigangle{#1}{2*\fractfourth}\relax
 \ifcase\fracttrigfourth\trans:count\relax
  \fr@ct:sin:cos:i\trans:def:a\trans:def:b\trans:count
  \wd\transbox=\dimexpr\trans:dim:a+\trans:dim:d*\trans:def:a/\trans:def:b\relax
  \savebp\trans:def:c=\dimexpr\trans:dim:c*\trans:def:a/\trans:def:b\relax
 \or
  \fr@ct:sin:cos:ii\trans:def:a\trans:def:b\trans:count
  \wd\transbox=\dimexpr\trans:dim:a-\trans:dim:d*\trans:def:a/\trans:def:b\relax
  \savebp\trans:def:c=\dimexpr-\trans:dim:b*\trans:def:a/\trans:def:b\relax
 \fi
 \edef\trans:def{\fdivide\trans:def:a\trans:def:b}\trans:count=\reducetrigangle{#2}{2*\fractfourth}\relax
 \ifcase\fracttrigfourth\trans:count\relax
  \fr@ct:sin:cos:i\trans:def:a\trans:def:b\trans:count
  \ht\transbox=\dimexpr\trans:dim:b+\trans:dim:a*\trans:def:a/\trans:def:b\relax
 \or
  \fr@ct:sin:cos:ii\trans:def:a\trans:def:b\trans:count
  \dp\transbox=\dimexpr\trans:dim:c-\trans:dim:a*\trans:def:a/\trans:def:b\relax
 \fi
 \ifbboxright
  \trans:dim:d=\dimexpr-\trans:dim:a*\trans:def:a/\trans:def:b\relax
  \ht\transbox=\dimexpr\ht\transbox+\trans:dim:d\relax
  \dp\transbox=\dimexpr\dp\transbox-\trans:dim:d\relax
  \savebp\trans:def:d\trans:dim:d
 \else
  \def\trans:def:d{0}\fi
 \edef\trans:def:a{\fdivide\trans:def:a\trans:def:b}\pdfliteral{q 1
               \trans:def:a\space
               \trans:def\space
               1
               \trans:def:c\space
               \trans:def:d\space cm}\savebp\trans:def=\wd\transbox
 \box\transbox
 \pdfliteral{Q 1 0 0 1 \trans:def\space 0 cm}}

\def\boxxform{\hbox\transboxdef
  \immediate\pdfxform\transbox
  \pdfrefxform\pdflastxform
 \transboxend}

\def\boxxformspec#1\boxxform{\hbox\transboxdef
  \immediate\pdfxform#1\transbox
  \pdfrefxform\pdflastxform
 \transboxend}

\def\boxraise#1{\hbox\transboxdef
  \raise\dimexpr#1\relax\box\transbox
 \transboxend}

\def\boxlower#1{\hbox\transboxdef
  \lower\dimexpr#1\relax\box\transbox
 \transboxend}

\def\boxbaselineat#1{\hbox\transboxdef
  \lower\dimexpr(\ht\transbox+\dp\transbox)*(#1)/\transfactor-\dp\transbox\relax
  \box\transbox
 \transboxend}

\def\boxmoveleft#1{\vbox\transboxdef
  \moveleft\dimexpr#1\relax\box\transbox
 \transboxend}

\def\boxmoveright#1{\vbox\transboxdef
  \moveright\dimexpr#1\relax\box\transbox
 \transboxend}

\def\b@x:rule#1#2#{#1\transboxdef
  \setbox0\hbox{\vrule width\wd\transbox height\ht\transbox depth\dp\transbox #2}\wd\transbox=\wd0
  \ht\transbox=\ht0
  \dp\transbox=\dp0
  \box\transbox
 \transboxend#1}

\def\hboxr{\b@x:rule\hbox}
\def\vboxr{\b@x:rule\vbox}
\def\vtopr{\b@x:rule\vtop}

\def\boxgs#1#2{\hbox\transboxdef
  \pdfliteral{q #1}\savebp\trans:def\wd\transbox
  \box\transbox
  \pdfliteral{#2 Q 1 0 0 1 \trans:def\space 0 cm}\transboxend}

\def\boxmarkers#1#2#3{\hbox\transboxdef
  \copy\transbox
  \trans:dim:a=\dimexpr#1\relax
  \trans:dim:b=\dimexpr#2\relax
  \pdfliteral{q #3}\savebp\trans:def-\dp\transbox
  \box:markers:h
  \savebp\trans:def\ht\transbox
  \box:markers:h
  \savebp\trans:def-\wd\transbox
  \box:markers:v
  \savebp\trans:def\z@
  \box:markers:v
  \pdfliteral{S Q}\setbox\transbox\box\voidb@x
 \transboxend}

\def\box:markers:h{\savebp\trans:def:a\trans:dim:a
  \savebp\trans:def:b\trans:dim:b
  \pdfliteral{\trans:def:a\space\trans:def\space m \trans:def:b\space\trans:def\space l}\savebp\trans:def:a\dimexpr-\wd\transbox-\trans:dim:a\relax
  \savebp\trans:def:b\dimexpr-\wd\transbox-\trans:dim:b\relax
  \pdfliteral{\trans:def:a\space\trans:def\space m \trans:def:b\space\trans:def\space l}}

\def\box:markers:v{\savebp\trans:def:a\dimexpr-\dp\transbox-\trans:dim:a\relax
  \savebp\trans:def:b\dimexpr-\dp\transbox-\trans:dim:b\relax
  \pdfliteral{\trans:def\space \trans:def:a\space m \trans:def\space \trans:def:b\space l}\savebp\trans:def:a\dimexpr\ht\transbox+\trans:dim:a\relax
  \savebp\trans:def:b\dimexpr\ht\transbox+\trans:dim:b\relax
  \pdfliteral{\trans:def\space \trans:def:a\space m \trans:def\space \trans:def:b\space l}}

\def\boxphantom{\hbox\transboxdef
  \hbox to\wd\transbox
   {\vrule width\z@ height\ht\transbox depth\dp\transbox\hss}\transboxend}

\def\boxsmash{\hbox\transhboxdef
  \wd\transbox=\z@
  \ht\transbox=\z@
  \dp\transbox=\z@
  \box\transbox
 \transboxend}

\def\hboxsmash{\hbox\transhboxdef
  \wd\transbox=\z@
  \box\transbox
 \transboxend}

\def\vboxsmash{\vbox\transhboxdef
  \ht\transbox=\z@
  \dp\transbox=\z@
  \box\transbox
 \transboxend}

\def\box:about#1{\hbox\bgroup
  \def\transboxtodo{\trans:dim:a=\wd\transbox
   \trans:dim:b=\ht\transbox
   \trans:dim:c=\dp\transbox
   \box\transbox
   \hbox to\z@{\hss
    \hbox to\trans:dim:a{\hss
     \lower\trans:dim:c\vbox to\z@{\vss
      \vbox to\dimexpr\trans:dim:b+\trans:dim:c{\vss\tt
       #1\vbox{\vskip1ex
        \halign{\hskip1ex plus 1fil####&####\hskip1ex plus 1fil\cr
         \trans:def\span\cr
         wd & \the\trans:dim:a\cr
         ht & \the\trans:dim:b\cr
         dp & \the\trans:dim:c\cr}\vskip1ex
       }\vss}}\hss}}\egroup}\def\trans:def{}\def\trans:def:a{}\box:@bout}

\def\box:@bout#1{\ifcase
  \ifx#1\hbox 0 \else
  \ifx#1\vbox 1 \else
  \ifx#1\vtop 2 \else
  \ifx#1\box  3 \else
  \ifx#1\copy 4 \else 5 \fi\fi\fi\fi\fi
 \edef\trans:def{\trans:def\string\hbox}\expandafter\transboxini\or
 \edef\trans:def{\trans:def\string\vbox}\expandafter\transboxini\or
 \edef\trans:def{\trans:def\string\vtop}\expandafter\transboxini\or
 \edef\trans:def{\trans:def\string\box}\expandafter\boxabout:register\or
 \edef\trans:def{\trans:def\string\copy}\expandafter\boxabout:register\or
 \ifx#1\trans:def:a\errmessage{`#1' is not a box}\fi\let\trans:def:a#1\edef\trans:def{\trans:def\string#1->}\expandafter\expandafter\expandafter\box:@bout\fi
 #1}

\def\boxabout:register#1{\let\trans:def:a#1\afterassignment\boxabout:r@gister\trans:count}

\def\boxabout:r@gister{\edef\trans:def{\trans:def\the\trans:count\space
  (\ifvoid\trans:count void\else
   \ifhbox\trans:count hbox\else
   \ifvbox\trans:count vbox\fi\fi\fi)}\afterassignment\transboxtodo
 \setbox\transbox\trans:def:a\trans:count}

\def\boxabout#1{\box:about{\boxgs{#1}{}}}

\def\expandnumberafter#1#2{\expandafter#1\expandafter{\number#2}}

\def\expandtwonumbersafter#1#2#3{\expandafter#1\expandafter
 {\number#2\expandafter}\expandafter
 {\number#3}}

\def\expandthreenumbersafter#1#2#3#4{\expandafter#1\expandafter
 {\number#2\expandafter}\expandafter
 {\number#3\expandafter}\expandafter
 {\number#4}}

\def\expandnumexprafter#1#2{\expandafter#1\expandafter{\number\numexpr#2}}

\def\expandtwonumexprafter#1#2#3{\expandafter#1\expandafter
 {\number\numexpr#2\expandafter}\expandafter
 {\number\numexpr#3}}

\def\expandthreenumexprafter#1#2#3#4{\expandafter#1\expandafter
 {\number\numexpr#2\expandafter}\expandafter
 {\number\numexpr#3\expandafter}\expandafter
 {\number\numexpr#4}}

\def\expanddimexprafter#1#2{\expandafter#1\expandafter{\the\dimexpr#2}}

\edef\pt:f@ctor{\number\dimexpr100pt} \edef\bp:f@ctor{\number\dimexpr100bp} 

\begingroup
 \catcode`\P=12
 \catcode`\T=12
 \lccode`P=`p
 \lccode`T=`t
 \lowercase{\gdef\with@ut:pt#1PT{#1}}
\endgroup

\def\withoutpt{\expandafter\with@ut:pt}
\def\negbp#1{\withoutpt\the\dimexpr-#1pt\relax}
\def\asbp#1{\withoutpt\the\dimexpr#1*\pt:f@ctor/\bp:f@ctor\relax}

\def\roundbp#1{\expandafter\r@undbp\the\dimexpr(#1)*\pt:f@ctor/\bp:f@ctor\relax0000\relax}
\def\r@undbp{\csname r@und:bp:\the\pdfdecimaldigits\expandafter\endcsname
	\with@ut:pt}

\expandafter\def\csname r@und:bp:0\endcsname #1.#2#3\relax{\number\numexpr#1#2/10\relax}
\expandafter\def\csname r@und:bp:1\endcsname #1.#2#3#4\relax{\round:bp:once{#1}{#2#3}\relax}
\expandafter\def\csname r@und:bp:2\endcsname #1.#2#3#4#5\relax{\round:bp:once{#1}{#2#3#4}\relax}
\expandafter\def\csname r@und:bp:3\endcsname #1.#2#3#4#5#6\relax{\round:bp:once{#1}{#2#3#4#5}\relax}
\expandafter\def\csname r@und:bp:4\endcsname #1.#2#3#4#5#6#7\relax{\round:bp:once{#1}{#2#3#4#5#6}\relax}

\def\round:bp:once#1#2{\ifnum#11<0-\number\numexpr-\else\number\numexpr\fi
 #1+(\m@ne+\expandafter\r@und:bp:once\number\numexpr1#2/10\relax}

\def\r@und:bp:once#1#2\relax{#1)\relax\ifnum#2>0.#2\fi}

\def\set:bp:rounder#1#2{\expandafter\edef\csname #1:\the\numexpr#2\relax\endcsname##1{\unexpanded{\expandafter\expandafter\expandafter}\expandafter\noexpand
  \csname r@und:bp:\the\numexpr#2\relax\endcsname
  \unexpanded{\expandafter\with@ut:pt\the}\dimexpr(##1)*\unexpanded{\pt:f@ctor/\bp:f@ctor}\relax0000\relax}}

\set:bp:rounder{roundbpto}{0}
\set:bp:rounder{roundbpto}{1}
\set:bp:rounder{roundbpto}{2}
\set:bp:rounder{roundbpto}{3}
\set:bp:rounder{roundbpto}{4}

\def\roundbpto#1{\csname roundbpto:#1\endcsname}

\def\enablebpround{\let\tobp\roundbp}
\def\disablebpround{\let\tobp\asbp}
\def\setbpround#1{\expandafter\let\expandafter\tobp\csname roundbpto:\the\numexpr#1\relax\endcsname}

\enablebpround

\def\savebp#1{\def\s@vebp{\edef#1{\tobp{\bp:dim@n}}}\afterassignment\s@vebp\bp:dim@n}

\newdimen\bp:dim@n

\def\absoluteint#1{\numexpr\ifnum#1<\z@-\fi#1}

\def\absolutedim#1{\dimexpr\ifdim#1<\z@-\fi#1}

\def\expanddivisionafter#1#2#3{\expandnumexprafter#1{#2/#3}{#2}{#3}}

\def\divfloor{\expandtwonumexprafter\dividefloor}
\def\dividefloor{\expanddivisionafter\divide:fl@@r}
\def\divide:fl@@r#1#2#3{\numexpr#1\ifcase\ifnum#2<0 \ifnum#3<0 1 \else 0 \fi
        \else      \ifnum#3<0 1 \else 0 \fi \fi
 \ifnum\numexpr#1*#3>#2-\@ne\fi\or
 \ifnum\numexpr#1*#3<#2-\@ne\fi\fi}

\def\divceil{\expandtwonumexprafter\divideceil}
\def\divideceil{\expanddivisionafter\divide:c@il}
\def\divide:c@il#1#2#3{\numexpr#1\ifcase\ifnum#2<0 \ifnum#3<0 1 \else 0 \fi
        \else      \ifnum#3<0 1 \else 0 \fi \fi
 \ifnum\numexpr#1*#3<#2+\@ne\fi\or
 \ifnum\numexpr#1*#3>#2+\@ne\fi\fi}

\def\divint{\expandtwonumexprafter\divideint}
\def\divideint{\expanddivisionafter\divide:int}
\def\divide:int#1#2#3{\numexpr#1\ifcase\ifnum#2<0 \ifnum#3<0 3 \else 1 \fi
        \else      \ifnum#3<0 2 \else 0 \fi \fi
 \ifnum\numexpr#1*#3>#2-\@ne\fi\or
 \ifnum\numexpr#1*#3<#2+\@ne\fi\or
 \ifnum\numexpr#1*#3>#2+\@ne\fi\or
 \ifnum\numexpr#1*#3<#2-\@ne\fi\fi}

\def\divnint{\expandtwonumexprafter\dividenint}
\def\dividenint#1#2{\numexpr#1/#2}

\def\modulo{\expanddivisionafter\do:m@dulo}
\def\do:m@dulo#1#2#3{\numexpr#2-#3*\divide:fl@@r{#1}{#2}{#3}\relax}

\def\dividefloorpos{\expanddivisionafter\divide:fl@@r:pos}
\def\divide:fl@@r:pos#1#2#3{\numexpr#1\ifnum\numexpr#1*#3>#2-\@ne\fi}

\def\divide:c@il:pos#1#2#3{\numexpr#1\ifnum\numexpr#1*#3<#2+\@ne\fi}

\def\modpos{\expandtwonumexprafter\modulopos}
\def\modulopos{\expanddivisionafter\modulo:p@s}
\def\modulo:p@s#1#2#3{\numexpr#2-#3*\divide:fl@@r:pos{#1}{#2}{#3}\relax}

\def\floatround#1{\divnint{\dimexpr#1pt}\p@}
\def\floatfloor#1{\divfloor{\dimexpr#1pt}\p@}
\def\floatceil#1{\divceil{\dimexpr#1pt}\p@}
\def\floatint#1{\divint{\dimexpr#1pt}\p@}
\def\floatnint#1{\divnint{\dimexpr#1pt}\p@}

\def\fdivide{\expandtwonumexprafter\flo@t:divide}
\def\flo@t:divide#1#2{\withoutpt\the\dimexpr\numexpr#1*\p@/#2\relax sp\relax}

\def\divfloat{\expandthreenumexprafter\dividefloat}

\def\dividefloat#1#2#3{\expandnumberafter\divide:flo@t {\absoluteint{\divideint{#1}{#2}}}{#1}{#2}{#3}}

\def\divide:flo@t#1#2#3{\ifnum#2<0 \ifnum#3>0 -\fi\else
 \ifnum#2>0 \ifnum#3<0 -\fi\fi\fi
 #1.\expandthreenumexprafter\divide:fl@@t
 {#1}{\absoluteint{#2}}{\absoluteint{#3}}}

\def\divide:fl@@t#1#2#3{\expandnumexprafter\divide:flo@t:modulo{#2-#1*#3}{#3}}

\def\divide:flo@t:modulo#1#2{\ifnum#1<214748365
  \expandtwonumbersafter\divide:flo@t:result{#10}{#2\expandafter}\else
  \expandtwonumexprafter\divide:flo@t:modulo{#1/2}{#2/2\expandafter}\fi}

\def\divide:flo@t:result#1#2#3{\ifnum#3>1
  \expandtwonumexprafter\divide:flo@t:repeat
  {#3-\@ne}{\dividefloorpos{#1}{#2}\expandafter}\else
  \number\divide:flo@t:last{#1}{#2}\relax
  \expandafter\gobbletwo
 \fi{#1}{#2}}

\def\divide:flo@t:repeat#1#2#3#4{#2\expandnumexprafter\divide:flo@t:modulo{#3-#2*#4}{#4}{#1}}

\newcount\floatprecision
\def\roundlast{\let\divide:flo@t:last\dividenint}
\def\floorlast{\let\divide:flo@t:last\divideint}
\roundlast

\def\tofixedbp#1{\divfloat{\dimexpr#1}\b@\floatprecision}
\def\roundfixedbp#1{\divfloat{\dimexpr#1}\b@\pdfdecimaldigits}

\def\fractdegree#1{\numexpr16*\dimexpr#1pt}           \edef\fractfactor{\number\numexpr\maxdimen+\@ne}      \edef\fractfourth{\number\numexpr90*\fractdegree\@ne} \edef\fractperiod{\number\numexpr4*\fractfourth}      

\def\reducefractangle#1{\expandnumberafter\reduce:fr@ct:angle{\fractdegree{#1}}}

\def\reduce:fr@ct:angle#1#2{\ifnum#1<0
  \numexpr#2-\modpos{-#1}{#2}\relax
 \else
  \modpos{#1}{#2}\fi}

\def\reduceintangle#1#2{\expandtwonumexprafter\reduce:int:@ngle{#1}{#2/\fractdegree\@ne}}

\def\reduce:int:@ngle#1#2{\fractdegree{\reduce:fr@ct:angle{#1}{#2}}}

\def\enablefractangle{\let\reducetrigangle\reducefractangle}
\def\disablefractangle{\let\reducetrigangle\reduceintangle}
\enablefractangle

\def\fracttrigfourth#1{\dividefloorpos{#1}\fractfourth}

\def\fr@ct:mul#1#2{#1*#2/\fractfactor}
\def\fr@ct:div#1{\fdivide{#1}\fractfactor}

\def\fr@ct:angle#1{\ifcase\numexpr#1\relax
62914560\or 47185920\or 31457280\or 16777216\or 8388608\or 4194304\or 2097152\or 1048576\or 524288\or 262144\or 131072\or 65536\or 32768\or 16384\or 8192\or 4096\or 2048\or 1024\or 512\or 256\or 128\or 64\or 32\or 16\or 8\or 4\or 2\or 1\fi}

\def\fr@ct:sin#1{\ifcase\numexpr#1\relax
929887697\or 759250125\or 536870912\or 295963357\or 149435979\or 74900443\or 37473049\or 18739379\or 9370046\or 4685068\or 2342539\or 1171270\or 585635\or 292818\or 146409\or 73204\or 36602\or 18301\or 9151\or 4575\or 2288\or 1144\or 572\or 286\or 143\or 71\or 36\or 18\fi}

\def\fr@ct:cos#1{\ifcase\numexpr#1\relax
 536870912\or 759250125\or 929887697\or 1032146887\or 1063292242\or 1071126243\or 1073087729\or 1073578288\or 1073700939\or 1073731603\or 1073739269\or 1073741185\or 1073741664\or 1073741784\or 1073741814\or 1073741822\or 1073741823\or 1073741824\or 1073741824\or 1073741824\or 1073741824\or 1073741824\or 1073741824\or 1073741824\or 1073741824\or 1073741824\or 1073741824\or 1073741824\fi}

\trans:count=0
\loop
 \expandafter\edef\csname
   fractangle:\the\trans:count\endcsname{\fr@ct:angle\trans:count}
 \expandafter\edef\csname
   fractsinvalue:\the\trans:count\endcsname{\fr@ct:sin\trans:count}
 \expandafter\edef\csname
   fractcosvalue:\the\trans:count\endcsname{\fr@ct:cos\trans:count}
 \ifnum\trans:count<27
 \advance\trans:count by1
\repeat
\def\fr@ct:angle#1{\csname fractangle:\number#1\endcsname}
\def\fr@ct:sin#1{\csname fractsinvalue:\number#1\endcsname}
\def\fr@ct:cos#1{\csname fractcosvalue:\number#1\endcsname}

\def\fracttrig#1{\expandnumexprafter\fr@ct:trig{\reducetrigangle{#1}\fractperiod}}

\def\fr@ct:trig#1{\csname fr@ct:trig:\romannumeral\fracttrigfourth{#1}+\@ne\endcsname
 {#1}}

\def\fr@ct:trig:i#1#2#3{\expandthreenumexprafter\fr@ct:trig:cont{#1}{#2}{#3}\z@}

\def\fr@ct:trig:ii#1#2#3{\expandthreenumexprafter\fr@ct:trig:cont{#1-\fractfourth}{#3}{-#2}\z@}

\def\fr@ct:trig:iii#1#2#3{\expandthreenumexprafter\fr@ct:trig:cont{#1-2*\fractfourth}{-#2}{-#3}\z@}

\def\fr@ct:trig:iv#1#2#3{\expandthreenumexprafter\fr@ct:trig:cont{#1-3*\fractfourth}{-#3}{#2}\z@}

\def\fr@ct:trig:cont#1#2#3#4{\ifcase
  \ifnum#1>0 \ifnum#1<\fr@ct:angle{#4} 0 \else 1 \fi \else 2 \fi
  \expandafter\fr@ct:trig:cont\expandafter
  {\number#1\expandafter}\expandafter
  {\number#2\expandafter}\expandafter
  {\number#3\expandafter}\expandafter
  {\number\numexpr#4+\@ne\expandafter}\or
  \expandafter\fr@ct:trig:cont\expandafter
  {\number\numexpr#1-\fr@ct:angle{#4}\expandafter}\expandafter
  {\number\numexpr\fr@ct:mul{#2}{\fr@ct:cos{#4}}+\fr@ct:mul{#3}{\fr@ct:sin{#4}}\expandafter}\expandafter
  {\number\numexpr\fr@ct:mul{#3}{\fr@ct:cos{#4}}-\fr@ct:mul{#2}{\fr@ct:sin{#4}}\expandafter}\expandafter
  {\number\numexpr#4+\@ne\expandafter}\or
  \fracttrigend{#2}{#3}\fi}

\def\fracttrigend#1#2\fi#3{\fi#3{#1}{#2}}

\def\fractsincos#1#2#3{\fracttrig{#3}\z@\fractfactor\fr@ct:sin:cos#1#2}
\def\fr@ct:sin:cos#1#2#3#4{\def#3{#1}\def#4{#2}}

\def\fr@ct:sin:cos:i#1#2#3{\fr@ct:trig:i{#3}\z@\fractfactor\fr@ct:sin:cos#1#2}
\def\fr@ct:sin:cos:ii#1#2#3{\fr@ct:trig:ii{#3}\z@\fractfactor\fr@ct:sin:cos#1#2}
\def\fr@ct:sin:cos:iii#1#2#3{\fr@ct:trig:iii{#3}\z@\fractfactor\fr@ct:sin:cos#1#2}
\def\fr@ct:sin:cos:iv#1#2#3{\fr@ct:trig:iv{#3}\z@\fractfactor\fr@ct:sin:cos#1#2}

\def\floatsincos#1#2#3{\fracttrig{#3}\z@\fractfactor\flo@t:sin:c@s#1#2}
\def\flo@t:sin:c@s#1#2#3#4{\edef#3{\fr@ct:div{#1}}\edef#4{\fr@ct:div{#2}}}

\endtrans
 \newbox\qbox
\def\usecolor#1{\csname\string\color@#1\endcsname\space}
\newcommand\bordercolor[1]{\colsplit{1}{#1}}
\newcommand\fillcolor[1]{\colsplit{0}{#1}}
\newcommand\outline[1]{\leavevmode \def\maltext{#1}\setbox\qbox=\hbox{\maltext}\boxgs{Q q 2 Tr \thickness\space w \fillcol\space \bordercol\space}{}\copy\qbox }
\makeatother
\newcommand\colsplit[2]{\colorlet{tmpcolor}{#2}\edef\tmp{\usecolor{tmpcolor}}\def\tmpB{}\expandafter\colsplithelp\tmp\relax \ifnum0=#1\relax\edef\fillcol{\tmpB}\else\edef\bordercol{\tmpC}\fi}
\def\colsplithelp#1#2 #3\relax{\edef\tmpB{\tmpB#1#2 }\ifnum `#1>`9\relax\def\tmpC{#3}\else\colsplithelp#3\relax\fi
}
\bordercolor{black}
\fillcolor{white}
\def\thickness{.3}

\renewcommand{\until}{\ensuremath{\mathbin{\mbox{\outline{$\bm{\mathsf{U}}$}}}}}
\renewcommand{\release}{\ensuremath{\mathbin{\mbox{\outline{$\bm{\mathsf{R}}$}}}}}
\renewcommand{\finally}{\ensuremath{\mbox{\outline{$\bm{\mathsf{F}}$}}}}

\newtheorem{definition}{Definition}
\newtheorem{theorem}{Theorem}
\newtheorem{proposition}{Proposition}
\newtheorem{corollary}{Corollary}
\newtheorem{lemma}{Lemma}
\newenvironment{proofof}[1]{\noindent {\bf Proof of #1.}}{\qed}

\newcommand{\sysfont}{\textit}

\newcommand{\clingo}{\sysfont{clingo}}

\newcommand{\telingo}{\sysfont{telingo}}

\newcommand{\mysubsection}[1]{\paragraph{#1.}}

\allowdisplaybreaks{}

\begin{document}

\title{Towards Metric Temporal Answer Set Programming}

\author[Cabalar et al.]{PEDRO CABALAR\thanks{Partially supported by MINECO, Spain, grant TIC2017-84453-P.}\\ {University of Corunna, Spain}
  \and
  MART\'{I}N DI\'EGUEZ\\ {LERIA, Universit\'e d'Angers, France}
  \and
  TORSTEN SCHAUB
  and
  ANNA SCHUHMANN\\ {University of Potsdam, Germany}
}

\maketitle

\begin{abstract}
  We elaborate upon the theoretical foundations of a metric temporal extension of Answer Set Programming.
  In analogy to previous extensions of ASP with constructs from Linear Temporal and Dynamic Logic,
  we accomplish this in the setting of the logic of Here-and-There and its non-monotonic extension, called Equilibrium Logic.
  More precisely, we develop our logic on the same semantic underpinnings as its predecessors and thus use a simple time domain of bounded time steps.
  This allows us to compare all variants in a uniform framework and ultimately combine them in a common implementation.

  \bigskip\noindent
  {\em This article is under consideration for acceptance in TPLP.}
\end{abstract}
 \section{Introduction}\label{sec:introduction}

Reasoning about action and change, or more generally reasoning about dynamic systems, is not only central to knowledge representation and reasoning
but at the heart of computer science.
We addressed this over the last years by combining traditional approaches, like Dynamic and Linear Temporal Logic (\DL~\cite{hatiko00a} and \LTL~\cite{pnueli77a}),
with the base logic of Answer Set Programming (ASP~\cite{lifschitz99b}), namely,
the logic of Here-and-There (\HT~\cite{heyting30a}) and its non-monotonic extension, called Equilibrium Logic (\EL~\cite{pearce96a}.
This resulted in non-monotonic linear dynamic and temporal equilibrium logics
(\DEL~\cite{bocadisc18a,cadisc19a} and \TEL~\cite{agcadipevi13a,cakascsc18a})
that gave rise to the temporal ASP system \telingo~\cite{cakamosc19a,cadilasc20a} extending the full-featured ASP system \clingo~\cite{gekakaosscwa16a}.
A key design decision has been to base both logics on the same semantic structures so that language constructs from both can be jointly used in an
implementation.
Another commonality of dynamic and temporal logics is that they abstract from specific time points and rather focus on capturing temporal relationships.
For instance, we can express in a temporal logic
that a machine has to be eventually cleaned after being used with the formula
\(
\alwaysF ( \mathit{use} \to \eventuallyF \mathit{clean})
\).
However, sometimes this is not enough to capture the desired relation.
That is, we might also want to quantify the time difference between both events.
For instance, whenever the machine is used, it has to be cleaned within less than 5 time units.
This can be expressed by means of metric temporal operators as follows:
\begin{align}\label{ex:use:clean}
  \alwaysF (\mathit{use} \to \metric{\eventuallyF}{5}\mathit{clean}) \ .
\end{align}

In this paper, we address this issue and elaborate upon a combination of Metric Temporal Logic (\MTL~\cite{aluhen92a,ouawor05a}
\footnote{Unlike traditional approaches usually having continuous time domains~\cite{aluhen92a},
  we deal with point-based semantics based on discrete linear time, similar to~\cite{ouawor05a}.})
with \HT\ and \EL.
Our development of \emph{Metric Equilibrium Logic} (\MEL) not only parallels the one of \TEL\ and \DEL\ mentioned above
but, moreover, builds on the same semantic foundations.
This allows us to relate all three systems in a uniform semantic setting and, ultimately, to integrate the corresponding language constructs in a common
implementation.

A full version of this paper including proofs of results can be found at \url{http://arxiv.org/abs/2008.02038}.

 \section{Metric Equilibrium Logic}\label{sec:mel}

Given a set $\mathcal{A}$ of atoms, or \emph{alphabet},
we define a \emph{(metric) formula} $\varphi$ by the grammar:
\begin{align*}
  \varphi & ::= a \mid \bot \mid \varphi_1 \bowtie \varphi_2 \mid
            \previous\varphi \mid  \varphi_1 \metric{\since}{n} \varphi_2 \mid \varphi_1 \metric{\trigger}{n} \varphi_2 \mid
            \next \varphi \mid \varphi_1 \metric{\until}{n} \varphi_2 \mid \varphi_1 \metric{\release}{n} \varphi_2
\end{align*}
where
$a\in\mathcal{A}$ is an atom
and
${\bowtie} \in \{\to,\wedge,\vee\}$ is a binary Boolean connective;
$n$ is a numeral constant (referring to some integer number) or the symbolic constant \last\
(standing for the length of a trace; see below).
The last six cases of $\varphi$ correspond to the metric past connectives
\emph{previous},
\emph{since},
\emph{trigger},
and their future counterparts
\emph{next},
\emph{until}, and
\emph{release},
where $n>0$ restricts the scope of each operator to the last (resp.~next) $n$ time points, including the current state.\footnote{Values $n\leq 0$ are tolerated but trivialize the subformula at hand, as made precise in Proposition~\ref{prop:validities}.}

We also define several derived operators like the Boolean connectives
\(
\top \eqdef \neg \bot
\),
\(
\neg \varphi \eqdef  \varphi \to \bot
\),
\(
\varphi \leftrightarrow \psi \eqdef (\varphi \to \psi) \wedge (\psi \to \varphi)
\),
and the following derived metric operators:
\[
\begin{array}{rcl}
   \initially\,                     & \eqdef & \neg \previous \top                     \\
   \wprevious \varphi               & \eqdef & \previous \varphi \vee \initially       \\
   \metric{\alwaysP}{n} \varphi     & \eqdef & \bot \metric{\trigger}{n} \varphi       \\
   \metric{\eventuallyP}{n} \varphi & \eqdef & \top \metric{\since}{n} \varphi         \\
\end{array}
\qquad
\begin{array}{rcl}
   \finally                         & \eqdef & \neg \next \top                         \\
   \wnext \varphi                   & \eqdef & \next \varphi \vee \finally             \\
   \metric{\alwaysF}{n} \varphi     & \eqdef & \bot \metric{\release}{n} \varphi       \\
   \metric{\eventuallyF}{n} \varphi & \eqdef & \top \metric{\until}{n} \varphi         \\
\end{array}
\]
On the left, we give past operators, namely,
\emph{initial},
\emph{weak previous},
\emph{always before},
\emph{eventually before},
while the right column lists their future counterparts
\emph{final},
\emph{weak next},
\emph{always afterward},
\emph{eventually afterward}.
We define the iterated application of the one step operators as
\(
\otimes^0\varphi \eqdef \varphi
\)
and
\(
\otimes^n\varphi \eqdef \otimes\otimes^{n-1}\varphi
\)
for $n>0$ and $\otimes\in\{\previous,\next,\wprevious,\wnext\}$.
For instance, $\next^2 p$ corresponds to $\next \next p$.
Here, the use of $n$ with temporal operators captures a number of iterations of some one step expression so that,
as we see below, for instance, $\metric{\eventuallyF}{3} \varphi$ amounts to $\varphi \vee \next \varphi \vee \next^2 \varphi$.

An example of metric formulas is the modeling of traffic lights.
While the light is red by default,
it changes to green within less than 3 time units whenever the button is pushed;
and it stays green for other 3 time units.
This can be represented by
\begin{align}
\alwaysF ( \mathit{red} \wedge \mathit{green} \to \bot) \label{ex:traffic:light}\\
\alwaysF ( \neg\mathit{green} \to \mathit{red} ) \label{ex:traffic:light:default}\\
\alwaysF ( \mathit{push} \to \metric{\eventuallyF}{3}\metric{\alwaysF}{4} \mathit{green} ) \label{ex:traffic:light:push}
\end{align}
Note that this example combines a default rule \eqref{ex:traffic:light:default} with a metric rule \eqref{ex:traffic:light:push},
describing the initiation and duration period of events.
This nicely illustrates the interest in non-monotonic metric representation and reasoning methods.

Given $a \in \mathbb{N}$ and $b \in \mathbb{N} \cup \{\omega\}$,
we let $\intervc{a}{b}$ stand for the set $\{i \in \mathbb{N} \mid a \leq i \leq b\}$,
$\intervo{a}{b}$ for $\{i \in \mathbb{N} \mid a \leq i < b\}$ and $\ointerv{a}{b}$ for $\{i \in \mathbb{N} \mid a < i \leq b\}$.
For the semantics, we start by defining a \emph{trace} of length $\lambda$ over alphabet $\mathcal{A}$ as a sequence
$(H_i)_{\rangeo{i}{0}{\lambda}}$ of sets $H_i\subseteq\mathcal{A}$.
A trace is \emph{infinite} if $\lambda=\omega$ and \emph{finite} if $\lambda=n$ for some natural number $n \in \mathbb{N}$.
Given traces $\H=(H_i)_{\rangeo{i}{0}{\lambda}}$ and $\H'=(H'_i)_{\rangeo{i}{0}{\lambda}}$ both of length $\lambda$,
we write $\H\leq\mathbf\H'$ if $H_i\subseteq H'_i$ for each $\rangeo{i}{0}{\lambda}$;
accordingly, $\mathbf{H}<\mathbf{H'}$ iff both $\mathbf{H}\leq\mathbf{H'}$ and $\mathbf{H}\neq\mathbf{H'}$.

Our semantics is based on \emph{Here-and-There traces} (for short \emph{\HT-traces}~\cite{cakascsc18a}) of length $\lambda$ over alphabet $\mathcal{A}$
being sequences of pairs
\(
(\tuple{H_i,T_i})_{\rangeo{i}{0}{\lambda}}
\)
such that $H_i\subseteq T_i\subseteq \mathcal{A}$ for any ${\rangeo{i}{0}{\lambda}}$.
We often represent an \HT-trace as a pair of traces $\tuple{\H,\T}$ of length $\lambda$
where $\H=(H_i)_{\rangeo{i}{0}{\lambda}}$ and $\T=(T_i)_{\rangeo{i}{0}{\lambda}}$ such that $\H \leq \T$.
When an \HT-trace $\tuple{\H,\T}$ satisfies $\H=\T$, it is called \emph{total}.

We assume a one-to-one correspondence between numeral constants and integers
and let $\numeral{n}$ stand for the number corresponding to numeral~$n$.
For the symbolic constant $\last$, we fix $\numeral{\last}=\lambda$ to the length~$\lambda$ of the trace.
For simplicity,
we let expressions like $n-m$, formed with numeral constants $n$ and $m$,
stand for the numeral representing the difference between $\numeral{n}$ and $\numeral{m}$.

We define the semantics of metric formulas in terms of \HT-traces.
\begin{definition}[Satisfaction]\label{def:satisfaction}
Let $\M=\tuple{\H,\T}$ be an $\HT$-trace of length $\lambda$ over alphabet $\mathcal{A}$,
and let $\varphi$ be a metric formula over $\mathcal{A}$.
The trace $\M$ satisfies $\varphi$ at time point $\rangeo{k}{0}{\lambda}$,
written $\M, k \models \varphi$,
if
\begin{enumerate} \setcounter{enumi}{0}
\item $\M, k \not\models \bot$
\item $\M, k \models a$
  iff $a \in H_k$, for any atom $a \in \mathcal{A}$
\item $\M, k \models \varphi \wedge \psi$
  iff
  $\M, k \models \varphi$
  and
  $\M, k \models \psi$
\item $\M, k \models \varphi \vee \psi$
  iff
  $\M, k \models \varphi$
  or
  $\M, k \models \psi$
\item $\M, k \models \varphi \to \psi$
  iff
  $\langle \mathbf{H}', \mathbf{T} \rangle, k \not \models \varphi$
  or
  $\langle \mathbf{H}', \mathbf{T} \rangle, k \models  \psi$, for all $\mathbf{H'} \in \{ \mathbf{H}, \mathbf{T} \}$
\item $\M, k \models \previous \varphi$
  iff
  $k>0$ and $\M, k{-}1 \models \varphi$
\item $\M, k \models \next \varphi$
  iff
  $k+1<\lambda$ and $\M, k{+}1 \models \varphi$
\item\label{def:satisfaction:until}
  $\M, k \models \varphi \metric{\until}{n} \psi$
  iff
    for some $\rangeo{j}{0}{\numeral{n}}$ such that
    $k{+}j \in \intervo{0}{\lambda}$ we have $\M, k{+}j \models \psi$
    and
    $\M, k{+}i \models \varphi$ for all $\rangeo{i}{0}{j}$
\item\label{def:satisfaction:release}
	  $\M, k \models \varphi \metric{\release}{n} \psi$
  iff
    for all $\rangeo{j}{0}{\numeral{n}}$ such that
    $k{+}j \in \intervo{0}{\lambda}$ we have $\M, k{+}j \models \psi$
    or
    $\M, k{+}i \models \varphi$ for some $\rangeo{i}{0}{j}$
\item\label{def:satisfaction:since}
     $\M, k \models \varphi \metric{\since}{n} \psi$
  iff
    for some $\rangeo{j}{0}{\numeral{n}}$ such that
    $k{-}j \in \intervo{0}{\lambda}$ we have $\M, k{-}j \models \psi$
    and
    $\M, k{-}i \models \varphi$ for all $\rangeo{i}{0}{j}$
\item\label{def:satisfaction:trigger}
  	 $\M, k \models \varphi \metric{\trigger}{n} \psi$
  iff
    for all $\rangeo{j}{0}{\numeral{n}}$ such that
    $k{-}j \in \intervo{0}{\lambda}$ we have $\M, k{-}j \models \psi$
    or
    $\M, k{-}i \models \varphi$ for some $\rangeo{i}{0}{j}$
\end{enumerate}
\end{definition}
The fundamental difference to standard temporal logics is clearly the satisfaction of implication `$\to$'
that is inherited from the (non-temporal) logic \HT,
an intermediate logic dealing with exactly two worlds $h,t$ with the reflexive accessibility relation $h \leq t$.
When traces are total $\tuple{\T,\T}$, we get $h=t$ and this distinction disappears, so implication becomes classical.
From the perspective of metric temporal logic,
the definition of release and trigger in~\emph{\ref{def:satisfaction:release}} and~\emph{\ref{def:satisfaction:trigger}}, respectively,
are additionally conditioned by the trace's limits and can thus be seen as weak variants of the standard counterparts.
Similarly, the satisfaction of until and since formulas in~\emph{\ref{def:satisfaction:until}} and~\emph{\ref{def:satisfaction:since}}, respectively,
is restricted to the time points within a trace.
Clearly, for infinite traces, the restriction of future operators vanishes.

An \HT-trace $\M$ is a \emph{model} of a metric theory $\Gamma$ if $\M,0 \models \varphi$ for all $\varphi \in \Gamma$.
A formula $\varphi$ is a \emph{tautology} (or is \emph{valid}),
written $\models \varphi$, iff $\M,k \models \varphi$ for any \HT-trace \M\ and any $k\in\intervo{0}{\lambda}$.
We call the logic induced by the set of all tautologies \emph{Metric logic of Here and There} (\MHT\ for short).
We say that an \HT-trace $\M$ is a \emph{model} of a set of formulas (or \emph{theory}) $\Gamma$ iff $\M,0 \models \varphi$ for any $\varphi \in \Gamma$.
Two formulas $\varphi, \psi$ are \emph{equivalent} if $\models \varphi \leftrightarrow \psi$.
Whenever two formulas $\varphi$ and $\psi$ are equivalent,
they are completely interchangeable in any theory
without altering the theory's semantics.

We write $\MHT(\Gamma,\lambda)$ to stand for the set of models of length $\lambda$ of a theory $\Gamma$,
and define $\MHT(\Gamma) \eqdef \MHT(\Gamma,\omega) \cup \bigcup_{\lambda\geq 0} \MHT(\Gamma,\lambda)$, that is, the whole set of models of $\Gamma$ of any length.
An interesting subset of $\MHT(\Gamma,\lambda)$ is the one formed by total traces $\tuple{\T,\T}$, we denote as $\MTL(\Gamma,\lambda)$.
We also use $\MTL(\Gamma)$ to stand for $\MHT(\Gamma,\omega) \cup \bigcup_{\lambda\geq 0} \MTL(\Gamma,\lambda)$.
In the non-metric version of temporal \HT, the restriction to total models turns out to correspond to Linear Temporal Logic (\LTL).
In our case, it defines a metric version of \LTL\ that we call \emph{Metric Temporal Logic} (\MTL\ for short).
It can be proved that $\MTL(\Gamma,\lambda)$ are those models of $\MHT(\Gamma,\lambda)$ satisfying the \emph{excluded middle axiom} schema:
\begin{align}\label{eq:excluded:middle}
\metric{\alwaysF}{\last} (p \vee \neg p) \qquad\qquad \text{(for any atom }p \in \mathcal{A})
\end{align}

The semantics of the derived operators in \MHT\ can be easily deduced.
\begin{proposition}[Satisfaction]\label{prop:satisfaction}
  Let $\M=\tuple{\H,\T}$ be an \HT-trace of length $\lambda$ over $\mathcal{A}$.
  Given the respective definitions of derived operators, we get the following satisfaction conditions:
  \begin{enumerate} \setcounter{enumi}{11}
  \item $\M, k \models \metric{\alwaysP}{n} \varphi$
    iff
    for all $\rangeo{j}{0}{\numeral{n}}$ such that
    $k{-}j \in \intervo{0}{\lambda}$ we have $\M, k{-}j \models \psi$
  \item $\M, k \models \metric{\eventuallyP}{n} \varphi$
    iff
    for some $\rangeo{j}{0}{\numeral{n}}$ such that
    $k{-}j \in \intervo{0}{\lambda}$ we have $\M, k{-}j \models \psi$
  \item $\M, k \models \initially$
    iff
    $k =0$
  \item $\M, k \models \metric{\alwaysF}{n} \varphi$
   iff
    for all $\rangeo{j}{0}{\numeral{n}}$ such that
    $k{+}j \in \intervo{0}{\lambda}$ we have $\M, k{+}j \models \psi$
  \item $\M, k \models \metric{\eventuallyF}{n} \varphi$
    iff
    for some $\rangeo{j}{0}{\numeral{n}}$ such that
    $k{+}j \in \intervo{0}{\lambda}$ we have $\M, k{+}j \models \psi$
  \item $\M, k \models \finally$
    iff
    $k+1 = \lambda$
  \item $\M, k \models \wprevious\varphi$
    iff
    $k =0$ or
    $\M, k{-}1 \models \varphi$
  \item $\M, k \models \wnext\varphi$
    iff
    $k + 1=\lambda$ or
    $\M, k{+}1 \models \varphi$
  \end{enumerate}
\end{proposition}
As in the temporal logic of \HT~\cite{cakascsc18a}, referred to as \THT,
the operators $\initially$ and $\finally$ exclusively depend on the value of time point $k$,
and are thus independent of \M.
In fact,
operator $\finally$ allows us to influence the length of models.
The inclusion of the axiom $\metric{\eventuallyF}{n} \finally$ for example forces its models to have length $\lambda \leq \numeral{n}$
with $\numeral{n} \in \mathbb{N}$.
On the other hand,
the inclusion of the axiom $\neg \metric{\eventuallyF}{\last} \finally$ forces models to be of infinite length.
As well,
we distinguish \MHT\ on finite and infinite traces, and refer to the respective logics as \MHTf\ and \MHTo.

Following the definitions of \TEL~\cite{cakascsc18a} and \DEL~\cite{cadisc19a},
we now introduce non-monotonicity by selecting a particular set of traces that we call \emph{temporal equilibrium models}.
First, given an arbitrary set $\mathfrak{S}$ of \HT-traces, we define the ones in equilibrium as follows.
\begin{definition}[Temporal Equilibrium/Stable models]\label{def:tem}
Let $\mathfrak{S}$ be some set of \HT-traces.
A total \HT-trace $\tuple{\T,\T} \in\mathfrak{S}$ is an \emph{equilibrium trace} of $\mathfrak{S}$ iff
there is no other $\tuple{\H,\T} \in\mathfrak{S}$ such that $\H < \T$.
\end{definition}
If $\tuple{\T,\T}$ is such an equilibrium trace, we also say that trace \T\ is a \emph{stable trace} of $\mathfrak{S}$.
We further talk about \emph{temporal equilibrium} or \emph{temporal stable models} of a theory $\Gamma$
when $\mathfrak{S}=\MHT(\Gamma)$, respectively.

We write $\MEL(\Gamma,\lambda)$ and $\MEL(\Gamma)$ to stand for the temporal equilibrium models of $\MHT(\Gamma,\lambda)$ and $\MHT(\Gamma)$, respectively.
Besides, as the ordering relation among traces is only defined for a fixed $\lambda$,
it is easy to see the following result:
\begin{proposition}
The set of temporal equilibrium models of $\Gamma$ can be partitioned by the trace length $\lambda$, that is,
$\bigcup_{\lambda=0}^\omega \MEL(\Gamma,\lambda) = \MEL(\Gamma)$.
\end{proposition}

\emph{Metric Equilibrium Logic} (\MEL) is the non-monotonic logic induced by temporal equilibrium models of metric theories.
We obtain the variants \MELf\ and \MELo\ by applying the respective restriction to finite or infinite traces, respectively.

Let us illustrate this by using the example of the pedestrian traffic light introduced above.
Consider the models of the theory
$\Gamma = \{\eqref{ex:traffic:light}, \eqref{ex:traffic:light:default}, \eqref{ex:traffic:light:push}\}$
for length $\lambda = 1$.
In this case, we only have time point $k=0$, and the metric or temporal aspect is less interesting, since HT-traces amount to pairs $\tuple{H_0,T_0}$.
Still, this helps to illustrate the difference among the different sets of models defined above.
In the example, we abbreviate a set of atoms as a string formed by their initials:
for instance $gp$ stands for $\{\mathit{green},\mathit{push}\}$.
Then, we obtain the following sets:
\begin{align*}
\MTL(\Gamma, 1) = & \left\lbrace \ \tuple{r}, \ \tuple{g}, \
\tuple{gp} \ \right\rbrace \\
\MHT(\Gamma, 1) = & \ \MTL(\Gamma,1) \cup \left\lbrace \
\tuple{\emptyset, g}, \
\tuple{\emptyset, gp}, \
\tuple{g, gp} \
\right\rbrace \\
\MEL(\Gamma, 1) = & \left\lbrace\ \tuple{r} \ \right\rbrace
\end{align*}
As we can see, total models $\MTL(\Gamma, 1)$ allow for choosing either $\mathit{red}$ or $\mathit{green}$ but, if we include $\mathit{push}$, then $\mathit{green}$ is mandatory because it is the only way to satisfy $\metric{\eventuallyF}{3}\metric{\alwaysF}{4} \mathit{green}$ with $\lambda=1$.
Note how the unique equilibrium model in $\MEL(\Gamma, 1)$ is the only total model $\tuple{\T,\T}$
with $\T=\tuple{r}$ in $\MTL(\Gamma, 1)$ for which there is no $\tuple{\H,\T} \in \MHT(\Gamma, 1)$ with smaller $\H < \T$.
Informally, this is because \eqref{ex:traffic:light:default} suffices to  justify $\mathit{red}$ by default,
while the other two total models in $\MTL(\Gamma,1)$, which assume $\mathit{green}$ or $\mathit{push}$ in $T_0$,
admit weaker $H_0$'s where these atoms are not justified.
As a result, $\MEL(\Gamma, \lambda) = \tuple{T_i} _{\rangeo{i}{0}{\lambda}}$
with $T_i = \{\mathit{red}\}$ for all ${\rangeo{i}{0}{\lambda}}$.
To illustrate non-monotonicity, suppose we add the formula
\begin{align}
\next \mathit{push} \label{ex:traffic:light:next:push}
\end{align}
that ensures the button is pushed in the second state of the trace (i.e.\ at $k=1$) and take $\lambda=3$.
For readability sake, we represent traces $(T_0, T_1, T_2)$ as $T_0 \cdot T_1 \cdot T_2$.
For length $\lambda=3$, formula \eqref{ex:traffic:light:push} amounts in \MTL\ to requiring $\mathit{green}$ at $k=2$ whenever $\mathit{push}$ holds at any point.
Thus, given $\Gamma' = \Gamma \cup \{\eqref{ex:traffic:light:next:push}\}$
and $\lambda=3$, we get
\begin{align*}
\MTL(\Gamma', 3)
& = \left\lbrace
\tuple{\T}=\tuple{T_0 \cdot T_1 \cdot T_2} \mid
T_0 \in \{r,g,rp,gp\}, T_1 \in \{rp,gp\}, T_2 \in \{g,gp\}
\right\rbrace
\end{align*}
Now, for those $\tuple{\T,\T}$ with $T_0 \neq r$ and $T_2 = gp$, we always have, among others, a smaller model $\tuple{ \emptyset \cdot T_1 \cdot g ,\T}$ in $\MHT(\Gamma',3)$.
This means that, in \MEL, we conclude by default that we do not push in other situations $k\neq 1$ and the traffic light is red at the initial state.
From the remaining possible total models $\tuple{r \cdot rp \cdot g}$ and $\tuple{r \cdot gp \cdot g}$,
the latter is not in equilibrium, since $\tuple{r \cdot p \cdot g , \ \; r \cdot gp \cdot g}$ is an model in \MHT\ that
reveals that $\mathit{green}$ at $k=0$ is not justified.
As a result $\MEL(\Gamma',3)$ contains the unique temporal equilibrium model $\tuple{r \cdot rp \cdot g}$ and the \MEL\ conclusion $\next^2 r$,
we could obtain from $\Gamma$ alone, is not derived any more once \eqref{ex:traffic:light:next:push} is added to the theory.
In this example, we obtain one equilibrium model because $\metric{\alwaysF}{4} \mathit{green}$ becomes trivially true on traces shorter than $\lambda=4$. When the trace is long enough ($\lambda\geq 7$),  $\Gamma'$ generates the three expected temporal equilibrium models:
\[\arraycolsep=1.4pt
\begin{array}{cccccccccccccccc}
\langle r & \cdot & gp & \cdot & g & \cdot & g & \cdot & g & \cdot & r & \cdot & r & \cdot & r & \ \dots \rangle\\
\langle r & \cdot & rp & \cdot & g & \cdot & g & \cdot & g & \cdot & g & \cdot & r & \cdot & r & \ \dots \rangle\\
\langle r & \cdot & rp & \cdot & r & \cdot & g & \cdot & g & \cdot & g & \cdot & g & \cdot & r & \ \dots \rangle
\end{array}
\]

 In the following, we elaborate on the formal characteristics of our approach.
At first, we show that a basic property of \HT\ is maintained in \MHT:
\begin{proposition}[Persistence]\label{prop:persistence}
  Let $\tuple{\H,\T}$ be an \HT-trace of length $\lambda$ and $\varphi$ be a metric formula.
Then, for any $\rangeo{k}{0}{\lambda}$,
  \begin{enumerate}
  \item if $\tuple{\H,\T}, k \models \varphi$ then $\tuple{\T,\T}, k \models \varphi$.
  \item $\tuple{\H,\T}, k \models \neg\varphi$ iff $\tuple{\T,\T}, k \not\models \varphi$
  \end{enumerate}
\end{proposition}

All \MHT\ tautologies are \MTL\ tautologies but not vice versa (cf.~\eqref{eq:excluded:middle} above).
However, they coincide for some types of equivalences, as stated below.
\begin{proposition}\label{prop:nonimpl}Let $\varphi$ and $\psi$ be metric formulas without implications (and so, without negations either).
  Then, $\varphi \equiv \psi$ in \MTL\ iff $\varphi \equiv \psi$ in \MHT.
\end{proposition}
Another useful tool that can save some effort when proving groups of equivalences in \MHT{} is the use of duality properties.
A first type of duality has to do with the temporal direction (future or past) of the modal operators.
Let $\next/\previous$, $\wnext/\wprevious$,
$\metric{\until}{n}/\metric{\since}{n}$, $\metric{\release}{n}/\metric{\trigger}{n}$,
$\metric{\alwaysF}{n}/\metric{\alwaysP}{n}$ and $\metric{\eventuallyF}{n}/\metric{\eventuallyP}{n}$
denote all pairs of swapped-time connectives and
let $\sigma(\varphi)$ denote the replacement in $\varphi$ of each connective by its swapped-time version.
If we restrict ourselves to finite traces, we get the following result.
\begin{lemma}\label{lem:temporal:duality}
Let \M\ be an \HT-trace of length $\lambda$ and $\varphi$ be a metric formula.
Then, there exists an \HT-trace $\M'$ of length $\lambda$ such that
$\M,k \models \varphi$ iff $\M',{n\!-\!k} \models \sigma(\varphi)$
  for any $\rangeo{k}{0}{\lambda}$.
\end{lemma}\begin{theorem}[Temporal Duality Theorem]\label{thm:temporal:duality}
  A metric formula $\varphi$ is a tautology in \MHTf\ iff $\sigma(\varphi)$ is a tautology in \MHTf.
\end{theorem}
For instance, suppose we obtain a proof for
\begin{eqnarray}
\eventuallyF_n \; p \leftrightarrow p \vee \next \eventuallyF_{n-1} \; p \label{f:eventually}
\end{eqnarray}
Then, we can immediately apply Theorem~\ref{thm:temporal:duality} to guarantee that ${\eventuallyP}_n p \leftrightarrow p \vee \previous \eventuallyP_{n-1} \; p$ is a tautology too.
A second kind of duality has to do the analogy between Boolean disjunction and conjunction.
Let us define all the pairs of dual connectives as follows:
$\wedge / \vee$,
$\top / \bot$,
$\metric{\until}{n}/\metric{\release}{n}$,
$\next / \wnext$,
$\metric{\eventuallyF}{n}/\metric{\alwaysF}{n}$,
$\metric{\since}{n}/\metric{\trigger}{n}$,
$\previous / \wprevious$,
$\metric{\eventuallyP}{n}/\metric{\alwaysP}{n}$.
For a formula $\varphi$ without implications, we define $\delta (\varphi)$ as the result of replacing each connective
by its dual operator. Then, we get the following result.

\begin{proposition}[Boolean Duality]\label{prop:boolean:duality}
Let $\varphi$ and $\psi$ be formulas without implication.\footnote{Note that this also means without negation.}
Then,
$\varphi \leftrightarrow \psi$ is a tautology in \MHT\ iff
$\delta(\varphi) \leftrightarrow \delta(\psi)$ is a tautology in \MHT.
\end{proposition}
Following with our example of equivalence~\eqref{f:eventually}, we can  now apply Theorem~\ref{prop:boolean:duality} to conclude:
$$
\alwaysF_n \; p \leftrightarrow p \wedge \wnext \alwaysF_{n-1} \; p
$$

Next, we show how metric operators on formulas can be characterized inductively.
\begin{proposition}\label{prop:validities}
  The following formulas are valid in \MHT. For any numeral $n$ with ${\numeral{n}}\leq 0$:
  \[
    \begin{array}{ccccc}
      \varphi \metric{\until}{n} \psi \leftrightarrow \bot
      & \varphi \metric{\release}{n} \psi \leftrightarrow \top
      & \varphi \metric{\since}{n} \psi \leftrightarrow \bot
      & \varphi \metric{\trigger}{n} \psi \leftrightarrow \top
    \end{array}
  \]
For any numeral $n$ with ${\numeral{n}}>0$, we have
  \[
    \begin{array}{cc}
      \varphi \metric{\until}{n} \psi \leftrightarrow \psi \vee \left(\varphi \wedge \next
      ( \varphi \metric{\until}{n-1} \psi )\right)
      & \varphi \metric{\release}{n} \psi \leftrightarrow \psi \wedge \left(\varphi \vee \wnext
      ( \varphi \metric{\release}{n-1} \psi ) \right)
      \\[5pt]
      \varphi \metric{\since}{n} \psi \leftrightarrow \psi \vee \left(\varphi \wedge \previous
      ( \varphi \metric{\since}{n-1} \psi ) \right)
      & \varphi \metric{\trigger}{n} \psi \leftrightarrow \psi \wedge \left(\varphi \vee \wprevious
      (\varphi \metric{\trigger}{n-1} \psi )\right)
    \end{array}
  \]
\end{proposition}
The propositions above allow us to unfold metric operators containing numerals.
It is easy to see that, for $n=1$, the four operators collapse to the formula $\psi$ on their right.
For instance,
$\metric{\eventuallyF}{5} \mathit{clean}
\leftrightarrow
\mathit{clean} \vee \next\metric{\eventuallyF}{4} \mathit{clean}$ whereas $\metric{\eventuallyF}{1} \mathit{clean} \leftrightarrow \mathit{clean}$ and $\metric{\eventuallyF}{0} \mathit{clean} \leftrightarrow \bot$.

For metric operators depending on the trace length, the value of which is not necessarily known, Proposition \ref{prop:validities} cannot be
applied.
Instead, we have the following tautologies.
\begin{proposition}\label{prop:validity:l}
  The following formulas are valid in \MHT:
  \[
    \begin{array}{cc}
      \varphi \metric{\until}{\last} \psi
      \leftrightarrow
      \psi \vee \left(\varphi \wedge \next ( \varphi \metric{\until}{\last} \psi )\right)
      & \varphi \metric{\release}{\last} \psi
        \leftrightarrow
        \psi \wedge \left(\varphi \vee \wnext ( \varphi \metric{\release}{\last} \psi )\right)
      \\[5pt]
      \varphi \metric{\since}{\last} \psi
      \leftrightarrow
      \psi \vee \left(\varphi \wedge \previous ( \varphi \metric{\since}{\last} \psi )\right)
      & \varphi \metric{\trigger}{\last} \psi
        \leftrightarrow
        \psi \wedge \left(\varphi \vee \wprevious ( \varphi \metric{\trigger}{\last} \psi ) \right)
    \end{array}
  \]
\end{proposition}
That is, when the limit is the trace length $\last$, the unfolding contains the same operator, it is not altered.
As an example, we consider the machine that has to be cleaned eventually before the end of the trace.
We then have $\metric{\eventuallyF}{\last} \mathit{clean}
\leftrightarrow
\mathit{clean} \vee \next\metric{\eventuallyF}{\last} \mathit{clean}$.

 Alternatively, metric operators may also be parametrized by intervals\footnote{This concept of interval operators should not be confounded with the ones of Allen's interval algebra~\cite{allen83a}.}
rather than a mere upper bound.
In our setting, this is however no restriction since such metric operators can also be expressed,
as we show next.

The definition of our interval operators then depends on the type of numeral expression involved.
For numeral constants $n,m$ where $\numeral{n}, \numeral{m} \in \intervo{0}{\lambda}$, we define:
\[
\begin{array}{rcl}
\alwaysP_{\left[ m;n\right)}\; \varphi      & \eqdef & \wprevious^{\numeral{m}} \; \metric{\alwaysP}{n{-}m} \varphi     \\
\eventuallyP_{\left[ m;n\right)} \; \varphi  & \eqdef & \previous^{\numeral{m}} \; \metric{\eventuallyP}{n{-}m} \varphi \\
\varphi \since_{\left[ m;n\right)} \psi   & \eqdef &  \previous^{\numeral{m}}\; ( \varphi \metric{\since}{n{-}m} \psi ) \\
\varphi \trigger_{\left[ m;n\right)} \psi & \eqdef & \wprevious^{\numeral{m}} \; (\varphi \metric{\trigger}{n{-}m} \psi)
\end{array}
\qquad\qquad
\begin{array}{rcl}
\alwaysF_{\left[ m;n\right)} \; \varphi      & \eqdef & \wnext^{\numeral{m}} \; \metric{\alwaysF}{n{-}m} \varphi     \\
\eventuallyF_{\left[ m;n\right)} \; \varphi  & \eqdef & \next^{\numeral{m}} \; \metric{\eventuallyF}{n{-}m} \varphi \\
\varphi \until_{\left[ m;n\right)} \psi   & \eqdef &  \next^{\numeral{m}} \; (\varphi \metric{\until}{n{-}m} \psi) \\
\varphi \release_{\left[ m;n\right)} \psi & \eqdef & \wnext^{\numeral{m}} \; (\varphi \metric{\release}{n{-}m} \psi)
\end{array}
\]

For intervals spanning to the end (or beginning) of the trace, we have:
\[
\begin{array}{rcl}
\alwaysP_{\left[ m;\last\right)} \; \varphi      & \eqdef & \wprevious^{\numeral{m}} \; \metric{\alwaysP}{\last} \varphi     \\
\eventuallyP_{\left[ m;\last\right)} \; \varphi  & \eqdef & \previous^{\numeral{m}} \; \metric{\eventuallyP}{\last} \varphi \\
\varphi \since_{\left[ m;\last\right)} \psi   & \eqdef &  \previous^{\numeral{m}} \; (\varphi \metric{\since}{\last} \psi) \\
\varphi \trigger_{\left[ m;\last\right)} \psi & \eqdef & \wprevious^{\numeral{m}} \; (\varphi \metric{\trigger}{\last} \psi)
\end{array}
\qquad\qquad
\begin{array}{rcl}
\alwaysF_{\left[ m;\last\right)} \; \varphi      & \eqdef & \wnext^{\numeral{m}} \; \metric{\alwaysF}{\last} \varphi     \\
\eventuallyF_{\left[ m;\last\right)} \; \varphi  & \eqdef & \next^{\numeral{m}} \; \metric{\eventuallyF}{\last} \varphi \\
\varphi \until_{\left[ m;\last\right)} \psi   & \eqdef &  \next^{\numeral{m}} \; (\varphi \metric{\until}{\last} \psi) \\
\varphi \release_{\left[ m;\last\right)} \psi & \eqdef & \wnext^{\numeral{m}} \; (\varphi \metric{\release}{\last} \psi)
\end{array}
\]
As an example of an interval formula, reconsider the machine, discussed in the beginning, and assume that
it cannot be cleaned immediately but only within 3 to 5 time steps after usage.
This can be expressed as
$
 \alwaysF (
 \mathit{use} \to \eventuallyF _{ \left[3;5\right)} \mathit{clean} )
$.
In the definitions, note that when $m \geq n$ the interval $[m;n)$ is empty and the operator is always reducible to a truth constant.
For instance, $\eventuallyF_{\left[ 5;3 \right)} \varphi$ becomes $\next^{5} \metric{\eventuallyF}{-2} \varphi$ which amounts to $\next^{5} \bot$ or simply $\bot$.
For this reason, we do not consider intervals with $\last$ as lower bound, since they are always empty by definition.
\begin{proposition}[Satisfaction]\label{prop:satisfaction:interval}
  Let $\M=\tuple{\H,\T}$ be an \HT-trace of length $\lambda$ over $\mathcal{A}$.
  Given the respective definitions of derived operators, we get the following satisfaction conditions:
  \begin{enumerate} \setcounter{enumi}{19}
\item $\M, k \models \varphi {\until}_{\left[ m;n\right)} \psi$
    iff
    for some $\rangeo{j}{\numeral{m}}{\numeral{n}}$ such that
    $k{+}j \in \intervo{0}{\lambda}$ we have $\M, k{+}j \models \psi$
    and
    $\M, k{+}i \models \varphi$ for all $\rangeo{i}{0}{j}$
\item $\M, k \models \varphi {\release}_{\left[ m;n\right)} \psi$
    iff
    for all $\rangeo{j}{\numeral{m}}{\numeral{n}}$ such that
    $k{+}j \in \intervo{0}{\lambda}$, we have
    $\M, k{+}j \models \psi$
    or
    $\M, k{+}i \models \varphi$ for some $\rangeo{i}{0}{j}$
\item $\M, k \models \varphi {\since}_{\left[ m;n\right)} \psi$
    iff
    for some $\rangeo{j}{\numeral{m}}{\numeral{n}}$ such that
    $k{-}j \in \intervo{0}{\lambda}$ we have $\M, k{-}j \models \psi$
    and
    $\M, k{-}i \models \varphi$ for all $\rangeo{i}{0}{j}$
\item $\M, k \models \varphi {\trigger}_{\left[ m;n\right)} \psi$
    iff
    for all $\rangeo{j}{\numeral{m}}{\numeral{n}}$ such that
    $k{-}j \in \intervo{0}{\lambda}$, we have
    $\M, k{-}j \models \psi$
    or
    $\M, k{-}i \models \varphi$ for some $\rangeo{i}{0}{j}$
\item $\M, k \models {\eventuallyF}_{\left[ m;n\right)} \varphi$
    iff
    for some $\rangeo{j}{\numeral{m}}{\numeral{n}}$ such that
    $k{+}j \in \intervo{0}{\lambda}$ we have $\M, k{+}j \models \varphi$
\item $\M, k \models {\alwaysF}_{\left[ m;n\right)} \varphi$
    iff
    for all $\rangeo{j}{\numeral{m}}{\numeral{n}}$ such that
    $k{+}j \in \intervo{0}{\lambda}$, we have
    $\M, k{+}j \models \varphi$
\item $\M, k \models {\eventuallyP}_{\left[ m;n\right)} \varphi$
    iff
    for some $\rangeo{j}{\numeral{m}}{\numeral{n}}$ such that
    $k{-}j \in \intervo{0}{\lambda}$ we have $\M, k{-}j \models \varphi$
\item $\M, k \models {\alwaysP}_{\left[ m;n\right)} \varphi$
    iff
    for all $\rangeo{j}{\numeral{m}}{\numeral{n}}$ such that
    $k{-}j \in \intervo{0}{\lambda}$, we have
    $\M, k{-}j \models \varphi$
  \end{enumerate}
\end{proposition}
Note that one-step operators can be represented in terms of intervals, since we have
\begin{align*}
\next \varphi = \next \metric{\eventuallyF}{1} \varphi = \eventuallyF_{[1;2)}\varphi
\quad & \quad
\previous \varphi = \previous \metric{\eventuallyP}{1} \varphi = \eventuallyP_{[1;2)}\varphi \\
\wnext \varphi = \wnext \metric{\alwaysF}{1} \varphi = \alwaysF_{[1;2)}\varphi
\quad & \quad
\wprevious \varphi = \wprevious \metric{\alwaysP}{1} \varphi = \alwaysP_{[1;2)}\varphi
\end{align*}

 Next, we present a three-valued semantics for \MHT\
which turns out to be particularly useful for formal elaborations.
In particular, this three-valued interpretation has an important advantage: it allows the interchange of subformulas in a larger formula provided that the interchanged subformulas have the same three-valued interpretations.
This characterization relies on temporal three-valued interpretation~\cite{cabalar10a} and is inspired,
in its turn,
by the characterization of \HT\ in terms of G\"odel's logic $G_3$~\cite{goedel32a}.
Under this orientation, we deal with three truth values $\{0,1,2\}$ standing for:
$2$ (or proved true) meaning satisfaction ``here'';
$0$ (or assumed false) meaning falsity ``there''; and
$1$ (potentially true) for formulas assumed but not proved to be true.
Given an \HT-trace $\tuple{\H,\T}$,
we define its associated \emph{truth valuation} as a function $\trival{k}{\varphi}$ that
assigns a truth value in the set $\{0,1,2\}$ to a metric formula $\varphi$ at time point $\rangeo{k}{0}{\lambda}$
as follows.
For propositional connectives, $\trival{k}{\varphi}$ directly corresponds to $G_3$, that is, conjunction is the minimum, disjunction the maximum and implication $\trival{k}{\varphi \to \psi}$ is 2 if $\trival{k}{\varphi}\leq \trival{k}{\psi}$, or is $\trival{k}{\psi}$ otherwise.
For the rest of cases, we have:
\begin{eqnarray*}
	\trival{k}{\bot} & \eqdef & 0 \\
	\trival{k}{\top} & \eqdef & 2 \\
	\trival{k}{p}    & \eqdef & \begin{cases}
                                      0 & \text{if} \ p \not\in T_k \\
                                      1 & \text{if} \ p \in T_k\setminus H_k \\
                                      2 & \text{if} \ p \in H_k
                                    \end{cases}
                                    \quad\text{ for any atom } p \in \mathcal{A}\\
	\trival{k}{\next \varphi}      & \eqdef & \begin{cases}
                                                    0 & \text{if } k+1=\lambda;\\
                                                    \trival{k+1}{\varphi} & \text{otherwise}
                                                  \end{cases}\\
	\trival{k}{\wnext \varphi}     & \eqdef & \begin{cases}
                                                    2 & \text{if } k+1=\lambda;\\
                                                    \trival{k+1}{\varphi} & \text{otherwise}
                                                  \end{cases}\\
	\trival{k}{\previous \varphi}  & \eqdef & \begin{cases}
                                                    0 & \text{if } k=0;\\
                                                    \trival{k-1}{\varphi} & \text{otherwise}
                                                  \end{cases}\\
	\trival{k}{\wprevious \varphi} & \eqdef & \begin{cases}
                                                    2 & \text{if } k=0;\\
                                                    \trival{k-1}{\varphi} & \text{otherwise}
                                                  \end{cases}
\end{eqnarray*}
For $n$ being a numeral constant or the constant $\last$,
we have\footnote{Here, we assume that $\max(\emptyset)=0$ and $\min(\emptyset)=2$.}
\begin{eqnarray*}
\trival{k}{\varphi \metric{\until}{n} \psi} & \eqdef & \max\lbrace \min\lbrace \trival{k+i}{\psi}, \trival{k+j}{\varphi} \mid \rangeo{j}{0}{i},\ k+i < \lambda \rbrace   \mid  0 \le i < \numeral{n} \rbrace \\
\trival{k}{\varphi \metric{\release}{n} \psi} &\eqdef& \min\lbrace  \max\lbrace \trival{k+i}{\psi}, \trival{k+j}{\varphi} \mid \rangeo{j}{0}{i},\ k+i < \lambda \rbrace \mid 0 \le i < \numeral{n} \rbrace\\
\trival{k}{\varphi \metric{\since}{n} \psi} & \eqdef & \max\lbrace \min\lbrace \trival{k - i}{\psi}, \trival{k - j}{\varphi} \mid \rangeo{j}{0}{i},\ k-i \ge 0 \rbrace   \mid 0 \le i < \numeral{n} \rbrace \\
\trival{k}{\varphi \metric{\trigger}{n} \psi} & \eqdef & \min\lbrace \max\lbrace \trival{k - i}{\psi}, \trival{k - j}{\varphi} \mid \rangeo{j}{0}{i},\ k-i \ge 0 \rbrace   \mid 0 \le i < \numeral{n} \rbrace
\end{eqnarray*}

\begin{proposition}\label{prop:three-valued}
  Let $\tuple{\H,\T}$ be a \HT-trace of length $\lambda$,
  ${\bm m}$ its associated valuation and $\rangeo{k}{0}{\lambda}$.
Then, for any formula $\varphi$, we have
  \begin{itemize}
  \item $\tuple{\H,\T}, k \models \varphi$ iff $\trival{k}{\varphi} = 2$ and
  \item $\tuple{\T,\T}, k \models \varphi$ iff $\trival{k}{\varphi} \not = 0$.
  \end{itemize}
\end{proposition}

 \section{Metric and Temporal Equilibrium Logic}
\label{sec:mel:tel}

In this section,
we study the relation between metric and temporal (equilibrium) logics.
In fact, temporal formulas constitute a subclass of metric formulas.
\[
  \begin{array}{r@{\ }c@{\ }lp{3pt}r@{\ }c@{\ }lp{3pt}r@{\ }c@{\ }lp{3pt}r@{\ }c@{\ }l}
    \eventuallyF \varphi  & \eqdef & \metric{\eventuallyF}{\last} \varphi  &&
    \alwaysF \varphi      & \eqdef & \metric{\alwaysF}{\last} \varphi      &&
    \eventuallyP \varphi  & \eqdef & \metric{\eventuallyP}{\last} \varphi  &&
    \alwaysP \varphi      & \eqdef & \metric{\alwaysP}{\last} \varphi
    \\
    \varphi \until \psi   & \eqdef & \varphi \metric{\until}{\last}\psi    &&
    \varphi \release \psi & \eqdef & \varphi \metric{\release}{\last} \psi &&
    \varphi \since \psi   & \eqdef & \varphi \metric{\since}{\last}\psi    &&
    \varphi \trigger \psi & \eqdef & \varphi \metric{\trigger}{\last} \psi
\end{array}
\]
The next result guarantees that the original semantics of the temporal operators in \THT~\cite{cakascsc18a} is preserved.
\begin{proposition}[Satisfaction]\label{prop:satisfaction:tel}
  Let $\M=\tuple{\H,\T}$ be an \HT-trace of length $\lambda$ over $\mathcal{A}$.
  Given the respective definitions of derived operators, we get the following satisfaction conditions:
  \begin{enumerate} \setcounter{enumi}{27}
  \item $\M, k \models \eventuallyF \varphi$
    iff
    $\M, i \models \varphi$ for some $\rangeo{i}{k}{\lambda}$
  \item $\M, k \models \alwaysF \varphi$
    iff
    $\M, i \models \varphi$ for all $\rangeo{i}{k}{\lambda}$
  \item $\M, k \models \varphi \until \psi$
    iff
    for some $\rangeo{j}{k}{\lambda}$, we have
    $\M, j \models \psi$
    and
    $\M, i \models \varphi$ for all $\rangeo{i}{k}{j}$
  \item $\M, k \models \varphi \release \psi$
    iff
    for all $\rangeo{j}{k}{\lambda}$, we have
    $\M, j \models \psi$
    or
    $\M, i \models \varphi$ for some $\rangeo{i}{k}{j}$
  \item $\M, k \models \eventuallyP\varphi $
    iff
    $\M, i \models \varphi$ for some $\rangec{i}{0}{k}$
  \item $\M, k \models \alwaysP\varphi$
    iff
    $\M, i \models \varphi$ for all $\rangec{i}{0}{k}$
  \item $\M, k \models \varphi \since \psi$
    iff
    for some $\rangec{j}{0}{k}$, we have
    $\M, j \models \psi$
    and
    $\M, i \models \varphi$ for all $\orange{i}{j}{k}$
  \item $\M, k \models \varphi \trigger \psi$
    iff
    for all $\rangec{j}{0}{k}$, we have
    $\M, j \models \psi$
    or
    $\M, i \models \varphi$ for some $\orange{i}{j}{k}$
  \end{enumerate}
\end{proposition}

Interestingly,
it turns out that metric formulas can also be translated into temporal formulas.
This is due to the discrete time domain of \MHT\ and the resulting semantic structure common to \MHT\ and \THT.
In fact,
we provide two alternative translations possessing complementary properties.

\mysubsection{Language-preserving translation}
Our first translation refrains from extending the original language and
is independent of the length of the trace (and so the specific value of \last).
Although this allows us to search for models of varying length without recompiling a formula,
when using temporal ASP solvers such as \telingo~\cite{cakamosc19a},
the translation suffers from an exponential blowup in the worst case.

We define the translation recursively as follows:
\par
\newcommand{\translation}[1]{\ensuremath{\tau({#1})}} \begin{align*}
\translation{a} &\eqdef a \text{ for } a \in \mathcal{A} \\
\translation{\oplus\varphi} &\eqdef \oplus \translation{\varphi} \text{ for } \oplus \in \{ \neg, \previous, \next, \wprevious, \wnext \} \\
\translation{\varphi {\otimes} \psi}
	&\eqdef \translation{\varphi} \otimes \translation{\psi} \text{ for } \otimes \in \{ \wedge, \vee, \to \} \\
\translation{\varphi \metric{\otimes}{\last} \psi}
	&\eqdef \translation{\varphi} \otimes \translation{\psi} \text{ for } \otimes \in \{ \until, \release, \since, \trigger \} \\
\translation{\varphi \metric{\otimes}{1} \psi}
	&\eqdef \translation{\psi} \text{ for } \otimes \in \{ \until, \release, \since, \trigger \} \\
\translation{\varphi \metric{\until}{n} \psi}
	&\eqdef \translation{\psi \vee (\varphi \wedge \next (\varphi \metric{\until}{n{-}1} \psi) )} \\
\translation{\varphi \metric{\release}{n} \psi}
	&\eqdef \translation{\psi \wedge (\varphi \vee \wnext (\varphi \metric{\release}{n{-}1} \psi ))} \\
\translation{\varphi \metric{\since}{n} \psi}
	&\eqdef \translation{\psi \vee (\varphi \wedge \previous (\varphi \metric{\since}{n{-}1} \psi ))} \\
\translation{\varphi \metric{\trigger}{n} \psi}
	&\eqdef \translation{\psi \wedge (\varphi \vee \wprevious ( \varphi \metric{\trigger}{n{-}1} \psi ))} \\
\end{align*}

Translating formula \eqref{ex:traffic:light:push} from the traffic light example into a temporal formula yields
\begin{equation}\label{ex:translation:traffic:light:language:preserving}
\begin{aligned}
&\hspace{-22mm}\translation{\alwaysF\big(
\mathit{push} \to \metric{\eventuallyF}{3} \metric{\alwaysF}{4} \mathit{green}
\big)} \\
=
 \alwaysF
(
\mathit{push} \to (
	&(\mathit{green} \wedge \wnext \mathit{green} \wedge
	\wnext\wnext \mathit{green} \wedge \wnext\wnext\wnext \mathit{green}) \\
&\vee
	(\next (\mathit{green} \wedge \wnext \mathit{green}
	\wedge \wnext\wnext \mathit{green} \wedge \wnext\wnext\wnext \mathit{green} )) \\
&\vee
	(\next\next (\mathit{green} 	\wedge \wnext \mathit{green}
	\wedge \wnext\wnext \mathit{green} \wedge \wnext\wnext\wnext \mathit{green} ))
))
\end{aligned}
\end{equation}
This illustrates the benefit of metric operators.
While we are also able to express the same formula with temporal operators,
it is much more concise and more readable with metric operators.

\begin{proposition}
  The translation \translation{\varphi} terminates for any metric formula $\varphi$.
\end{proposition}
\begin{proposition}\label{lem:translation:correctness}
  For any \HT-trace \M\ and any time point $\rangeo{k}{0}{\lambda}$, we have

  $\M, k \models \varphi$ in \MHT\ iff $\M, k \models \translation{\varphi}$ in \THT.
\end{proposition}
\begin{corollary}\label{thm:melf:telf}
  For any metric formula $\varphi$ over $\mathcal{A}$,
  there exists
  a  temporal formula $\psi$ over $\mathcal{A}$ such that
  an \HT-trace \M\ is a model for $\psi$ in \THT\
  iff
  \M\ is a model for $\varphi$ in \MHT.
\end{corollary}

\mysubsection{Language-extending translation}
For a complement,
we provide an alternative translation using an extended alphabet.
This translation has the advantage of avoiding applications of distributivity, a source of an exponential increase in size.

To this end,
we adapt the notion of closure from~\cite{cadilasc20a} to provide a translation of metric into temporal formulas.
The original definition of closure is due to~\cite{fislad79a}.
\begin{definition}[Closure]
  The closure $\cl(\gamma)$ of a metric formula $\gamma$ is the subset minimal set of formulas satisfying the inductive conditions:
  \begin{enumerate}
  \item $\gamma \in \cl(\gamma)$
  \item $(\varphi \otimes \psi) \in \cl(\gamma)$ implies $\varphi \in \cl(\gamma)$ and $\psi \in \cl(\gamma)$
    \par for $\otimes \in \lbrace \wedge, \vee, \rightarrow, \metric{\until}{n},\metric{\release}{n},\metric{\since}{n},\metric{\trigger}{n}\rbrace$ with $n \in \mathbb{N}\cup\last$
  \item If $\otimes \psi \in \cl(\gamma)$ then $\psi \in \cl(\gamma)$
    for $\otimes \in \lbrace \next, \wnext,\previous,\wprevious \rbrace$
  \item If $\varphi \metric{\until}{n} \psi \in \cl(\gamma)$ and $n > 1$ then $\next \left(\varphi \metric{\until}{n-1} \psi\right) \in \cl(\gamma)$
  \item If $\varphi \metric{\release}{n} \psi \in \cl(\gamma)$ and $n > 1$ then $\wnext \left(\varphi \metric{\release}{n-1} \psi\right) \in \cl(\gamma)$
  \item If $\varphi \metric{\since}{n} \psi \in \cl(\gamma)$ and $n > 1$ then $\previous \left(\varphi \metric{\since}{n-1} \psi\right) \in \cl(\gamma)$
  \item If $\varphi \metric{\trigger}{n} \psi \in \cl(\gamma)$ and $n > 1$ then $\wprevious \left(\varphi \metric{\trigger}{n-1} \psi\right) \in \cl(\gamma)$
  \item If $\varphi \metric{\until}{\last} \psi \in \cl(\gamma)$ then $\next \left(\varphi \metric{\until}{\last} \psi\right) \in \cl(\gamma)$
  \item If $\varphi \metric{\release}{\last} \psi \in \cl(\gamma)$ then $\wnext \left(\varphi \metric{\release}{\last} \psi\right) \in \cl(\gamma)$
  \item If $\varphi \metric{\since}{\last} \psi \in \cl(\gamma)$ then $\previous \left(\varphi \metric{\since}{\last} \psi\right) \in \cl(\gamma)$
  \item If $\varphi \metric{\trigger}{\last} \psi \in \cl(\gamma)$ then $\wprevious \left(\varphi \metric{\trigger}{\last} \psi\right) \in \cl(\gamma)$
  \end{enumerate}
  Any set satisfying these conditions is called \emph{closed}.
\end{definition}

\begin{proposition}\label{prop:closure:finite}
  For any metric formula $\varphi$, $\cl(\varphi)$ is finite. Moreover, the total size of all the formulas in $\cl(\varphi)$, $\lvert \cl(\varphi)\rvert$ is bound by $\lvert \cl(\varphi)\rvert \le 2k_\varphi \lvert \varphi \rvert$, where $k_\varphi = \max\{1,n_\varphi\}$ and $n_\varphi$ is the maximum (metric) integer subindex occurring in $\varphi$.
\end{proposition}

Thus, given a metric formula $\varphi$ over alphabet $\mathcal{A}$ at hand,
we define the extended alphabet $\mathcal{A}^{\varphi} \eqdef \mathcal{A} \cup \{\Lab{\mu} \mid \mu \in \cl(\varphi)\}$.
For convenience, we simply use $\Lab{\varphi} \overset{\mathit{def}}{=} \varphi $ if $\varphi$ is $\top , \bot$ or an atom $a \in \mathcal{A}$.

As happened with the normal form reduction for \TELf\ in~\cite{cakascsc18a}, the translation is done in two phases: we first obtain a temporal theory containing double implications, and then we unfold them into  temporal rules.
We start by defining the temporal theory $\upsilon(\varphi)$ that introduces new labels $\Lab{\mu}$ for each formula $\mu \in \cl(\varphi)$.
This theory contains the formula $\Lab{\varphi}$ and, per each label $\Lab{\mu}$,
a set of formulas $\Tra{\mu}$ fixing the label's truth value.
Formally, we define that
\begin{align*}
  \upsilon(\varphi) =
  \left\lbrace \Lab{\varphi} \right\rbrace
  \cup \left\lbrace \Tra{\mu}
  \mid \mu \in \cl(\varphi)\right\rbrace
  \qquad\text{ and }\qquad
  \upsilon(\Gamma)=\{\upsilon(\varphi)\mid\varphi\in\Gamma\}.
\end{align*}

Table~\ref{tab:translation} shows the definitions $\Tra{\mu}$ for each $\mu$ in the closure $\cl(\varphi)$ depending on the outer modality in the formula.
\begin{table}[h!]
\centering
	\[
	\renewcommand{\arraystretch}{1.5}
	\begin{array}{|c|l|}
	\cline{1-2}
	\mu\in \cl(\varphi) & \Tra{\mu} \\ \cline{1-2}
	\next\alpha &\wnext \alwaysF (\previous \Lab{\mu} \leftrightarrow \Lab{\alpha}) \qquad  \alwaysF (\finally \to \neg \Lab{\mu}) \\ \cline{1-2}
	\previous \alpha & \wnext \alwaysF ( \Lab{\mu} \leftrightarrow \previous \Lab{\alpha}) \qquad   \neg \Lab{\mu} \\ \cline{1-2}
	\wnext \alpha & \wnext \alwaysF (\previous \Lab{\mu} \leftrightarrow \Lab{\alpha}) \qquad \alwaysF (\finally \to \Lab{\mu}) \\ \cline{1-2}
	\wprevious \alpha & \wnext \alwaysF ( \Lab{\mu} \leftrightarrow \previous \Lab{\alpha}) \qquad   \Lab{\mu} \\ \cline{1-2}
\alpha \metric{\until}{1}\beta &\alwaysF (\Lab{\mu} \leftrightarrow \Lab{\beta}) \\ \cline{1-2}
	\alpha \metric{\release}{1}\beta &\alwaysF (\Lab{\mu} \leftrightarrow \Lab{\beta}) \\ \cline{1-2}
	\alpha \metric{\since}{1}\beta & \alwaysF(\Lab{\mu} \leftrightarrow \Lab{\beta}) \\ \cline{1-2}
	\alpha \metric{\trigger}{1}\beta & \alwaysF (\Lab{\mu} \leftrightarrow \Lab{\beta}) \\ \cline{1-2}
\alpha \metric{\until}{n}\beta &\alwaysF \left(\Lab{\mu} \leftrightarrow \Lab{\beta} \vee \left(\Lab{\alpha} \wedge  \Lab{\alpha'}\right)\right) \qquad \text{with } \alpha'=\next \left( \alpha \metric{\until}{n-1} \beta \right) \\ \cline{1-2}
	\alpha \metric{\release}{n}\beta &\alwaysF \left(\Lab{\mu} \leftrightarrow \Lab{\beta} \wedge \left(\Lab{\alpha} \vee  \Lab{\alpha'}\right)\right) \qquad \text{with } \alpha'=\wnext \left( \alpha \metric{\release}{n-1} \beta \right) \\ \cline{1-2}
	\alpha \metric{\since}{n}\beta &\alwaysF \left(\Lab{\mu} \leftrightarrow \Lab{\beta} \vee \left(\Lab{\alpha} \wedge  \Lab{\alpha'}\right)\right) \qquad \text{with } \alpha'=\previous \left( \alpha \metric{\since}{n-1} \beta \right) \\ \cline{1-2}
	\alpha \metric{\trigger}{n}\beta &\alwaysF \left(\Lab{\mu} \leftrightarrow \Lab{\beta} \wedge \left(\Lab{\alpha} \vee  \Lab{\alpha'}\right)\right) \qquad \text{with } \alpha'=\wprevious \left( \alpha \metric{\trigger}{n-1} \beta \right) \\ \cline{1-2}
	\alpha \metric{\until}{\last}\beta &\alwaysF \left(\Lab{\mu} \leftrightarrow \Lab{\beta} \vee \left(\Lab{\alpha} \wedge  \Lab{\alpha'}\right)\right) \qquad \text{with } \alpha'=\next \left( \alpha \metric{\until}{\last} \beta \right) \\ \cline{1-2}
\alpha \metric{\release}{\last}\beta &\alwaysF \left(\Lab{\mu} \leftrightarrow \Lab{\beta} \wedge \left(\Lab{\alpha} \vee  \Lab{\alpha'}\right)\right) \qquad \text{with } \alpha'=\wnext \left( \alpha \metric{\release}{\last} \beta \right) \\ \cline{1-2}
\alpha \metric{\since}{\last}\beta &\alwaysF \left(\Lab{\mu} \leftrightarrow \Lab{\beta} \vee \left(\Lab{\alpha} \wedge  \Lab{\alpha'}\right)\right) \qquad \text{with } \alpha'=\previous \left( \alpha \metric{\since}{\last} \beta \right) \\ \cline{1-2}
\alpha \metric{\trigger}{\last}\beta &\alwaysF \left(\Lab{\mu} \leftrightarrow \Lab{\beta} \wedge \left(\Lab{\alpha} \vee  \Lab{\alpha'}\right)\right) \qquad \text{with } \alpha'=\wprevious \left( \alpha \metric{\trigger}{\last} \beta \right) \\ \cline{1-2}
	\end{array}
	\]
	\caption{Translation of metric modal operators}
	\label{tab:translation}
\end{table}

As we have seen, in the general case, formulas in $\Tra{\mu}$ are not temporal rules, since they sometimes contain double implications.
However, they all have the forms $\varphi$, $\alwaysF \varphi$, $\wnext \alwaysF \varphi$ or $\alwaysF (\finally \to \varphi)$, for some inner propositional formula $\varphi$ formed with temporal literals.

Given an \HT-trace\
\(
\tuple{\H,\T} = \tuple{H_i,T_i}_{i \in \intervo{0}{\lambda}}
\),
we define its restriction to alphabet $\mathcal{A}$
as
\(
\tuple{\H,\T}|_{\mathcal{A}}
\eqdef
\tuple{H_i\cap\mathcal{A},T_i\cap\mathcal{A}}_{\rangeo{i}{0}{\lambda}}
\).
Similarly, for any set $\mathfrak{S}$ of \HT-traces, we write $\mathfrak{S}|_{\mathcal{A}}$ to stand for
$\{ \tuple{\H,\T}|_{\mathcal{A}} \mid \tuple{\H,\T} \in \mathfrak{S}\}$.

The following lemma shows that $\Lab{\mu}$ and $\mu$ are equivalent:
\begin{proposition}\label{lem:nf2}
  Let $\gamma$ be a metric formula over $\mathcal{A}$ and
  let $\tuple{\H,\T}$ be a model in \MHTf\
  of $\upsilon(\gamma)$ being associated with the three-valuation ${\bm m}$.

  Then, for any $\mu \in \cl(\gamma)$ and any $\rangeo{k}{0}{\lambda}$,
  we have $\trival{k}{\Lab{\mu}} = \trival{k}{\mu}$.
\end{proposition}
\begin{theorem}\label{lem:nf1}
  For any metric formula $\varphi$ and any length $\lambda$, we have
  $$
  \MHTf(\varphi,\lambda)
  = \MHTf(\upsilon(\varphi),\lambda)|_{\mathcal{A}}
  = \THTf(\upsilon(\varphi),\lambda)|_{\mathcal{A}}.
  $$
\end{theorem}
\begin{corollary}
  For any temporal formula $\varphi$ and any length $\lambda$, we have
  $$
  \THT(\varphi, \lambda) = \MHT(\varphi, \lambda)
  $$
\end{corollary}
For example, the formula $\alwaysF (\neg \mathit{green} \to \mathit{red})$
in \eqref{ex:traffic:light:default}  has the same models whether it is interpreted
as a temporal or a metric formula.

\begin{corollary}\label{tel:in:mel}
  Let $\varphi$ be a metric formula over $\mathcal{A}$.
Then, translation $\upsilon(\varphi)$ is \emph{strongly faithful}, that is:
  $$
  \MELf(\varphi \wedge \varphi') = \MELf(\upsilon(\varphi)\wedge \varphi')|_{\mathcal{A}}.
  $$
  for any arbitrary metric formula $\varphi'$ over $\mathcal{A}$.
\end{corollary}

The translation of formula \eqref{ex:traffic:light:push} from the traffic light example yields the resulting set of rules in Table~\ref{tab:example:translation} computed from
$\cl(\ref{ex:traffic:light:push})$.\begin{table}[ht]
\centering\newcommand{\LabX}[1]{\raisebox{11pt}{}{\Lab{#1}}}
\begin{tabular}{|c|c|c|}\cline{1-3}
\LabX{\mu} & $\mu \in \cl(\ref{ex:traffic:light:push})$ & \Tra{\mu}\\\cline{1-3}
\LabX{1} & $\alwaysF\left(\mathit{push} \to \metric{\eventuallyF}{3} \metric{\alwaysF}{4} \mathit{green}\right)$ & $\alwaysF\left(\Lab{1} \leftrightarrow \Lab{3} \wedge \Lab{2}\right)$\\\cline{1-3}
\LabX{2} & $\wnext  \alwaysF\left(\mathit{push} \to \metric{\eventuallyF}{3} \metric{\alwaysF}{4} \mathit{green}\right)$ & $\wnext \alwaysF\left(\previous \Lab{2} \leftrightarrow \Lab{1}\right)$ \qquad $\alwaysF\left( \finally \to \Lab{2}\right)$ \\\cline{1-3}
\LabX{3} & $\mathit{push} \to \metric{\eventuallyF}{3} \metric{\alwaysF}{4} \mathit{green}$ & $\alwaysF\left(\Lab{3} \leftrightarrow  \left(\mathit{push} \rightarrow \Lab{4} \right)\right)$\\\cline{1-3}
\LabX{4} & $\metric{\eventuallyF}{3} \metric{\alwaysF}{4} \mathit{green}$ & $\alwaysF\left(\Lab{4} \leftrightarrow  \Lab{9} \vee \Lab{5}\right)$\\\cline{1-3}
\LabX{5} & $\next\metric{\eventuallyF}{2} \metric{\alwaysF}{4} \mathit{green}$ & $\wnext\alwaysF\left(\previous \Lab{5} \leftrightarrow   \Lab{6}\right)$  \qquad  $\alwaysF\left( \finally \to \neg \Lab{5}\right)$\\\cline{1-3}
\LabX{6} & $\metric{\eventuallyF}{2} \metric{\alwaysF}{4} \mathit{green}$ & $\alwaysF\left(\Lab{6} \leftrightarrow   \Lab{9} \vee \Lab{7} \right)$  \\\cline{1-3}
\LabX{7} & $\next \metric{\eventuallyF}{1} \metric{\alwaysF}{4} \mathit{green}$ & $\wnext \alwaysF\left(\previous \Lab{7} \leftrightarrow   \Lab{8}\right)$  \qquad  $\alwaysF\left( \finally \to \neg \Lab{7}\right)$ \\\cline{1-3}
\LabX{8} & $\metric{\eventuallyF}{1} \metric{\alwaysF}{4} \mathit{green}$ & $\alwaysF\left(\Lab{8} \leftrightarrow  \Lab{9}\right)$\\\cline{1-3}
\LabX{9} & $\metric{\alwaysF}{4} \mathit{green}$ & $\alwaysF\left(\Lab{9} \leftrightarrow  \mathit{green} \wedge \Lab{10}\right)$\\\cline{1-3}
\LabX{10} & $\wnext\metric{\alwaysF}{3} \mathit{green}$ & $\wnext\alwaysF\left(\previous\Lab{10} \leftrightarrow  \Lab{11}\right)$  \qquad  $\alwaysF\left( \finally \to \Lab{10}\right)$\\\cline{1-3}
\LabX{11} & $\metric{\alwaysF}{3} \mathit{green}$ & $\alwaysF\left(\Lab{11} \leftrightarrow  \mathit{green} \wedge \Lab{12}\right)$\\\cline{1-3}
\LabX{12} & $\wnext\metric{\alwaysF}{2} \mathit{green}$ & $\wnext\alwaysF\left(\previous\Lab{12} \leftrightarrow  \Lab{13}\right)$  \qquad  $\alwaysF\left( \finally \to \Lab{12}\right)$\\\cline{1-3}
\LabX{13} & $\metric{\alwaysF}{2} \mathit{green}$ & $\alwaysF\left(\Lab{13} \leftrightarrow  \mathit{green} \wedge \Lab{14}\right)$ \\\cline{1-3}
\LabX{14} & $\wnext\metric{\alwaysF}{1} \mathit{green}$ & $\wnext \alwaysF\left(\previous \Lab{14} \leftrightarrow   \Lab{15}\right)$  \qquad  $\alwaysF\left( \finally \to \Lab{14}\right)$\\\cline{1-3}
\LabX{15} & $\metric{\alwaysF}{1} \mathit{green}$ & $\alwaysF\left(\Lab{15} \leftrightarrow  \mathit{green}\right)$ \\\cline{1-3}
\end{tabular}
\caption{Translating formula \eqref{ex:traffic:light:push} from the traffic light example.}
\label{tab:example:translation}
\end{table}
Although in this case, $\cl(\ref{ex:traffic:light:push})$ is larger than the language-preserving translation $\translation{\ref{ex:traffic:light:push}}$ = \eqref{ex:translation:traffic:light:language:preserving}, note that the latter is not in the form of a logic program yet, while double implications in Table~\ref{tab:example:translation} can be linearly reduced to a logic program.
In general, reducing $\translation{\gamma}$ to a (language-preserving) logic program induces an exponential size increase due to the application of distributivity laws.
For instance, the formula $\translation{p \until_n q}$ amounts to a combination of conjunctions and disjunctions that, to become a logic program, must be previously reduced to conjunctive normal form.
On the other hand, $\cl(\gamma)$ preserves a polynomial size, although at the cost of introducing auxiliary
atoms, one per formula in $\cl(\gamma)$.
 \section{Discussion}\label{sec:discussion}

We have defined and elaborated upon a metric temporal extension of the logic of Here-and-There
in order to lay the theoretical foundations of metric Answer Set Programming.
The resulting logics, \MHT\ and its non-monotonic extension \MEL,
have a point-based semantics based on discrete linear time.
The choice of such a simple time domain was motivated by the desire to base the approach on the same semantic structures as used in
previous extensions of \HT\ and ASP with constructs from Linear Temporal and Dynamic Logic,
namely, linear \HT-traces.
As a result, we were able to inter-translate our approach and the temporal logics of \THT\ and \TEL.
This is of practical relevance since it allows us to use temporal ASP solvers such as \telingo~\cite{cakamosc19a} for implementation.

There exist other approaches that introduce metric temporal operators in logic programming.
For instance, a first metric extension for Horn Prolog was presented back in~\cite{brzoska93a}.
This approach is more limited than \MEL{} for several reasons: it does not consider default negation, metric operators
can only be used under a restricted syntax and only for unary temporal operators,
and the interpretation of programs relies on resolution rather than on a purely model-theoretic description.
Other more recent related approaches introduce temporal metrics for
stream reasoning for ASP~\cite{bedaeifi15a,bedaei16a} and DATALOG~\cite{brkaryxiza18a,wakagr19a}.
The former is not only closely related due to its ASP-based approach but also because it can be characterized
in terms of \EL~\cite{bedaei16a}, although extra conditions are needed to guarantee the persistence property of \HT.
While the DATALOG-based approach lacks the rich language of ASP,
they rely on more expressive metric operators.
For instance, \cite{brkaryxiza18a} deal with intervals over $\mathbb{Q} \cup \{ -\infty, +\infty \}$.
It will be interesting to investigate the formal connection to such stream-oriented approaches and
metric temporal logic programming formalisms in general.

Another aspect to explore has to do with the fact that traditional metric logics rely on continuous time domains (cf.~\cite{aluhen92a}), which often brings about undecidability.
Although we do not strive for such expressiveness, it will be interesting future work to generalize our approach to more fine-grained time domains
using integer or rational numbers and to explore implementations with hybrid ASP systems.
However, the presented approach should already furnish a host system for action languages with durative actions~\cite{sobatu04a} or even traditional
event calculus~\cite{kowser86a}, although this remains to be worked out.

 \bibliographystyle{acmtrans}
\newpage
\appendix

\section{Proofs}
\label{sec:proofs}

\begin{proofof}{Proposition~\ref{prop:persistence}}
	The first item is proved by structural induction. We consider the different cases below:
	\begin{itemize}
		\item Case $\varphi \metric{\until}{n} \psi$: if $\tuple{\H,\T},k \models \varphi \metric{\until}{n} \psi$ then there exists $0\le i < \numeral{n}$ such that $k+i < \lambda$, $\tuple{\H,\T}, k + i \models \psi$ and for all $0 \le j < i$, $\tuple{\H,\T}, k + j \models \varphi$. By induction hypothesis, $\tuple{\T,\T}, k + i \models \psi$ and for all $0 \le j < i$, $\tuple{\T,\T}, k + j \models \varphi$. Therefore, $\tuple{\H,\T},k \models \varphi \metric{\until}{n} \psi$.
		
		\item Case $\varphi \metric{\release}{n} \psi$: assume by contradiction that $\tuple{\T,\T}, k \not \models \varphi \metric{\release}{n} \psi$. This means that there exists $0\le i < \numeral{n}$ such that $k+i < \lambda$, $\tuple{\T,\T}, k + i \not \models \psi$ and for all $0 \le j < i$, $\tuple{\T,\T}, k + j \not \models \varphi$. By induction, $\tuple{\H,\T}, k + i \not \models \psi$ and for all $0 \le j < i$, $\tuple{\H,\T}, k + j \not \models \varphi$. Therefore, $\tuple{\H,\T}, k \not \models \varphi \metric{\release}{n} \psi$: a contradiction.
		
		\item Case $\varphi \metric{\since}{n} \psi$: if $\tuple{\H,\T}, k \models \varphi \metric{\since}{n} \psi$ then there exists $0\le i < \numeral{n}$ such that $k-i \ge 0$, $\tuple{\H,\T}, k - i \models \psi$ and for all $0 \le j < i$, $\tuple{\H,\T}, k - j \models \varphi$. By the induction hypothesis, $\tuple{\T,\T}, k - i \models \psi$ and for all $0 \le j < i$, $\tuple{\T,\T}, k - j \models \varphi$. Therefore, $\tuple{\T,\T}, k \models \varphi \metric{\since}{n} \psi$

		\item Case $\varphi \metric{\trigger}{n} \psi$: assume by contradiction that $\tuple{\T,\T}, k \not \models \varphi \metric{\trigger}{n} \psi$. This means that there exists $0\le i < \numeral{n}$ such that $k-i \ge 0$, $\tuple{\T,\T}, k - i \not \models \psi$ and for all $0 \le j < i$, $\tuple{\T,\T}, k - j \not \models \varphi$. By induction hypothesis, $\tuple{\H,\T}, k - i \not \models \psi$ and for all $0 \le j < i$, $\tuple{\H,\T}, k - j \not \models \varphi$, so $\tuple{\H,\T}, k \not \models \varphi \metric{\trigger}{n} \psi$: a contradiction.

		\item Case $\varphi \metric{\until}{\last} \psi$: if $\tuple{\H,\T},k \models \varphi \metric{\until}{\last} \psi$ then there exists $0\le i < \lambda$ such that $k+i < \lambda$, $\tuple{\H,\T}, k + i \models \psi$ and for all $0 \le j < i$, $\tuple{\H,\T}, k + j \models \varphi$. By induction hypothesis, $\tuple{\T,\T}, k + i \models \psi$ and for all $0 \le j < i$, $\tuple{\T,\T}, k + j \models \varphi$. Therefore, $\tuple{\H,\T},k \models \varphi \metric{\until}{\last} \psi$.
		
		\item Case $\varphi \metric{\release}{\last} \psi$: assume by contradiction that $\tuple{\T,\T}, k \not \models \varphi \metric{\release}{\last} \psi$. This means that there exists $0\le i < \lambda$ such that $k+i < \lambda$, $\tuple{\T,\T}, k + i \not \models \psi$ and for all $0 \le j < i$, $\tuple{\T,\T}, k + j \not \models \varphi$. By induction, $\tuple{\H,\T}, k + i \not \models \psi$ and for all $0 \le j < i$, $\tuple{\H,\T}, k + j \not \models \varphi$. Therefore, $\tuple{\H,\T}, k \not \models \varphi \metric{\release}{\last} \psi$: a contradiction.
		
		\item Case $\varphi \metric{\since}{\last} \psi$: if $\tuple{\H,\T}, k \models \varphi \metric{\since}{\last} \psi$ then there exists $0\le i < \lambda$ such that $k-i \ge 0$, $\tuple{\H,\T}, k - i \models \psi$ and for all $0 \le j < i$, $\tuple{\H,\T}, k - j \models \varphi$. By the induction hypothesis, $\tuple{\T,\T}, k - i \models \psi$ and for all $0 \le j < i$, $\tuple{\T,\T}, k - j \models \varphi$. Therefore, $\tuple{\T,\T}, k \models \varphi \metric{\since}{\last} \psi$

		\item Case $\varphi \metric{\trigger}{\last} \psi$: assume by contradiction that $\tuple{\T,\T}, k \not \models \varphi \metric{\trigger}{\last} \psi$. This means that there exists $0\le i < \lambda$ such that $k-i \ge 0$, $\tuple{\T,\T}, k - i \not \models \psi$ and for all $0 \le j < i$, $\tuple{\T,\T}, k - j \not \models \varphi$. By induction hypothesis, $\tuple{\H,\T}, k - i \not \models \psi$ and for all $0 \le j < i$, $\tuple{\H,\T}, k - j \not \models \varphi$, so $\tuple{\H,\T}, k \not \models \varphi \metric{\trigger}{\last} \psi$: a contradiction.
		
	\end{itemize}
	
	For the second item, assume by contradiction that $\tuple{\H,\T},k \models \neg \varphi$  but $\tuple{\T,\T},k \models \varphi$. By persistency,  $\tuple{\T,\T},k \models \neg \varphi$ and, therefore, $\tuple{\T,\T},k \models \bot$: a contradiction.
	Conversely, Assume by contradiction that $\tuple{\H,\T},k \not\models \neg \varphi$. Therefore, either  $\tuple{\H,\T},k \not\models \varphi$ or  $\tuple{\H,\T},k \not\models \varphi$.
	In any case, by persistence, we conclude that $\tuple{\T,\T},k \not\models \varphi$: a contradiction.
	
\end{proofof}

\begin{proofof}{Proposition~\ref{prop:validities}}

For $\varphi \metric{\until}{1} \psi \leftrightarrow \psi$ we reason as follows: from left to right, if $\tuple{\H,\T}, k \models \varphi \metric{\until}{1} \psi$ then $\tuple{\H,\T}, k \models \psi$ by definition. Conversely, if $\tuple{\H,\T}, k + 0 \models \psi$ then $\tuple{\H,\T}, k \models \varphi \metric{\until}{1}\psi$ by definition. 

\noindent The proofs for  $\varphi \metric{\release}{1} \psi \leftrightarrow \psi$,  $\varphi \metric{\since}{1} \psi \leftrightarrow \psi$ and  $\varphi \metric{\trigger}{1} \psi \leftrightarrow \psi$ are done in a similar way.

\noindent From now on we will consider $n$ with $\numeral{n} > 1$. 
\begin{itemize}
	\item For $\varphi \metric{\until}{n} \psi \leftrightarrow \psi \vee \left(\varphi \wedge \next \varphi \metric{\until}{n-1} \psi \right)$, let us consider from left to right that $\tuple{\H,\T}, k \models \varphi \metric{\until}{n} \psi$. This means that there exists $0\le i < \numeral{n}$ such that $k+i < \lambda$, $\tuple{\H,\T}, k + i \models \psi$ and for all $0 \le j < i$, $\tuple{\H,\T}, k + j \models \varphi$. If $i  = 0$ then we can easily conclude $\tuple{\H,\T}, k \models \psi$. If $i > 0$ then $\tuple{\H,\T}, k+1 \models \varphi \metric{\until}{n-1} \psi$ and $\tuple{\H,\T}, k \models \varphi$. As a consequence, $\tuple{\H,\T}, k \models \psi \vee \left( \varphi \wedge \next \left( \varphi \metric{\until}{n-1} \psi\right)\right)$.
	
	For the converse direction, assume that $\tuple{\H,\T}, k \models \psi \vee \left( \varphi \wedge \next \left( \varphi \metric{\until}{n-1} \psi\right)\right)$. If $\tuple{\H,\T}, k \models \psi$ then $\tuple{\H,\T}, k \models \varphi \metric{\until}{n} \psi$ by definition (just take $i=0$).  If $\tuple{\H,\T}, k \models \varphi \wedge \next \left( \varphi \metric{\until}{n-1} \psi\right)$ then $\tuple{\H,\T}, k \models \varphi$, $k + 1 < \lambda$ and  $\tuple{\H,\T}, k + 1 \models  \varphi \metric{\until}{n-1} \psi$. This means that there exists $0 \le i < \numeral{n-1}$  such that $k + 1 + i < \lambda$, $\tuple{\H,\T}, k + 1 + i \models \psi$ and for all $0 \le j < i$, $\tuple{\H,\T}, k + 1 + j \models \varphi$. From this and $\tuple{\H,\T}, k \models \varphi$ we conclude that there exists $0 \le i' < \numeral{n}$ such that $k + i' < \lambda$ and $\tuple{\H,\T}, k + i' \models \psi$ and for all $0 \le j' < i$, $\tuple{\H,\T}, k + j' \models \varphi$. Therefore, $\tuple{\H,\T}, k \models \varphi \metric{\until}{n} \psi$. 

	\item For $\varphi \metric{\release}{n} \psi \leftrightarrow \psi \wedge \left(\varphi \vee \wnext \varphi \metric{\release}{n-1} \psi \right)$, let us consider first the left to right direction. For this, let us assume for the sake of contradiction that $\tuple{\H,\T}, k \not \models \psi \wedge \left( \varphi \vee \wnext \left( \varphi \metric{\release}{n-1} \psi\right)\right)$. If $\tuple{\H,\T}, k \not \models \psi$ then $\tuple{\H,\T}, k \not \models \varphi \metric{\release}{n} \psi$ by definition (just take $i=0$).  If $\tuple{\H,\T}, k \not \models \varphi \wedge  \wnext \left( \varphi \metric{\release}{n-1} \psi\right)$ then $\tuple{\H,\T}, k \not \models \varphi$, $k+1 < \lambda$ and $\tuple{\H,\T}, k + 1 \not \models  \varphi \metric{\release}{n-1} \psi$. This means that there exists $0 \le i < \numeral{n-1}$  such that $k + 1 + i < \lambda$, $\tuple{\H,\T}, k + 1 + i \not \models \psi$ and for all $0 \le j < i$, $\tuple{\H,\T}, k + 1 + j \not \models \varphi$. From this and $\tuple{\H,\T}, k \not \models \varphi$ we conclude that there exists $0 \le i' < \numeral{n}$ such that $k + i' < \lambda$ and $\tuple{\H,\T}, k + i' \not \models \psi$ and for all $0 \le j' < i$, $\tuple{\H,\T}, k + j' \not \models \varphi$. Therefore, $\tuple{\H,\T}, k \not \models \varphi \metric{\release}{n} \psi$: a contradiction. 
	
	From right to left, let us assume by contradiction that $\tuple{\H,\T}, k \not \models \varphi \metric{\release}{n} \psi$. This means that there exists $0\le i < \numeral{n}$ such that $k+i < \lambda$, $\tuple{\H,\T}, k + i \not \models \psi$ and for all $0 \le j < i$, $\tuple{\H,\T}, k + j \not \models \varphi$. If $i  = 0$ then we can easily conclude $\tuple{\H,\T}, k \not \models \psi$. If $i > 0$ then $\tuple{\H,\T}, k+1 \not \models \varphi \metric{\release}{n-1} \psi$ and $\tuple{\H,\T}, k \not \models \varphi$. As a consequence, $\tuple{\H,\T}, k \not \models \psi \wedge \left( \varphi \vee \wnext \left( \varphi \metric{\release}{n-1} \psi\right)\right)$.
	
	\item For $\varphi \metric{\since}{n} \psi \leftrightarrow \psi \vee \left(\varphi \wedge \previous \varphi \metric{\since}{n-1} \psi \right)$, let us assume that $\tuple{\H,\T}, k \models \varphi \metric{\since}{n} \psi$. This means that there exists $0\le i < \numeral{n}$ such that $k-i \ge 0$, $\tuple{\H,\T}, k - i \models \psi$ and for all $0 \le j < i$, $\tuple{\H,\T}, k - j \models \varphi$. If $i  = 0$ then we can easily conclude $\tuple{\H,\T}, k \models \psi$. If $i > 0$ then $\tuple{\H,\T}, k-1 \models \varphi \metric{\since}{n-1} \psi$ and $\tuple{\H,\T}, k \models \varphi$. As a consequence, $\tuple{\H,\T}, k \models \psi \vee \left( \varphi \wedge \previous \left( \varphi \metric{\since}{n-1} \psi\right)\right)$.
	
	For the converse direction, assume that $\tuple{\H,\T}, k \models \psi \vee \left( \varphi \wedge \previous \left( \varphi \metric{\until}{n-1} \psi\right)\right)$. If $\tuple{\H,\T}, k \models \psi$ then $\tuple{\H,\T}, k \models \varphi \metric{\since}{n} \psi$ by definition (just take $i=0$).  If $\tuple{\H,\T}, k \models \varphi \wedge \previous \left( \varphi \metric{\since}{n-1} \psi\right)$ then $\tuple{\H,\T}, k \models \varphi$, $k - 1 \ge 0$ and  $\tuple{\H,\T}, k - 1 \models  \varphi \metric{\since}{n-1} \psi$. This means that there exists $0 \le i < \numeral{n-1}$  such that $k - 1 - i \ge 0$, $\tuple{\H,\T}, k - 1 - i \models \psi$ and for all $0 \le j < i$, $\tuple{\H,\T}, k - 1 - j \models \varphi$. From this and $\tuple{\H,\T}, k \models \varphi$ we conclude that there exists $0 \le i' < \numeral{n}$ such that $k - i' \ge 0$ and $\tuple{\H,\T}, k - i' \models \psi$ and for all $0 \le j' < i$, $\tuple{\H,\T}, k - j' \models \varphi$. Therefore, $\tuple{\H,\T}, k \models \varphi \metric{\since}{n} \psi$. 

	\item For $\varphi \metric{\trigger}{n} \psi \leftrightarrow \psi \wedge \left(\varphi \vee \wprevious \varphi \metric{\trigger}{n-1} \psi \right)$, let us consider first the left to right direction. For this, let us assume for the sake of contradiction that $\tuple{\H,\T}, k \not \models \psi \wedge \left( \varphi \vee \wprevious \left( \varphi \metric{\trigger}{n-1} \psi\right)\right)$. If $\tuple{\H,\T}, k \not \models \psi$ then $\tuple{\H,\T}, k \not \models \varphi \metric{\trigger}{n} \psi$ by definition (just take $i=0$).  If $\tuple{\H,\T}, k \not \models \varphi \wedge  \wprevious \left( \varphi \metric{\trigger}{n-1} \psi\right)$ then $\tuple{\H,\T}, k \not \models \varphi$, $k > 0$ and $\tuple{\H,\T}, k - 1 \not \models  \varphi \metric{\trigger}{n-1} \psi$. This means that there exists $0 \le i < \numeral{n-1}$  such that $k - 1 - i \ge 0$, $\tuple{\H,\T}, k - 1 - i \not \models \psi$ and for all $0 \le j < i$, $\tuple{\H,\T}, k - 1 - j \not \models \varphi$. From this and $\tuple{\H,\T}, k \not \models \varphi$ we conclude that there exists $0 \le i' < \numeral{n}$ such that $k - i' \ge 0$ and $\tuple{\H,\T}, k - i' \not \models \psi$ and for all $0 \le j' < i$, $\tuple{\H,\T}, k - j' \not \models \varphi$. Therefore, $\tuple{\H,\T}, k \not \models \varphi \metric{\trigger}{n} \psi$: a contradiction. 

From right to left, let us assume by contradiction that $\tuple{\H,\T}, k \not \models \varphi \metric{\trigger}{n} \psi$. This means that there exists $0\le i < \numeral{n}$ such that $k-i \ge 0$, $\tuple{\H,\T}, k - i \not \models \psi$ and for all $0 \le j < i$, $\tuple{\H,\T}, k - j \not \models \varphi$. If $i  = 0$ then we can easily conclude $\tuple{\H,\T}, k \not \models \psi$. If $i > 0$ then $\tuple{\H,\T}, k-1 \not \models \varphi \metric{\trigger}{n-1} \psi$ and $\tuple{\H,\T}, k \not \models \varphi$. As a consequence, $\tuple{\H,\T}, k \not \models \psi \wedge \left( \varphi \vee \wprevious \left( \varphi \metric{\trigger}{n-1} \psi\right)\right)$.
\end{itemize}
\end{proofof}

\begin{proofof}{Proposition~\ref{prop:validity:l}}
	We consider the different cases below:
	\begin{itemize}
		\item For $\varphi \metric{\until}{\last} \psi \leftrightarrow \psi \vee \left(\varphi \wedge \next \varphi \metric{\until}{\last} \psi \right)$, let us consider from left to right that $\tuple{\H,\T}, k \models \varphi \metric{\until}{\last} \psi$. This means that there exists $0\le i < \lambda$ such that $k+i < \lambda$, $\tuple{\H,\T}, k + i \models \psi$ and for all $0 \le j < i$, $\tuple{\H,\T}, k + j \models \varphi$. If $i  = 0$ then we can easily conclude $\tuple{\H,\T}, k \models \psi$. If $0 < i < \lambda$ then $\tuple{\H,\T}, k+1 \models \varphi \metric{\until}{\last} \psi$ and $\tuple{\H,\T}, k \models \varphi$. As a consequence, $\tuple{\H,\T}, k \models \psi \vee \left( \varphi \wedge \next \left( \varphi \metric{\until}{\last} \psi\right)\right)$.
		
		For the converse direction, assume that $\tuple{\H,\T}, k \models \psi \vee \left( \varphi \wedge \next \left( \varphi \metric{\until}{\last} \psi\right)\right)$. If $\tuple{\H,\T}, k \models \psi$ then $\tuple{\H,\T}, k \models \varphi \metric{\until}{\last} \psi$ by definition (just take $i=0$).  If $\tuple{\H,\T}, k \models \varphi \wedge \next \left( \varphi \metric{\until}{\last} \psi\right)$ then $\tuple{\H,\T}, k \models \varphi$, $k + 1 < \lambda$ and  $\tuple{\H,\T}, k + 1 \models  \varphi \metric{\until}{\last} \psi$. This means that there exists $0 \le i < \lambda$  such that $k + 1 + i < \lambda$, $\tuple{\H,\T}, k + 1 + i \models \psi$ and for all $0 \le j < i$, $\tuple{\H,\T}, k + 1 + j \models \varphi$. From this and $\tuple{\H,\T}, k \models \varphi$ we conclude that there exists $0 \le i' < \lambda$ such that $k + i' < \lambda$ and $\tuple{\H,\T}, k + i' \models \psi$ and for all $0 \le j' < i$, $\tuple{\H,\T}, k + j' \models \varphi$. Therefore, $\tuple{\H,\T}, k \models \varphi \metric{\until}{\last} \psi$.
		
		\item For $\varphi \metric{\release}{\last} \psi \leftrightarrow \psi \wedge \left(\varphi \vee \wnext \varphi \metric{\release}{\last} \psi \right)$, let us consider first the left to right direction. For this, let us assume for the sake of contradiction that $\tuple{\H,\T}, k \not \models \psi \wedge \left( \varphi \vee \wnext \left( \varphi \metric{\release}{\last} \psi\right)\right)$. If $\tuple{\H,\T}, k \not \models \psi$ then $\tuple{\H,\T}, k \not \models \varphi \metric{\release}{\last} \psi$ by definition (just take $i=0$).  If $\tuple{\H,\T}, k \not \models \varphi \vee  \wnext \left( \varphi \metric{\release}{\last} \psi\right)$ then $\tuple{\H,\T}, k \not \models \varphi$, $k+1 < \lambda$ and $\tuple{\H,\T}, k + 1 \not \models  \varphi \metric{\release}{\last} \psi$. This means that there exists $0 \le i < \lambda$  such that $k + 1 + i < \lambda$, $\tuple{\H,\T}, k + 1 + i \not \models \psi$ and for all $0 \le j < i$, $\tuple{\H,\T}, k + 1 + j \not \models \varphi$. From this and $\tuple{\H,\T}, k \not \models \varphi$ we conclude that there exists $0 \le i' < \lambda$ such that $k + i' < \lambda$ and $\tuple{\H,\T}, k + i' \not \models \psi$ and for all $0 \le j' < i$, $\tuple{\H,\T}, k + j' \not \models \varphi$. Therefore, $\tuple{\H,\T}, k \not \models \varphi \metric{\release}{\last} \psi$: a contradiction.
		
		From right to left, let us assume by contradiction that $\tuple{\H,\T}, k \not \models \varphi \metric{\release}{\last} \psi$. This means that there exists $0\le i < \lambda$ such that $k+i < \lambda$, $\tuple{\H,\T}, k + i \not \models \psi$ and for all $0 \le j < i$, $\tuple{\H,\T}, k + j \not \models \varphi$. If $i  = 0$ then we can easily conclude $\tuple{\H,\T}, k \not \models \psi$. If $i > 0$ then $\tuple{\H,\T}, k+1 \not \models \varphi \metric{\release}{\last} \psi$ and $\tuple{\H,\T}, k \not \models \varphi$. As a consequence, $\tuple{\H,\T}, k \not \models \psi \wedge \left( \varphi \vee \wnext \left( \varphi \metric{\release}{\last} \psi\right)\right)$.
		
		\item For $\varphi \metric{\since}{\last} \psi \leftrightarrow \psi \vee \left(\varphi \wedge \previous \varphi \metric{\since}{\last} \psi \right)$, let us assume that $\tuple{\H,\T}, k \models \varphi \metric{\since}{\last} \psi$. This means that there exists $0\le i < \lambda$ such that $k-i \ge 0$, $\tuple{\H,\T}, k - i \models \psi$ and for all $0 \le j < i$, $\tuple{\H,\T}, k - j \models \varphi$. If $i  = 0$ then we can easily conclude $\tuple{\H,\T}, k \models \psi$. If $i > 0$ then $\tuple{\H,\T}, k-1 \models \varphi \metric{\since}{\last} \psi$ and $\tuple{\H,\T}, k \models \varphi$. As a consequence, $\tuple{\H,\T}, k \models \psi \vee \left( \varphi \wedge \previous \left( \varphi \metric{\since}{\last} \psi\right)\right)$.
		
		For the converse direction, assume that $\tuple{\H,\T}, k \models \psi \vee \left( \varphi \wedge \previous \left( \varphi \metric{\until}{\last} \psi\right)\right)$. If $\tuple{\H,\T}, k \models \psi$ then $\tuple{\H,\T}, k \models \varphi \metric{\since}{\last} \psi$ by definition (just take $i=0$).  If $\tuple{\H,\T}, k \models \varphi \wedge \previous \left( \varphi \metric{\since}{\lambda} \psi\right)$ then $\tuple{\H,\T}, k \models \varphi$, $k - 1 \ge 0$ and  $\tuple{\H,\T}, k - 1 \models  \varphi \metric{\since}{n} \psi$. This means that there exists $0 \le i < \lambda$  such that $k - 1 - i \ge 0$, $\tuple{\H,\T}, k - 1 - i \models \psi$ and for all $0 \le j < i$, $\tuple{\H,\T}, k - 1 - j \models \varphi$. From this and $\tuple{\H,\T}, k \models \varphi$ we conclude that there exists $0 \le i' < \lambda$ such that $k - i' \ge 0$ and $\tuple{\H,\T}, k - i' \models \psi$ and for all $0 \le j' < i$, $\tuple{\H,\T}, k - j' \models \varphi$. Therefore, $\tuple{\H,\T}, k \models \varphi \metric{\since}{\last} \psi$.
		
		\item For $\varphi \metric{\trigger}{\last} \psi \leftrightarrow \psi \wedge \left(\varphi \vee \wprevious \varphi \metric{\trigger}{\lambda} \psi \right)$, let us consider first the left to right direction. For this, let us assume for the sake of contradiction that $\tuple{\H,\T}, k \not \models \psi \wedge \left( \varphi \vee \wprevious \left( \varphi \metric{\trigger}{\lambda} \psi\right)\right)$. If $\tuple{\H,\T}, k \not \models \psi$ then $\tuple{\H,\T}, k \not \models \varphi \metric{\trigger}{\lambda} \psi$ by definition (just take $i=0$).  If $\tuple{\H,\T}, k \not \models \varphi \wedge  \wprevious \left( \varphi \metric{\trigger}{\lambda} \psi\right)$ then $\tuple{\H,\T}, k \not \models \varphi$, $k > 0$ and $\tuple{\H,\T}, k - 1 \not \models  \varphi \metric{\trigger}{\last} \psi$. This means that there exists $0 \le i < \lambda$  such that $k - 1 - i \ge 0$, $\tuple{\H,\T}, k - 1 - i \not \models \psi$ and for all $0 \le j < i$, $\tuple{\H,\T}, k - 1 - j \not \models \varphi$. From this and $\tuple{\H,\T}, k \not \models \varphi$ we conclude that there exists $0 \le i' < \lambda$ such that $k - i' \ge 0$ and $\tuple{\H,\T}, k - i' \not \models \psi$ and for all $0 \le j' < i$, $\tuple{\H,\T}, k - j' \not \models \varphi$. Therefore, $\tuple{\H,\T}, k \not \models \varphi \metric{\trigger}{\last} \psi$: a contradiction.
		
		From right to left, let us assume by contradiction that $\tuple{\H,\T}, k \not \models \varphi \metric{\trigger}{\last} \psi$. This means that there exists $0\le i < \lambda$ such that $k-i \ge 0$, $\tuple{\H,\T}, k - i \not \models \psi$ and for all $0 \le j < i$, $\tuple{\H,\T}, k - j \not \models \varphi$. If $i  = 0$ then we can easily conclude $\tuple{\H,\T}, k \not \models \psi$. If $i > 0$ then $\tuple{\H,\T}, k-1 \not \models \varphi \metric{\trigger}{\last} \psi$ and $\tuple{\H,\T}, k \not \models \varphi$. As a consequence, $\tuple{\H,\T}, k \not \models \psi \wedge \left( \varphi \vee \wprevious \left( \varphi \metric{\trigger}{\last} \psi\right)\right)$.		
	\end{itemize}
\end{proofof}

 \begin{proofof}{Lemma~\ref{lem:translation:correctness}}
The proof is done by using double induction on the complexity of the formula and on $\numeral{n}$. We consider the metric operators below

\begin{itemize}
	\item Case $\varphi \metric{\until}{1} \psi$: from left to right, if $\M, k \models \varphi \metric{\until}{1} \psi$ in \MHT{} then, by Proposition~\ref{prop:validities}, $\M, k \models \psi$ in \MHT{}. By induction hypothesis, $\M, k \models \translation{\psi}$ in \THT{} and, consequently, $\M, k \models \translation{\varphi \metric{\until}{1} \psi}$ in \THT{}. Conversely, if $\M, k \models \translation{\varphi \metric{\until}{1} \psi}$ in \THT{} then, by definition,  $\M, k \models \translation{\psi}$ in \THT{} and, by induction hypothesis, $\M, k \models \psi$ in \MHT{} and, thanks to Proposition~\ref{prop:validities} we conclude $\M, k \models \varphi \metric{\until}{1} \psi$.
	\item Cases $\varphi \metric{\release}{1} \psi$, $\varphi \metric{\since}{1} \psi$ and $\varphi \metric{\trigger}{1} \psi$ are proved in a similar way.

	\item Case  $\varphi \metric{\until}{n} \psi$ with $n > 1$: from left to right, from $\M, k \models \varphi \metric{\until}{n} \psi$ and Proposition~\ref{prop:validities}, $\M, k \models \psi \vee \left(\varphi \wedge \next \left(\varphi \metric{\until}{n-1} \psi\right)\right)$ in \MHT{}. By using induction on the subformulas and on $\numeral{n}$ we can conclude that $\M, k \models \translation{\psi}$ in \THT{} or both $\M, k \models \translation{\varphi}$ and $\M, k \models \next \translation{\varphi \metric{\until}{n-1} \psi}$ in \THT{}. By using the definition of $\translation{~}$ it follows that $\M, k \models \translation{\psi \vee \left(\varphi \wedge \next \left(\varphi \metric{\until}{n-1} \psi\right)\right)}$ and so $\M, k \models \translation{\varphi \metric{\until}{n} \psi}$ in \THT{}. Conversely, if $\M, k \models \translation{\varphi \metric{\until}{n} \psi}$ in \THT{} then, by using the definition of $\translation{~}$, we conclude that $\M,k \models \translation{\psi} \vee \left( \translation{\varphi} \wedge \next \translation{\varphi \metric{\until}{n-1}\psi}\right)$ in \THT{}. By applying induction on $\numeral{n}$ and on the subformulas we can easily conclude that $\M, k \models \psi \vee \left( \varphi \wedge \next \left(\varphi \metric{\until}{n-1}\psi\right)\right)$ in \MHT{}. Thanks to Proposition~\ref{prop:validities} we conclude that $\M, k \models \varphi \metric{\until}{n} \psi$ in \MHT{}.

	\item Case  $\varphi \metric{\release}{n} \psi$ with $\numeral{n} > 1$: from left to right, assume by contradiction that $\M, k \not \models \translation{\varphi \metric{\release}{n} \psi}$ in \THT{}. By definition of $\translation{~}$ we can derive that $\M,k \not\models \translation{\psi} \wedge \left( \translation{\varphi} \vee \wnext \translation{\varphi \metric{\release}{n-1}\psi}\right)$ in \THT{}. By applying induction on $\numeral{n}$ and on the subformulas we can easily conclude that $\M, k \not \models \psi \wedge \left( \varphi \vee \wnext \left(\varphi \metric{\release}{n-1}\psi\right)\right)$ in \MHT{}. Thanks to Proposition~\ref{prop:validities} we conclude that $\M, k \not \models \varphi \metric{\release}{n} \psi$ in \MHT{}: a contradiction. For the converse direction assume, again, by contradiction that $\M, k \not \models \varphi \metric{\release}{n} \psi$ and Proposition~\ref{prop:validities}, $\M, k \not \models \psi \wedge \left(\varphi \vee \wnext \left(\varphi \metric{\release}{n-1} \psi\right)\right)$ in \MHT{}. By using induction on the subformulas and on $\numeral{n}$ we can conclude that $\M, k \not \models \translation{\psi}$ in \THT{} or both $\M, k \not \models \translation{\varphi}$ and $\M, k \not \models \wnext \translation{\varphi \metric{\release}{n-1} \psi}$ in \THT{}. By using the definition of $\translation{~}$ it follows that $\M, k \not \models \translation{\psi \wedge \left(\varphi \vee \wnext \left(\varphi \metric{\release}{n-1} \psi\right)\right)}$ and so $\M, k \not \models \translation{\varphi \metric{\release}{n} \psi}$ in \THT{}: a contradiction.

	\item Case  $\varphi \metric{\since}{n} \psi$ with $\numeral{n} > 1$: from left to right, from $\M, k \models \varphi \metric{\since}{n} \psi$ and Proposition~\ref{prop:validities}, $\M, k \models \psi \vee \left(\varphi \wedge \previous \left(\varphi \metric{\since}{n-1} \psi\right)\right)$ in \MHT{}. By using induction on the subformulas and on $\numeral{n}$ we can conclude that $\M, k \models \translation{\psi}$ in \THT{} or both $\M, k \models \translation{\varphi}$ and $\M, k \models \previous \translation{\varphi \metric{\since}{n-1} \psi}$ in \THT{}. By using the definition of $\translation{~}$ it follows that $\M, k \models \translation{\psi \vee \left(\varphi \wedge \previous \left(\varphi \metric{\since}{n-1} \psi\right)\right)}$ and so $\M, k \models \translation{\varphi \metric{\since}{n} \psi}$ in \THT{}. Conversely, if $\M, k \models \translation{\varphi \metric{\since}{n} \psi}$ in \THT{} then, by using the definition of $\translation{~}$, we conclude that $\M,k \models \translation{\psi} \vee \left( \translation{\varphi} \wedge \previous \translation{\varphi \metric{\since}{n-1}\psi}\right)$ in \THT{}. By applying induction on $\numeral{n}$ and on the subformulas we can easily conclude that $\M, k \models \psi \vee \left( \varphi \wedge \previous \left(\varphi \metric{\since}{n-1}\psi\right)\right)$ in \MHT{}. Thanks to Proposition~\ref{prop:validities} we conclude that $\M, k \models \varphi \metric{\since}{n} \psi$ in \MHT{}.

	\item Case  $\varphi \metric{\trigger}{n} \psi$ with $\numeral{n} > 1$: from left to right, assume by contradiction that $\M, k \not \models \translation{\varphi \metric{\trigger}{n} \psi}$ in \THT{}. By definition of $\translation{~}$ we can derive that $\M,k \not\models \translation{\psi} \wedge \left( \translation{\varphi} \vee \wprevious \translation{\varphi \metric{\trigger}{n-1}\psi}\right)$ in \THT{}. By applying induction on $\numeral{n}$ and on the subformulas we can easily conclude that $\M, k \not \models \psi \wedge \left( \varphi \vee \wprevious \left(\varphi \metric{\trigger}{n-1}\psi\right)\right)$ in \MHT{}. Thanks to Proposition~\ref{prop:validities} we conclude that $\M, k \not \models \varphi \metric{\trigger}{n} \psi$ in \MHT{}: a contradiction. For the converse direction assume, again, by contradiction that $\M, k \not \models \varphi \metric{\trigger}{n} \psi$ and Proposition~\ref{prop:validities}, $\M, k \not \models \psi \wedge \left(\varphi \vee \wprevious \left(\varphi \metric{\trigger}{n-1} \psi\right)\right)$ in \MHT{}. By using induction on the subformulas and on $\numeral{n}$ we can conclude that $\M, k \not \models \translation{\psi}$ in \THT{} or both $\M, k \not \models \translation{\varphi}$ and $\M, k \not \models \wprevious \translation{\varphi \metric{\trigger}{n-1} \psi}$ in \THT{}. By using the definition of $\translation{~}$ it follows that $\M, k \not \models \translation{\psi \wedge \left(\varphi \vee \wprevious \left(\varphi \metric{\trigger}{n-1} \psi\right)\right)}$ and so $\M, k \not \models \translation{\varphi \metric{\trigger}{n} \psi}$ in \THT{}: a contradiction.

	\item Case  $\varphi \metric{\until}{\last} \psi$: from left to right, from $\M, k \models \varphi \metric{\until}{\last} \psi$ and Proposition~\ref{prop:validity:l}, $\M, k \models \psi \vee \left(\varphi \wedge \next \left(\varphi \metric{\until}{\last} \psi\right)\right)$ in \MHT{}. By using induction on the subformulas we can conclude that $\M, k \models \translation{\psi}$ in \THT{} or both $\M, k \models \translation{\varphi}$ and $\M, k \models \next \translation{\varphi \metric{\until}{\last} \psi}$ in \THT{}. By using the definition of $\translation{~}$ it follows that $\M, k \models \translation{\psi \vee \left(\varphi \wedge \next \left(\varphi \metric{\until}{\last} \psi\right)\right)}$ and so $\M, k \models \translation{\varphi \metric{\until}{\last} \psi}$ in \THT{}. Conversely, if $\M, k \models \translation{\varphi \metric{\until}{\last} \psi}$ in \THT{} then, by using the definition of $\translation{~}$, we conclude that $\M,k \models \translation{\psi} \vee \left( \translation{\varphi} \wedge \next \translation{\varphi \metric{\until}{\last}\psi}\right)$ in \THT{}. By applying induction on the subformulas we can easily conclude that $\M, k \models \psi \vee \left( \varphi \wedge \next \left(\varphi \metric{\until}{\last}\psi\right)\right)$ in \MHT{}. Thanks to Proposition~\ref{prop:validity:l} we conclude that $\M, k \models \varphi \metric{\until}{\last} \psi$ in \MHT{}.

	\item Case  $\varphi \metric{\release}{\last} \psi$:  from left to right, assume by contradiction that $\M, k \not \models \translation{\varphi \metric{\release}{\last} \psi}$ in \THT{}. By definition of $\translation{~}$ we can derive that $\M,k \not\models \translation{\psi} \wedge \left( \translation{\varphi} \vee \wnext \translation{\varphi \metric{\release}{\last}\psi}\right)$ in \THT{}. By applying induction on the subformulas we can easily conclude that $\M, k \not \models \psi \wedge \left( \varphi \vee \wnext \left(\varphi \metric{\release}{\last}\psi\right)\right)$ in \MHT{}. Thanks to Proposition~\ref{prop:validity:l} we conclude that $\M, k \not \models \varphi \metric{\release}{\last} \psi$ in \MHT{}: a contradiction. For the converse direction assume, again, by contradiction that $\M, k \not \models \varphi \metric{\release}{\last} \psi$ and Proposition~\ref{prop:validity:l}, $\M, k \not \models \psi \wedge \left(\varphi \vee \wnext \left(\varphi \metric{\release}{\last} \psi\right)\right)$ in \MHT{}. By using induction on the subformulas we can conclude that $\M, k \not \models \translation{\psi}$ in \THT{} or both $\M, k \not \models \translation{\varphi}$ and $\M, k \not \models \wnext \translation{\varphi \metric{\release}{\last} \psi}$ in \THT{}. By using the definition of $\translation{~}$ it follows that $\M, k \not \models \translation{\psi \wedge \left(\varphi \vee \wnext \left(\varphi \metric{\release}{\last} \psi\right)\right)}$ and so $\M, k \not \models \translation{\varphi \metric{\release}{\last} \psi}$ in \THT{}: a contradiction.

	\item Case  $\varphi \metric{\since}{\last} \psi$: from left to right, from $\M, k \models \varphi \metric{\since}{\last} \psi$ and Proposition~\ref{prop:validity:l}, $\M, k \models \psi \vee \left(\varphi \wedge \previous \left(\varphi \metric{\since}{\last} \psi\right)\right)$ in \MHT{}. By using induction on the subformulas we can conclude that $\M, k \models \translation{\psi}$ in \THT{} or both $\M, k \models \translation{\varphi}$ and $\M, k \models \previous \translation{\varphi \metric{\since}{\last} \psi}$ in \THT{}. By using the definition of $\translation{~}$ it follows that $\M, k \models \translation{\psi \vee \left(\varphi \wedge \previous \left(\varphi \metric{\since}{\last} \psi\right)\right)}$ and so $\M, k \models \translation{\varphi \metric{\since}{\last} \psi}$ in \THT{}. Conversely, if $\M, k \models \translation{\varphi \metric{\since}{\last} \psi}$ in \THT{} then, by using the definition of $\translation{~}$, we conclude that $\M,k \models \translation{\psi} \vee \left( \translation{\varphi} \wedge \previous \translation{\varphi \metric{\since}{\last}\psi}\right)$ in \THT{}. By applying induction on the subformulas we can easily conclude that $\M, k \models \psi \vee \left( \varphi \wedge \previous \left(\varphi \metric{\since}{\last} \psi\right)\right)$ in \MHT{}. Thanks to Proposition~\ref{prop:validity:l} we conclude that $\M, k \models \varphi \metric{\since}{\last} \psi$ in \MHT{}.

	\item Case  $\varphi \metric{\trigger}{\last} \psi$: from left to right, assume by contradiction that $\M, k \not \models \translation{\varphi \metric{\trigger}{\last} \psi}$ in \THT{}. By definition of $\translation{~}$ we can derive that $\M,k \not\models \translation{\psi} \wedge \left( \translation{\varphi} \vee \wprevious \translation{\varphi \metric{\trigger}{\last}\psi}\right)$ in \THT{}. By applying induction on the subformulas we can easily conclude that $\M, k \not \models \psi \wedge \left( \varphi \vee \wprevious \left(\varphi \metric{\trigger}{\last}\psi\right)\right)$ in \MHT{}. Thanks to Proposition~\ref{prop:validity:l} we conclude that $\M, k \not \models \varphi \metric{\trigger}{\last} \psi$ in \MHT{}: a contradiction. For the converse direction assume, again, by contradiction that $\M, k \not \models \varphi \metric{\trigger}{\last} \psi$ and Proposition~\ref{prop:validity:l}, $\M, k \not \models \psi \wedge \left(\varphi \vee \wprevious \left(\varphi \metric{\trigger}{\last} \psi\right)\right)$ in \MHT{}. By using induction on the subformulas we can conclude that $\M, k \not \models \translation{\psi}$ in \THT{} or both $\M, k \not \models \translation{\varphi}$ and $\M, k \not \models \wprevious \translation{\varphi \metric{\trigger}{\last} \psi}$ in \THT{}. By using the definition of $\translation{~}$ it follows that $\M, k \not \models \translation{\psi \wedge \left(\varphi \vee \wprevious \left(\varphi \metric{\trigger}{\last} \psi\right)\right)}$ and so $\M, k \not \models \translation{\varphi \metric{\trigger}{\last} \psi}$ in \THT{}: a contradiction.
\end{itemize}

The proof is done by structural induction. For each case, the proof can be done by using the equivalences of Proposition~\ref{prop:validities} and Proposition~\ref{prop:validity:l} together with some propositional reasoning.
\end{proofof}
 
\begin{proofof}{Proposition~\ref{prop:three-valued}}
The proof is by structural induction. We consider only until/release and since/trigger.
\begin{itemize}
	\item $\varphi \metric{\until}{n} \psi$: from left to right assume that $\tuple{\H,\T}, k \models\varphi \metric{\until}{n} \psi$. Therefore, there exists $\rangeo{i}{0}{\numeral{n}}$ such that $k + i < \lambda$, $\tuple{\H,\T}, k+i \models\psi$ and for all $\rangeo{j}{0}{i}$, $\tuple{\H,\T}, k + j \models\varphi$. By induction hypothesis, $\trival{k+i}{\psi} = 2$ and $\trival{k+j}{\varphi} = 2$, for all $\rangeo{j}{0}{i}$. Therefore, $\min\lbrace\trival{k+i}{\psi}, \trival{k+j}{\varphi}\mid \rangeo{j}{0}{i},\ k + i < \lambda \rbrace = 2$ and so
	$\max\lbrace\min\lbrace\trival{k+i}{\psi}, \trival{k+j}{\varphi}\mid \rangeo{j}{0}{i},\ k + i < \lambda \rbrace \mid  \rangeo{i}{0}{\numeral{n}} \rbrace= 2$. Conversely, assume that 
	$\trival{k}{\varphi \metric{\until}{n}\psi} = \max\lbrace\min\lbrace\trival{k+i}{\psi}, \trival{k+j}{\varphi}\mid \rangeo{j}{0}{i},\ k + i < \lambda \rbrace \mid  \rangeo{i}{0}{\numeral{n}} \rbrace= 2$. Therefore there exists $0 \le i < \numeral{n}$ such that $\min\lbrace\trival{k+i}{\psi}, \trival{k+j}{\varphi}\mid \rangeo{j}{0}{i},\ k + i < \lambda \rbrace = 2$. As a consequence, $k + i < \lambda$, $\trival{k+i}{\psi} = 2$ and $\trival{k+j}{\varphi} = 2$, for all $0 \le j < i$. By induction induction it follows that $\tuple{\H, \T}, k+i \models \psi$ and $\tuple{\H, \T}, k+j \models \varphi$, for all $0 \le j < i$. This means that $\tuple{\H, \T}, k \models \varphi \metric{\until}{n}\psi$. The second item is proved in a similar way.
	
	\item $\varphi \metric{\release}{n} \psi$: from left to right, assume that $\trival{k}{\varphi \metric{\release}{n}\psi} = \min\lbrace\max\lbrace\trival{k+i}{\psi}, \trival{k+j}{\varphi}\mid \rangeo{j}{0}{i},\ k + i < \lambda \rbrace \mid  \rangeo{i}{0}{\numeral{n}} \rbrace\not = 2$. therefore, there exists $0 \le i < \numeral{n}$ such that $\max\lbrace\trival{k+i}{\psi}, \trival{k+j}{\varphi}\mid \rangeo{j}{0}{i},\ k + i < \lambda \rbrace \not =  2$. This means that $k + i < \lambda$, $\trival{k+i}{\psi} \not = 2$ and $\trival{k+j}{\varphi} \not =2$ for all $0 \le j < i$. By induction hypothesis, $\tuple{\H, \T}, k+i \not\models \psi$ and $\tuple{\H, \T}, k+j \not\models \varphi$, for all $0 \le j < i$. From this we conclude that $\tuple{\H, \T}, k \not \models \varphi \metric{\release}{n}\psi$. Conversely, assume that $\tuple{\H,\T}, k\not \models \varphi\metric{\release}{n}\psi$. This means that there exists $\rangeo{i}{0}{\numeral{n}}$ such that $k + i < \lambda$, $\tuple{\H,\T}, k+i \not \models\psi$ and for all $\rangeo{j}{0}{i}$, $\tuple{\H,\T}, k + j \not\models\varphi$. 
	By induction hypothesis, $\trival{k+i}{\psi} \not= 2$ and $\trival{k+j}{\varphi} \not= 2$, for all $\rangeo{j}{0}{i}$. Therefore, $\max\lbrace\trival{k+i}{\psi}, \trival{k+j}{\varphi}\mid \rangeo{j}{0}{i},\ k + i < \lambda \rbrace \not = 2$ and so $\min\lbrace\max\lbrace\trival{k+i}{\psi}, \trival{k+j}{\varphi}\mid \rangeo{j}{0}{i},\ k + i < \lambda \rbrace \mid  \rangeo{i}{0}{\numeral{n}} \rbrace= 2$: a contradiction. The second item is proved in a similar way.
	\item $\varphi \metric{\since}{n} \psi$ From left to right, assume that $\tuple{\H,\T}, k \models\varphi \metric{\since}{n} \psi$. Therefore, there exists $\rangeo{i}{0}{\numeral{n}}$ such that $k-i \ge 0$, $\tuple{\H,\T}, k-i \models\psi$ and for all $\rangeo{j}{0}{i}$, $\tuple{\H,\T}, k - j \models\varphi$. By induction hypothesis, $\trival{k-i}{\psi} = 2$ and $\trival{k-j}{\varphi} = 2$, for all $\rangeo{j}{0}{i}$. Therefore, $\min\lbrace\trival{k-i}{\psi}, \trival{k-j}{\varphi}\mid \rangeo{j}{0}{i},\ k - i \ge 0 \rbrace = 2$ and so $\max\lbrace\min\lbrace\trival{k-i}{\psi}, \trival{k-j}{\varphi}\mid \rangeo{j}{0}{i},\ k - i\ge 0  \rbrace \mid  \rangeo{i}{0}{\numeral{n}} \rbrace= 2$. Conversely, assume that $\trival{k}{\varphi \metric{\since}{n}\psi} = \max\lbrace\min\lbrace\trival{k-i}{\psi}, \trival{k-j}{\varphi}\mid \rangeo{j}{0}{i},\ k - i \ge 0 \rbrace \mid  \rangeo{i}{0}{\numeral{n}} \rbrace= 2$. Therefore there exists $0 \le i < \numeral{n}$ such that $\min\lbrace\trival{k-i}{\psi}, \trival{k-j}{\varphi}\mid \rangeo{j}{0}{i},\ k-i \ge 0 \rbrace = 2$. As a consequence, $k - i \ge 0$, $\trival{k-i}{\psi} = 2$ and $\trival{k-j}{\varphi} = 2$, for all $0 \le j < i$. By induction induction it follows that $\tuple{\H, \T}, k-i \models \psi$ and $\tuple{\H, \T}, k-j \models \varphi$, for all $0 \le j < i$. This means that $\tuple{\H, \T}, k \models \varphi \metric{\since}{n}\psi$. The second item is proved in a similar way.
	
	\item $\varphi \metric{\trigger}{n} \psi$: from left to right, assume that $\trival{k}{\varphi \metric{\trigger}{n}\psi} = \min\lbrace\max\lbrace\trival{k-i}{\psi}, \trival{k-j}{\varphi}\mid \rangeo{j}{0}{i},\ k - i \ge 0 \rbrace \mid  \rangeo{i}{0}{\numeral{n}} \rbrace\not = 2$. therefore, there exists $0 \le i < \numeral{n}$ such that $\max\lbrace\trival{k-i}{\psi}, \trival{k-j}{\varphi}\mid \rangeo{j}{0}{i},\ k - i \ge 0 \rbrace \not =  2$. This means that $k - i \ge 0$, $\trival{k-i}{\psi} \not = 2$ and $\trival{k-j}{\varphi} \not =2$ for all $0 \le j < i$. By induction hypothesis, $\tuple{\H, \T}, k-i \not\models \psi$ and $\tuple{\H, \T}, k-j \not\models \varphi$, for all $0 \le j < i$. From this we conclude that $\tuple{\H, \T}, k \not \models \varphi \metric{\trigger}{n}\psi$: a contradiction. Conversely, assume that $\tuple{\H,\T}, k\not \models \varphi\metric{\trigger}{n}\psi$. This means that there exists $\rangeo{i}{0}{\numeral{n}}$ such that $k - i \ge 0$, $\tuple{\H,\T}, k-i \not \models\psi$ and for all $\rangeo{j}{0}{i}$, $\tuple{\H,\T}, k - j \not\models\varphi$. By induction hypothesis, $\trival{k-i}{\psi} \not= 2$ and $\trival{k-j}{\varphi} \not= 2$, for all $\rangeo{j}{0}{i}$. Therefore, $\max\lbrace\trival{k-i}{\psi}, \trival{k-j}{\varphi}\mid \rangeo{j}{0}{i},\ k-i\ge 0 \rbrace \not = 2$ and so $\min\lbrace\max\lbrace\trival{k-i}{\psi}, \trival{k-j}{\varphi}\mid \rangeo{j}{0}{i},\ k - i \ge 0\rbrace \mid  \rangeo{i}{0}{\numeral{n}} \rbrace= 2$. The second item is proved in a similar way: a contradiction. The second item is proved in a similar way.

	\item $\varphi \metric{\until}{\last} \psi$: from left to right assume that $\tuple{\H,\T}, k \models\varphi \metric{\until}{\last} \psi$. Therefore, there exists $\rangeo{i}{0}{\lambda}$ such that $k + i < \lambda$, $\tuple{\H,\T}, k+i \models\psi$ and for all $\rangeo{j}{0}{i}$, $\tuple{\H,\T}, k + j \models\varphi$. By induction hypothesis, $\trival{k+i}{\psi} = 2$ and $\trival{k+j}{\varphi} = 2$, for all $\rangeo{j}{0}{i}$. Therefore, $\min\lbrace\trival{k+i}{\psi}, \trival{k+j}{\varphi}\mid \rangeo{j}{0}{i},\ k + i < \lambda \rbrace = 2$ and so
	$\max\lbrace\min\lbrace\trival{k+i}{\psi}, \trival{k+j}{\varphi}\mid \rangeo{j}{0}{i},\ k + i < \lambda \rbrace \mid  \rangeo{i}{0}{\lambda} \rbrace= 2$. Conversely, assume that 
	$\trival{k}{\varphi \metric{\until}{\last}\psi} = \max\lbrace\min\lbrace\trival{k+i}{\psi}, \trival{k+j}{\varphi}\mid \rangeo{j}{0}{i},\ k + i < \lambda \rbrace \mid  \rangeo{i}{0}{\lambda} \rbrace= 2$. Therefore there exists $0 \le i < \lambda$ such that $\min\lbrace\trival{k+i}{\psi}, \trival{k+j}{\varphi}\mid \rangeo{j}{0}{i},\ k + i < \lambda \rbrace = 2$. As a consequence, $k + i < \lambda$, $\trival{k+i}{\psi} = 2$ and $\trival{k+j}{\varphi} = 2$, for all $0 \le j < i$. By induction induction it follows that $\tuple{\H, \T}, k+i \models \psi$ and $\tuple{\H, \T}, k+j \models \varphi$, for all $0 \le j < i$. This means that $\tuple{\H, \T}, k \models \varphi \metric{\until}{\last}\psi$. The second item is proved in a similar way.
	
	\item $\varphi \metric{\release}{\last} \psi$: from left to right, assume that $\trival{k}{\varphi \metric{\release}{\last}\psi} = \min\lbrace\max\lbrace\trival{k+i}{\psi}, \trival{k+j}{\varphi}\mid \rangeo{j}{0}{i},\ k + i < \lambda \rbrace \mid  \rangeo{i}{0}{\lambda} \rbrace\not = 2$. therefore, there exists $0 \le i < \lambda$ such that $\max\lbrace\trival{k+i}{\psi}, \trival{k+j}{\varphi}\mid \rangeo{j}{0}{i},\ k + i < \lambda \rbrace \not =  2$. This means that $k + i < \lambda$, $\trival{k+i}{\psi} \not = 2$ and $\trival{k+j}{\varphi} \not =2$ for all $0 \le j < i$. By induction hypothesis, $\tuple{\H, \T}, k+i \not\models \psi$ and $\tuple{\H, \T}, k+j \not\models \varphi$, for all $0 \le j < i$. From this we conclude that $\tuple{\H, \T}, k \not \models \varphi \metric{\release}{\last}\psi$. Conversely, assume that $\tuple{\H,\T}, k\not \models \varphi\metric{\release}{\last}\psi$. This means that there exists $\rangeo{i}{0}{\lambda}$ such that $k + i < \lambda$, $\tuple{\H,\T}, k+i \not \models\psi$ and for all $\rangeo{j}{0}{i}$, $\tuple{\H,\T}, k + j \not\models\varphi$. 
	By induction hypothesis, $\trival{k+i}{\psi} \not= 2$ and $\trival{k+j}{\varphi} \not= 2$, for all $\rangeo{j}{0}{i}$. Therefore, $\max\lbrace\trival{k+i}{\psi}, \trival{k+j}{\varphi}\mid \rangeo{j}{0}{i},\ k + i < \lambda \rbrace \not = 2$ and so $\min\lbrace\max\lbrace\trival{k+i}{\psi}, \trival{k+j}{\varphi}\mid \rangeo{j}{0}{i},\ k + i < \lambda \rbrace \mid  \rangeo{i}{0}{\lambda} \rbrace= 2$: a contradiction. The second item is proved in a similar way.
	\item $\varphi \metric{\since}{\last} \psi$ From left to right, assume that $\tuple{\H,\T}, k \models\varphi \metric{\since}{\last} \psi$. Therefore, there exists $\rangeo{i}{0}{\lambda}$ such that $k-i \ge 0$, $\tuple{\H,\T}, k-i \models\psi$ and for all $\rangeo{j}{0}{i}$, $\tuple{\H,\T}, k - j \models\varphi$. By induction hypothesis, $\trival{k-i}{\psi} = 2$ and $\trival{k-j}{\varphi} = 2$, for all $\rangeo{j}{0}{i}$. Therefore, $\min\lbrace\trival{k-i}{\psi}, \trival{k-j}{\varphi}\mid \rangeo{j}{0}{i},\ k - i \ge 0 \rbrace = 2$ and so $\max\lbrace\min\lbrace\trival{k-i}{\psi}, \trival{k-j}{\varphi}\mid \rangeo{j}{0}{i},\ k - i\ge 0  \rbrace \mid  \rangeo{i}{0}{\lambda} \rbrace= 2$. Conversely, assume that $\trival{k}{\varphi \metric{\since}{\last}\psi} = \max\lbrace\min\lbrace\trival{k-i}{\psi}, \trival{k-j}{\varphi}\mid \rangeo{j}{0}{i},\ k - i \ge 0 \rbrace \mid  \rangeo{i}{0}{\lambda} \rbrace= 2$. Therefore there exists $0 \le i < \lambda$ such that $\min\lbrace\trival{k-i}{\psi}, \trival{k-j}{\varphi}\mid \rangeo{j}{0}{i},\ k-i \ge 0 \rbrace = 2$. As a consequence, $k - i \ge 0$, $\trival{k-i}{\psi} = 2$ and $\trival{k-j}{\varphi} = 2$, for all $0 \le j < i$. By induction it follows that $\tuple{\H, \T}, k-i \models \psi$ and $\tuple{\H, \T}, k-j \models \varphi$, for all $0 \le j < i$. This means that $\tuple{\H, \T}, k \models \varphi \metric{\since}{\last}\psi$. The second item is proved in a similar way.
	
	\item $\varphi \metric{\trigger}{\last} \psi$: from left to right, assume that $\trival{k}{\varphi \metric{\trigger}{\last}\psi} = \min\lbrace\max\lbrace\trival{k-i}{\psi}, \trival{k-j}{\varphi}\mid \rangeo{j}{0}{i},\ k - i \ge 0 \rbrace \mid  \rangeo{i}{0}{\lambda} \rbrace\not = 2$. therefore, there exists $0 \le i < \lambda$ such that $\max\lbrace\trival{k-i}{\psi}, \trival{k-j}{\varphi}\mid \rangeo{j}{0}{i},\ k - i \ge 0 \rbrace \not =  2$. This means that $k - i \ge 0$, $\trival{k-i}{\psi} \not = 2$ and $\trival{k-j}{\varphi} \not =2$ for all $0 \le j < i$. By induction hypothesis, $\tuple{\H, \T}, k-i \not\models \psi$ and $\tuple{\H, \T}, k-j \not\models \varphi$, for all $0 \le j < i$. From this we conclude that $\tuple{\H, \T}, k \not \models \varphi \metric{\trigger}{\last}\psi$: a contradiction. Conversely, assume that $\tuple{\H,\T}, k\not \models \varphi\metric{\trigger}{\last}\psi$. This means that there exists $\rangeo{i}{0}{\lambda}$ such that $k - i \ge 0$, $\tuple{\H,\T}, k-i \not \models\psi$ and for all $\rangeo{j}{0}{i}$, $\tuple{\H,\T}, k - j \not\models\varphi$. By induction hypothesis, $\trival{k-i}{\psi} \not= 2$ and $\trival{k-j}{\varphi} \not= 2$, for all $\rangeo{j}{0}{i}$. Therefore, $\max\lbrace\trival{k-i}{\psi}, \trival{k-j}{\varphi}\mid \rangeo{j}{0}{i},\ k-i\ge 0 \rbrace \not = 2$ and so $\min\lbrace\max\lbrace\trival{k-i}{\psi}, \trival{k-j}{\varphi}\mid \rangeo{j}{0}{i},\ k - i \ge 0\rbrace \mid  \rangeo{i}{0}{\lambda} \rbrace= 2$. The second item is proved in a similar way: a contradiction. The second item is proved in a similar way.
\end{itemize}
	
\end{proofof} 
\begin{proofof}{Proposition~\ref{prop:closure:finite}}
We assume $k_\varphi$ is defined for any metric formula $\varphi$ as stated in the proposition.
	We proceed by structural induction.
	\begin{itemize}
		\item case $\varphi = p$: in this case $\lvert p \rvert = 1$ and $\cl(p) = \lbrace p \rbrace$, so $\lvert \cl(p)\rvert = 1$. 
		Since there is no metric connective in $p$, $k_p = 1$. Therefore, $\lvert \cl(p)\rvert = 1 \le 2*1 * \lvert p \rvert = 2$.
		\item case $\varphi \wedge \psi$: in this case, $\cl(\varphi \wedge \psi) = \lbrace \varphi \wedge \psi \rbrace \cup \cl(\varphi) \cup \cl(\psi)$ and $\lvert\cl(\varphi \wedge \psi) \rvert \le 1 + \lvert \cl(\varphi) \rvert + \lvert \cl(\psi)\rvert$.
		We assume, without loss of generality, $\max(k_\varphi,k_\psi) = k_\psi$ when needed. 
		We prove this case next:
		
		\begin{align*}
		\lvert \cl(\varphi \wedge \psi)\rvert& \le 1 + \lvert \cl(\varphi) \rvert + \lvert \cl(\psi)\rvert &  \\
											 & \le 1 + 2 k_\varphi \lvert \varphi\rvert + 2 k_\psi\lvert \psi\rvert & \hbox{by induction on }\varphi \hbox{ and } \psi\\
											 & \le 1 + 2 k_\varphi \lvert \varphi\rvert + 2 k_\psi\lvert \psi\rvert  + 2 (k_\psi - k_\varphi) \lvert \varphi \rvert & 2 (k_\psi - k_\varphi) \lvert \varphi \rvert \ge 0\\ 
											 & \le 1 + 2 k_\varphi \lvert \varphi\rvert + 2 k_\psi\lvert \psi\rvert + 2 k_\psi \lvert \varphi \rvert - 2 k_\varphi	\lvert \varphi \rvert \\	
											 & \le 1 + 2 k_\psi\lvert \psi\rvert + 2 k_\psi \lvert \varphi \rvert \\
											 & \le 1 + 2 k_\psi \left( \lvert \psi\rvert + \lvert \varphi \rvert\right) \\
											 & \le 1 + 2 k_\psi \left( \lvert \varphi \wedge \psi\rvert  - 1 \right) \\
											 & \le 1 - 2 k_\psi +  2 k_\psi \lvert \varphi \wedge \psi\rvert\\
											 & \le 2 k_\psi \lvert \varphi \wedge \psi\rvert. & 1 - 2 k_\psi < 0.
		\end{align*}	

		\item the proof for the formulas $\varphi \vee \psi$ and $\varphi \rightarrow \psi$ is done as for $\varphi \wedge \psi$.
		\item case $\next \varphi$: in this case, $\cl(\next \varphi) = \lbrace \next \varphi \rbrace \cup \cl(\varphi)$ and $\lvert\next \varphi\rvert = 1 + \lvert \varphi \rvert$.
		It follows that

		\begin{align*}
			\lvert \cl(\next \varphi)\rvert & \le 1 + \lvert \cl(\varphi) \rvert\\
											& \le 1 + 2 k_\varphi \lvert \varphi \rvert & \hbox{by induction on }\varphi\\
											& \le 1 + 2 k_\varphi \left( \lvert \next \varphi \rvert -1 \right)\\	
											& \le 1 - 2 k_\varphi + 2 k_\varphi \lvert \next \varphi \rvert\\
											& \le  2 k_\varphi \lvert \next \varphi \rvert & 1 - 2 k_\varphi < 0\\																			
		\end{align*}

		\item  The proof for $\wnext\varphi$, $\previous \varphi$ and $\wprevious \varphi$ follows the same line of reasoning as for $\next \varphi$.

		\item Case $\varphi \metric{\until}{\last} \psi$: in this case, $\cl(\varphi \metric{\until}{\last} \psi) = \lbrace \varphi \metric{\until}{\last} \psi, \next \left( \varphi \metric{\until}{\last} \psi \right) \rbrace \cup \cl(\varphi) \cup \cl(\psi)$ and  $\lvert\varphi \metric{\until}{\last} \psi \rvert = 1 + \lvert \varphi \rvert + \lvert \psi \rvert$.
		We proceed in a similar way as in the previous cases. We will also assume, without loss of generality, that $\max(k_\varphi,k_\psi) = k_\psi$ when needed.

		\begin{align*} 
			\lvert \cl(\varphi \metric{\until}{\last} \psi)\rvert& \le 2 + \lvert \cl(\varphi) \rvert + \lvert \cl(\psi)\rvert \\
												 & \le 2 + 2k_\varphi \lvert \varphi\rvert +2 k_\psi\lvert \psi\rvert & \hbox{by induction on }\varphi \hbox{ and } \psi \\
												 & \le 2 + 2 k_\varphi \lvert \varphi\rvert +2 k_\psi\lvert \psi\rvert  + 2 (k_\psi - k_\varphi) \lvert \varphi \rvert & 2 (k_\psi - k_\varphi) \lvert \varphi \rvert \ge 0\\ 
												 & \le 2 +2 k_\varphi \lvert \varphi\rvert +2 k_\psi\lvert \psi\rvert +2 k_\psi \lvert \varphi \rvert - 2k_\varphi	\lvert \varphi \rvert \\	
												 & \le 2 + 2 k_\psi\lvert \psi\rvert + 2 k_\psi \lvert \varphi \rvert \\
												 & \le 2 + 2 k_\psi \left( \lvert \psi\rvert + \lvert \varphi \rvert\right) \\
												 & \le 2 + 2 k_\psi \left( \lvert \varphi \metric{\until}{\last} \psi\rvert  - 1 \right) \\
												 & \le 2 - 2 k_\psi +  2 k_\psi \lvert \varphi \metric{\until}{\last} \psi \rvert\\
												 &\le 2 k_\psi \lvert \varphi \metric{\until}{\last} \psi \rvert. & 2 - 2 k_\psi \le 0
		\end{align*}	

		\item The proof for $\varphi \metric{\release}{\last} \psi$, $\varphi \metric{\since}{\last} \psi$ and $\varphi \metric{\trigger}{\last} \psi$ follows the same reasoning as for the case $\varphi \metric{\until}{\last} \psi$.
		
		\item Case $\varphi \metric{\until}{n} \psi$: in this case we have that $\lvert\varphi \metric{\until}{n} \psi \rvert = 1 + \lvert \varphi \rvert + \lvert \psi \rvert$ and $\cl(\varphi \metric{\until}{n} \psi) = \lbrace \varphi \metric{\until}{i} \psi \mid 1 \le i \le n\rbrace \cup \lbrace \next \left(\varphi \metric{\until}{i} \psi\right) \mid 1 \le i < n \rbrace \cup \cl(\varphi) \cup \cl(\psi)$. Therefore, $$\lvert\cl(\varphi \metric{\until}{n} \psi) \rvert \le 2n -1 + \lvert \cl(\varphi) \rvert + \lvert \cl(\psi)\rvert.$$ 
We consider two different cases:
		\begin{enumerate}
			\item $n = \max\lbrace n, k_\varphi,k_\psi \rbrace$. Consequently, $2(n - k_\varphi) \lvert \varphi \rvert \ge 0$ and $2(n - k_\psi) \lvert \psi \rvert\ge 0$:
\begin{align*} 
				\lvert \cl(\varphi \metric{\until}{n} \psi)\rvert & \le 2n -1 + \lvert \cl(\varphi) \rvert + \lvert \cl(\psi)\rvert \\
													  & \le 2n - 1 + 2k_\varphi \lvert \varphi\rvert + 2k_\psi\lvert \psi\rvert \hspace{50pt} \hbox{by induction on }\varphi \hbox{ and } \psi \\
													  & \le 2n -1 + 2k_\varphi \lvert \varphi\rvert + 2k_\psi\lvert \psi\rvert  + 2(n - k_\varphi) \lvert \varphi \rvert + 2(n - k_\psi) \lvert \psi \rvert\\ 
													  & \le 2 n -1 + 2k_\varphi \lvert \varphi\rvert + 2k_\psi\lvert \psi\rvert  + 2n \lvert \varphi \rvert - 2 k_\varphi \lvert \varphi \rvert + 2n \lvert \psi \rvert - 2 k_\psi \lvert \psi \rvert\\ 
													  & \le 2n -1 + 2n \lvert \varphi \rvert + 2n \lvert \psi \rvert\\ 
													  & \le 2n -1 + 2n \left(\lvert \varphi \rvert + \lvert \psi \rvert\right)\\
													  & \le 2n -1 + 2n \left(\lvert \varphi \metric{\until}{n} \psi \rvert - 1\right)\\ 	
													  & \le 2n -1 -2n  + 2n \lvert \varphi \metric{\until}{n} \psi \rvert\\ 	
													  & \le -1 + 2n \lvert \varphi \metric{\until}{n} \psi \rvert\\ 
													  & \le 2n \lvert \varphi \metric{\until}{n} \psi \rvert.
			\end{align*}	
					
			\item otherwise, assume that $k_\psi  = \max\lbrace k_\varphi,k_\psi \rbrace$. Therefore, $2(k_\psi - k_\varphi) \lvert \varphi \rvert \ge 0$ :

			\begin{align*} 
						\lvert \cl(\varphi \metric{\until}{n} \psi)\rvert & \le 2n -1 + \lvert \cl(\varphi) \rvert + \lvert \cl(\psi)\rvert \\
						& \le 2 n -1 + 2k_\varphi \lvert \varphi\rvert + 2k_\psi\lvert \psi\rvert \hspace{50pt} \hbox{by induction on }\varphi \hbox{ and } \psi \\
					    & \le 2n -1 + 2k_\varphi \lvert \varphi\rvert + 2k_\psi\lvert \psi\rvert  + 2(k_\psi - k_\varphi) \lvert \varphi \rvert \\ 
						& \le 2n -1 + 2k_\varphi \lvert \varphi\rvert + 2k_\psi\lvert \psi\rvert  + 2k_\psi \lvert \varphi \rvert - 2 k_\varphi \lvert \varphi \rvert\\ 
						& \le 2n -1 + 2k_\psi \lvert \varphi \rvert + 2k_\psi \lvert \psi \rvert\\ 
						& \le 2n -1 + 2k_\psi \left(\lvert \varphi \rvert + \lvert \psi \rvert\right)\\
						& \le 2n -1 + 2k_\psi \left(\lvert \varphi \metric{\until}{n} \psi \rvert - 1\right)\\ 	
						& \le 2n -1 -2k_\psi  + 2k_\psi \lvert \varphi \metric{\until}{n} \psi \rvert \\
						& \le 2k_\psi \lvert \varphi \metric{\until}{n} \psi \rvert. \hspace{120pt} 2n -1 -2k_\psi < 0 \\
			\end{align*}	
		\end{enumerate}
			\item The proof for $\varphi \metric{\release}{n} \psi$, $\varphi \metric{\since}{n} \psi$ and $\varphi \metric{\trigger}{n} \psi$ follows the same line of reasoning as for $\varphi \metric{\until}{n} \psi$.
\end{itemize}
\end{proofof}

\begin{proofof}{Proposition~\ref{lem:nf2}}
	We proceed by structural induction on $\mu$. We consider only the the metric operators.
	
	\begin{itemize}
		\item If $\mu = \varphi \metric{\until}{n}\psi$ we proceed by cases. If $\numeral{n} = 1$ we conclude that 
		\begin{equation*}
		\trival{k}{\Lab\mu} \stackrel{\eta(\mu)}{=} \trival{k}{\Lab{\psi}}\stackrel{IH}{\text{=}}\trival{k}{\psi} \stackrel{Prop.~\ref{prop:validities}}{=} \trival{k}{\mu}.
		\end{equation*}
		
		\noindent If $\numeral{n} > 1$, let us take $\varphi' = \next \left( \varphi \metric{\until}{n-1} \psi\right)$ to conclude that 
		\begin{align*}
		\trival{k}{\Lab\mu} & \stackrel{\eta(\mu)}{=} \trival{k}{\Lab{\psi} \vee \left(\Lab{\varphi} \wedge  \Lab{\varphi'}\right)} \stackrel{IH}{\text{=}}\trival{k}{\psi \vee \left(\varphi \wedge  \varphi'\right)} \stackrel{Prop.~\ref{prop:validities}}{=} \trival{k}{\mu}.
		\end{align*}
	
		\noindent For the case of $\last$ let us take $\varphi' = \next \left( \varphi \metric{\until}{\last} \psi\right)$ to conclude that
	
			\begin{align*}
				\trival{k}{\Lab\mu} & \stackrel{\eta(\mu)}{=} \trival{k}{\Lab{\psi} \vee \left(\Lab{\varphi} \wedge  \Lab{\varphi'}\right)} \stackrel{IH}{\text{=}}\trival{k}{\psi \vee \left(\varphi \wedge  \varphi'\right)} \stackrel{Prop.~\ref{prop:validity:l}}{=} \trival{k}{\mu}.
			\end{align*}

		\item If $\mu = \varphi \metric{\release}{n}\psi$ we proceed by cases. If $\numeral{n} = 1$ we use the first formula to conclude that 
		\begin{equation*}
			\trival{k}{\Lab\mu} \stackrel{\eta(\mu)}{=} \trival{k}{\Lab{\psi}}\stackrel{IH}{\text{=}}\trival{k}{\psi} \stackrel{}{=} \trival{k}{\mu}
		\end{equation*}
				
		\noindent If $\numeral{n} > 1$, let us take $\varphi' = \wnext \left( \varphi \metric{\release}{n-1} \psi\right)$ to conclude that 
		\begin{align*}
			\trival{k}{\Lab\mu}  \stackrel{\eta(\mu)}{=} \trival{k}{\Lab{\psi} \wedge \left(\Lab{\varphi} \vee  \Lab{\varphi'}\right)} 
			\stackrel{IH}{\text{=}}\trival{k}{\psi \wedge \left(\varphi \vee  \varphi'\right)} 
			\stackrel{Prop.~\ref{prop:validities}}{=} \trival{k}{\mu}.
		\end{align*}
	
		\noindent For the case of $\last$ let us take $\varphi' = \wnext \left( \varphi \metric{\release}{\last} \psi\right)$ to conclude that 
		\begin{align*}
			\trival{k}{\Lab\mu} & \stackrel{\eta(\mu)}{=} \trival{k}{\Lab{\psi} \wedge \left(\Lab{\varphi} \vee  \Lab{\varphi'}\right)}
			 \stackrel{IH}{\text{=}}\trival{k}{\psi \wedge \left(\varphi \vee  \varphi'\right)}
			 \stackrel{Prop.~\ref{prop:validity:l}}{=} \trival{k}{\mu}.
		\end{align*}
		
		\item If $\mu = \varphi \metric{\since}{n}\psi$ we proceed by cases. If $\numeral{n} = 1$ we use the first formula to conclude that 
		\begin{equation*}
			\trival{k}{\Lab\mu} \stackrel{\eta(\mu)}{=} \trival{k}{\Lab{\psi}}\stackrel{IH}{\text{=}}\trival{k}{\psi} \stackrel{}{=} \trival{k}{\mu}
		\end{equation*}
	\noindent If $\numeral{n} > 1$, let us take $\varphi' = \previous \left( \varphi \metric{\since}{n-1} \psi\right)$ to conclude that 
	\begin{align*}
		\trival{k}{\Lab\mu} & \stackrel{\eta(\mu)}{=} \trival{k}{\Lab{\psi} \vee \left(\Lab{\varphi} \wedge  \Lab{\varphi'}\right)}
		 \stackrel{IH}{\text{=}}\trival{k}{\psi \vee \left(\varphi \wedge  \varphi'\right)}
		 \stackrel{Prop.~\ref{prop:validities}}{=} \trival{k}{\mu}.
	\end{align*}

	\noindent For the case of $\last$, let us take $\varphi' = \previous \left( \varphi \metric{\since}{\last} \psi\right)$ to conclude that 
	\begin{align*}
		\trival{k}{\Lab\mu} & \stackrel{\eta(\mu)}{=} \trival{k}{\Lab{\psi} \vee \left(\Lab{\varphi} \wedge  \Lab{\varphi'}\right)}
		 \stackrel{IH}{\text{=}}\trival{k}{\psi \vee \left(\varphi \wedge  \varphi'\right)}
		 \stackrel{Prop.~\ref{prop:validity:l}}{=} \trival{k}{\mu}.
	\end{align*}

		\item If $\mu = \varphi \metric{\trigger}{n}\psi$ we proceed by cases. If $\numeral{n} = 1$ we use the first formula to conclude that 
		\begin{equation*}
			\trival{k}{\Lab\mu} \stackrel{\eta(\mu)}{=} \trival{k}{\Lab{\psi}}\stackrel{IH}{\text{=}}\trival{k}{\psi} \stackrel{}{=} \trival{k}{\mu}
		\end{equation*}
		
		\noindent If $\numeral{n} > 1$, let us take $\varphi' = \wprevious \left( \varphi \metric{\trigger}{n-1} \psi\right)$ to conclude that 
		\begin{align*}
			\trival{k}{\Lab\mu} & \stackrel{\eta(\mu)}{=} \trival{k}{\Lab{\psi} \wedge \left(\Lab{\varphi} \vee  \Lab{\varphi'}\right)}
			 \stackrel{IH}{\text{=}}\trival{k}{\psi \wedge \left(\varphi \vee  \varphi'\right)}
			 \stackrel{Prop.~\ref{prop:validities}}{=} \trival{k}{\mu}.
		\end{align*}
	\noindent For the case of $\last$, let us take $\varphi' = \wprevious \left( \varphi \metric{\trigger}{\last} \psi\right)$ to conclude that 
	\begin{align*}
		\trival{k}{\Lab\mu} & \stackrel{\eta(\mu)}{=} \trival{k}{\Lab{\psi} \wedge \left(\Lab{\varphi} \vee  \Lab{\varphi'}\right)}
		 \stackrel{IH}{\text{=}}\trival{k}{\psi \wedge \left(\varphi \vee  \varphi'\right)}
		 \stackrel{Prop.~\ref{prop:validity:l}}{=} \trival{k}{\mu}.
	\end{align*}	
	\end{itemize}

\end{proofof}

\begin{proofof}{Theorem~\ref{lem:nf1}}
	Take the \MELf{}-trace $\tuple{\H',\T'}$ whose three valued interpretation ${\bm m}'$ satisfies:
	\begin{align*}
	\trivalp{k}{\Lab{\varphi}} \overset{\label{proofNF_equalityExtAlph}}{=} \trival{k}{\varphi}
	\end{align*}
	
	for any formula $\varphi$ over $\mathcal{A}$ and for all $\rangeo{i}{k}{\lambda}$.
	When $\varphi$ is an atom $a \in \mathcal{A}$ then $\trivalp{k}{a} = \trivalp{k}{\Lab{a}} = \trival{k}{a}$, which implies that both valuations coincide for atoms, and so, $\tuple{\H',\T'} |_{\mathcal{A}} = \tuple{\H,\T}$.
	It remains to be shown that
	$\tuple{\H',\T'} \models \upsilon(\varphi)$,
	which is equivalent to
	\begin{align*}
	\tuple{\H',\T'} &\models
	\left\lbrace \Lab{\varphi}\right\rbrace
	\cup \left\lbrace \eta(\mu)
	\mid \mu \in \cl(\varphi)\right\rbrace\\
	\Leftrightarrow \tuple{\H',\T'} &\models
	\left\lbrace \Lab{\varphi} \right\rbrace
	\text{ and }
	\tuple{\H',\T'} \models \left\lbrace \eta(\mu)
	\mid \mu \in \cl(\varphi)\right\rbrace
	\end{align*}
	
	The first satisfaction relation follows directly from the definition of $\tuple{\H',\T'}$ since $\trivalp{0}{\Lab{\varphi}}=2$ iff $\trival{0}{\varphi}=2$ and we had that $\tuple{\H,\T}$ is a model of $\varphi$.
	
	For the second part, we consider the following cases depending on the structure of $\mu$:
	\begin{itemize}
	\item The boolean connectives and temporal formulas $\next \varphi$, $\previous\varphi$, $\wprevious\varphi$ and $\wnext\varphi$ are left to the reader.
	
	\item For the formula $\mu = \varphi \metric{\until}{1}\psi$, note that $\varphi,\psi \in \cl(\mu)$. Let us reason as follows
	
	\begin{align*}
	\trivalp{k}{\Lab{\mu}} & = \trival{k}{\mu} = \trival{k}{\varphi \metric{\until}{1}\psi} \stackrel{Prop.~\ref{prop:validities}}{=} \trival{k}{\psi} = \trival{k}{\Lab{\psi}}.
	\end{align*}
	
	\noindent The cases $\varphi \metric{\release}{1}\psi$, $\varphi \metric{\since}{1}\psi$ and $\varphi \metric{\trigger}{1}\psi$ are proved in a similar way.
	
	\item For the formula $\mu = \varphi \metric{\until}{n}\psi$, with $\numeral{n}>1$, let us take $\varphi' = \next \left( \varphi \metric{\until}{n-1} \psi\right)$. Note that, by definition $\varphi, \psi, \varphi' \in \cl(\mu)$. Having said that, we reason as follows
	
	\begin{align*}
	\trivalp{k}{\Lab{\mu}} & = \trival{k}{\mu} = \trival{k}{\varphi \metric{\until}{n}\psi}\\
	& \stackrel{Prop.~\ref{prop:validities}}{=} \trival{k}{\psi \vee \left(\varphi \wedge \next \left( \varphi \metric{\until}{n-1} \psi\right)\right)}\\
	&= \max\lbrace\trival{k}{\psi},\min(\trival{k}{\varphi},\trival{k}{\next \left( \varphi \metric{\until}{n-1} \psi\right)})\rbrace \\
	&= \max\lbrace\trival{k}{\Lab{\psi}},\min(\trival{k}{\Lab\varphi},\trival{k}{\Lab{\varphi'}})\rbrace \\
	&=  \trival{k}{\Lab{\psi} \vee \left(\Lab\varphi \wedge \Lab{\varphi'} \right)}.\\
	\end{align*}

\item For the formula $\mu = \varphi \metric{\release}{n}\psi$, with $\numeral{n}>1$, let us take $\varphi' = \wnext \left( \varphi \metric{\release}{n-1} \psi\right)$. Note that, by definition $\varphi, \psi, \varphi' \in \cl(\mu)$. Having said that, we reason as follows

\begin{align*}
	\trivalp{k}{\Lab{\mu}} & = \trival{k}{\mu} = \trival{k}{\varphi \metric{\release}{n}\psi}\\
	& \stackrel{Prop.~\ref{prop:validities}}{=} \trival{k}{\psi \wedge \left(\varphi \vee \wnext \left( \varphi \metric{\release}{n-1} \psi\right)\right)}\\
	&= \min\lbrace\trival{k}{\psi},\max(\trival{k}{\varphi},\trival{k}{\wnext \left( \varphi \metric{\release}{n-1} \psi\right)})\rbrace \\
	&= \min\lbrace\trival{k}{\Lab{\psi}},\max(\trival{k}{\Lab\varphi},\trival{k}{\Lab{\varphi'}})\rbrace \\
	&=  \trival{k}{\Lab{\psi} \vee \left(\Lab\varphi \wedge \Lab{\varphi'} \right)}.\\
\end{align*}

	\item For the formula $\mu \varphi \metric{\since}{n}\psi$, with $\numeral{n}>1$, let us take $\varphi' = \previous \left( \varphi \metric{\since}{n-1} \psi\right)$. Note that, by definition $\varphi, \psi, \varphi' \in \cl(\mu)$. Having said that, we reason as follows

	\begin{align*}
	\trivalp{k}{\Lab{\mu}} & = \trival{k}{\mu} = \trival{k}{\varphi \metric{\since}{n}\psi}\\
	& \stackrel{Prop.~\ref{prop:validities}}{=} \trival{k}{\psi \vee \left(\varphi \wedge \previous \left( \varphi \metric{\since}{n-1} \psi\right)\right)}\\
	&= \max\lbrace\trival{k}{\psi},\min(\trival{k}{\varphi},\trival{k}{\previous \left( \varphi \metric{\since}{n-1} \psi\right)})\rbrace \\
	&= \max\lbrace\trival{k}{\Lab{\psi}},\min(\trival{k}{\Lab\varphi},\trival{k}{\Lab{\varphi'}})\rbrace \\
	&=  \trival{k}{\Lab{\psi} \vee \left(\Lab\varphi \wedge \Lab{\varphi'} \right)}.\\
	\end{align*}

\item For the formula $\mu = \varphi \metric{\trigger}{n}\psi$, with $\numeral{n}>1$, let us take $\varphi' = \wprevious \left( \varphi \metric{\trigger}{n-1} \psi\right)$. Note that, by definition $\varphi, \psi, \varphi' \in \cl(\mu)$. Having said that, we reason as follows

\begin{align*}
	\trivalp{k}{\Lab{\mu}} & = \trival{k}{\mu} = \trival{k}{\varphi \metric{\trigger}{n}\psi}\\
	& \stackrel{Prop.~\ref{prop:validities}}{=} \trival{k}{\psi \wedge \left(\varphi \vee \wprevious \left( \varphi \metric{\trigger}{n-1} \psi\right)\right)}\\
	&= \min\lbrace\trival{k}{\psi},\max(\trival{k}{\varphi},\trival{k}{\wprevious \left( \varphi \metric{\trigger}{n-1} \psi\right)})\rbrace \\
	&= \min\lbrace\trival{k}{\Lab{\psi}},\max(\trival{k}{\Lab\varphi},\trival{k}{\Lab{\varphi'}})\rbrace \\
	&=  \trival{k}{\Lab{\psi} \vee \left(\Lab\varphi \wedge \Lab{\varphi'} \right)}.\\
\end{align*}

\item For the formula $\mu = \varphi \metric{\until}{\last}\psi$, let us take $\varphi' = \next \left( \varphi \metric{\until}{\last} \psi\right)$. Note that, by definition $\varphi, \psi, \varphi' \in \cl(\mu)$. Having said that, we reason as follows

\begin{align*}
	\trivalp{k}{\Lab{\mu}} & = \trival{k}{\mu} = \trival{k}{\varphi \metric{\until}{\last}\psi}\\
	& \stackrel{Prop.~\ref{prop:validity:l}}{=} \trival{k}{\psi \vee \left(\varphi \wedge \next \left( \varphi \metric{\until}{\last} \psi\right)\right)}\\
	&= \max\lbrace\trival{k}{\psi},\min(\trival{k}{\varphi},\trival{k}{\next \left( \varphi \metric{\until}{\last} \psi\right)})\rbrace \\
	&= \max\lbrace\trival{k}{\Lab{\psi}},\min(\trival{k}{\Lab\varphi},\trival{k}{\Lab{\varphi'}})\rbrace \\
	&=  \trival{k}{\Lab{\psi} \vee \left(\Lab\varphi \wedge \Lab{\varphi'} \right)}.\\
\end{align*}

\item For the formula $\mu= \varphi \metric{\release}{\last}\psi$, let us take $\varphi' = \wnext \left( \varphi \metric{\release}{\last} \psi\right)$. Note that, by definition $\varphi, \psi, \varphi' \in \cl(\mu)$. Having said that, we reason as follows

\begin{align*}
	\trivalp{k}{\Lab{\mu}} & = \trival{k}{\mu} = \trival{k}{\varphi \metric{\release}{\last}\psi}\\
	& \stackrel{Prop.~\ref{prop:validity:l}}{=} \trival{k}{\psi \wedge \left(\varphi \vee \wnext \left( \varphi \metric{\release}{\last} \psi\right)\right)}\\
	&= \min\lbrace\trival{k}{\psi},\max(\trival{k}{\varphi},\trival{k}{\wnext \left( \varphi \metric{\release}{\last} \psi\right)})\rbrace \\
	&= \min\lbrace\trival{k}{\Lab{\psi}},\max(\trival{k}{\Lab\varphi},\trival{k}{\Lab{\varphi'}})\rbrace \\
	&=  \trival{k}{\Lab{\psi} \vee \left(\Lab\varphi \wedge \Lab{\varphi'} \right)}.\\
\end{align*}

\item For the formula $\mu = \varphi \metric{\since}{\last}\psi$, let us take $\varphi' = \previous \left( \varphi \metric{\since}{\last} \psi\right)$. Note that, by definition $\varphi, \psi, \varphi' \in \cl(\mu)$. Having said that, we reason as follows

\begin{align*}
	\trivalp{k}{\Lab{\mu}} & = \trival{k}{\mu} = \trival{k}{\varphi \metric{\since}{\last}\psi}\\
	& \stackrel{Prop.~\ref{prop:validity:l}}{=} \trival{k}{\psi \vee \left(\varphi \wedge \previous \left( \varphi \metric{\since}{\last} \psi\right)\right)}\\
	&= \max\lbrace\trival{k}{\psi},\min(\trival{k}{\varphi},\trival{k}{\previous \left( \varphi \metric{\since}{\last} \psi\right)})\rbrace \\
	&= \max\lbrace\trival{k}{\Lab{\psi}},\min(\trival{k}{\Lab\varphi},\trival{k}{\Lab{\varphi'}})\rbrace \\
	&=  \trival{k}{\Lab{\psi} \vee \left(\Lab\varphi \wedge \Lab{\varphi'} \right)}.\\
\end{align*}

\item For the formula $\mu = \varphi \metric{\trigger}{\last}\psi$, let us take $\varphi' = \wprevious \left( \varphi \metric{\trigger}{\last} \psi\right)$. Note that, by definition $\varphi, \psi, \varphi' \in \cl(\mu)$. Having said that, we reason as follows

\begin{align*}
	\trivalp{k}{\Lab{\mu}} & = \trival{k}{\mu} = \trival{k}{\varphi \metric{\trigger}{\last}\psi}\\
	& \stackrel{Prop.~\ref{prop:validity:l}}{=} \trival{k}{\psi \wedge \left(\varphi \vee \wprevious \left( \varphi \metric{\trigger}{\last} \psi\right)\right)}\\
	&= \min\lbrace\trival{k}{\psi},\max(\trival{k}{\varphi},\trival{k}{\wprevious \left( \varphi \metric{\trigger}{\last} \psi\right)})\rbrace \\
	&= \min\lbrace\trival{k}{\Lab{\psi}},\max(\trival{k}{\Lab\varphi},\trival{k}{\Lab{\varphi'}})\rbrace \\
	&=  \trival{k}{\Lab{\psi} \vee \left(\Lab\varphi \wedge \Lab{\varphi'} \right)}.\\
\end{align*}

	\end{itemize}
\end{proofof}


\begin{thebibliography}{}

\bibitem[\protect\citeauthoryear{Aguado, Cabalar, Di{\'{e}}guez, P{\'{e}}rez,
  and Vidal}{Aguado et~al\mbox{.}}{2013}]{agcadipevi13a}
{\sc Aguado, F.}, {\sc Cabalar, P.}, {\sc Di{\'{e}}guez, M.}, {\sc P{\'{e}}rez,
  G.}, {\sc and} {\sc Vidal, C.} 2013.
\newblock Temporal equilibrium logic: a survey.
\newblock {\em Journal of Applied Non-Classical Logics\/}~{\em 23,\/}~1-2,
  2--24.

\bibitem[\protect\citeauthoryear{Allen}{Allen}{1983}]{allen83a}
{\sc Allen, J.} 1983.
\newblock Maintaining knowledge about temporal intervals.
\newblock {\em Communications of the {ACM}\/}~{\em 26,\/}~11, 832--843.

\bibitem[\protect\citeauthoryear{Alur and Henzinger}{Alur and
  Henzinger}{1992}]{aluhen92a}
{\sc Alur, R.} {\sc and} {\sc Henzinger, T.} 1992.
\newblock Logics and models of real time: A survey.
\newblock In {\em Real-Time: Theory in Practice}. Springer, 74--106.

\bibitem[\protect\citeauthoryear{Balduccini, Lierler, and Woltran}{Balduccini
  et~al\mbox{.}}{2019}]{lpnmr19}
{\sc Balduccini, M.}, {\sc Lierler, Y.}, {\sc and} {\sc Woltran, S.}, Eds.
  2019.
\newblock {\em Proceedings of the Fifteenth International Conference on Logic
  Programming and Nonmonotonic Reasoning (LPNMR'19)}. Springer.

\bibitem[\protect\citeauthoryear{Beck, Dao-Tran, and Eiter}{Beck
  et~al\mbox{.}}{2016}]{bedaei16a}
{\sc Beck, H.}, {\sc Dao-Tran, M.}, {\sc and} {\sc Eiter, T.} 2016.
\newblock Equivalent stream reasoning programs.
\newblock In {\em Proceedings of the Twenty-fifth International Joint
  Conference on Artificial Intelligence (IJCAI'16)}, {R.~Kambhampati}, Ed.
  IJCAI/AAAI Press, 929--935.

\bibitem[\protect\citeauthoryear{Beck, Dao{-}Tran, Eiter, and Fink}{Beck
  et~al\mbox{.}}{2015}]{bedaeifi15a}
{\sc Beck, H.}, {\sc Dao{-}Tran, M.}, {\sc Eiter, T.}, {\sc and} {\sc Fink, M.}
  2015.
\newblock {LARS:} {A} logic-based framework for analyzing reasoning over
  streams.
\newblock In {\em Proceedings of the Twenty-Ninth National Conference on
  Artificial Intelligence (AAAI'15)}, {B.~Bonet} {and} {S.~Koenig}, Eds. AAAI
  Press, 1431--1438.

\bibitem[\protect\citeauthoryear{Bosser, Cabalar, Di\'eguez, and Schaub}{Bosser
  et~al\mbox{.}}{2018}]{bocadisc18a}
{\sc Bosser, A.}, {\sc Cabalar, P.}, {\sc Di\'eguez, M.}, {\sc and} {\sc
  Schaub, T.} 2018.
\newblock Introducing temporal stable models for linear dynamic logic.
\newblock In {\em Proceedings of the Sixteenth International Conference on
  Principles of Knowledge Representation and Reasoning (KR'18)},
  {M.~Thielscher}, {F.~Toni}, {and} {F.~Wolter}, Eds. AAAI Press, 12--21.

\bibitem[\protect\citeauthoryear{Brandt, Kalayci, Ryzhikov, Xiao, and
  Zakharyaschev}{Brandt et~al\mbox{.}}{2018}]{brkaryxiza18a}
{\sc Brandt, S.}, {\sc Kalayci, E.}, {\sc Ryzhikov, V.}, {\sc Xiao, G.}, {\sc
  and} {\sc Zakharyaschev, M.} 2018.
\newblock Querying log data with metric temporal logic.
\newblock {\em Journal of Artificial Intelligence Research\/}~{\em 62},
  829--877.

\bibitem[\protect\citeauthoryear{Brzoska}{Brzoska}{1995}]{brzoska93a}
{\sc Brzoska, C.} 1995.
\newblock Temporal logic programming with metric and past operators.
\newblock In {\em Proceedings of the Workshop on Executable Modal and Temporal
  Logics}, {M.~Fisher} {and} {R.~Owens}, Eds. Springer, 21--39.

\bibitem[\protect\citeauthoryear{Cabalar}{Cabalar}{2010}]{cabalar10a}
{\sc Cabalar, P.} 2010.
\newblock A normal form for linear temporal equilibrium logic.
\newblock In {\em Proceedings of the Twelfth European Conference on Logics in
  Artificial Intelligence (JELIA'10)}, {T.~Janhunen} {and} {I.~Niemel{\"a}},
  Eds. Springer,
  64--76.

\bibitem[\protect\citeauthoryear{Cabalar, Di\'eguez, Laferriere, and
  Schaub}{Cabalar et~al\mbox{.}}{2020}]{cadilasc20a}
{\sc Cabalar, P.}, {\sc Di\'eguez, M.}, {\sc Laferriere, F.}, {\sc and} {\sc
  Schaub, T.} 2020.
\newblock Implementing dynamic answer set programming over finite traces.
\newblock In {\em Proceedings of the Twenty-fourth European Conference on
  Artificial Intelligence (ECAI'20)}, {G.~{De Giacomo}}, Ed. IOS Press.
\newblock To appear.

\bibitem[\protect\citeauthoryear{Cabalar, Diéguez, and Schaub}{Cabalar
  et~al\mbox{.}}{2019}]{cadisc19a}
{\sc Cabalar, P.}, {\sc Diéguez, M.}, {\sc and} {\sc Schaub, T.} 2019.
\newblock Towards dynamic answer set programming over finite traces.
\newblock See \citeN{lpnmr19}, 148--162.

\bibitem[\protect\citeauthoryear{Cabalar, Kaminski, Morkisch, and
  Schaub}{Cabalar et~al\mbox{.}}{2019}]{cakamosc19a}
{\sc Cabalar, P.}, {\sc Kaminski, R.}, {\sc Morkisch, P.}, {\sc and} {\sc
  Schaub, T.} 2019.
\newblock telingo = {ASP} + time.
\newblock See \citeN{lpnmr19}, 256--269.

\bibitem[\protect\citeauthoryear{Cabalar, Kaminski, Schaub, and
  Schuhmann}{Cabalar et~al\mbox{.}}{2018}]{cakascsc18a}
{\sc Cabalar, P.}, {\sc Kaminski, R.}, {\sc Schaub, T.}, {\sc and} {\sc
  Schuhmann, A.} 2018.
\newblock Temporal answer set programming on finite traces.
\newblock {\em Theory and Practice of Logic Programming\/}~{\em 18,\/}~3-4,
  406--420.

\bibitem[\protect\citeauthoryear{Fischer and Ladner}{Fischer and
  Ladner}{1979}]{fislad79a}
{\sc Fischer, M.} {\sc and} {\sc Ladner, R.} 1979.
\newblock Propositional dynamic logic of regular programs.
\newblock {\em Journal of Computer and System Sciences\/}~{\em 18,\/}~2,
  194--211.

\bibitem[\protect\citeauthoryear{Gebser, Kaminski, Kaufmann, Ostrowski, Schaub,
  and Wanko}{Gebser et~al\mbox{.}}{2016}]{gekakaosscwa16a}
{\sc Gebser, M.}, {\sc Kaminski, R.}, {\sc Kaufmann, B.}, {\sc Ostrowski, M.},
  {\sc Schaub, T.}, {\sc and} {\sc Wanko, P.} 2016.
\newblock Theory solving made easy with clingo~5.
\newblock In {\em Technical Communications of the Thirty-second International
  Conference on Logic Programming (ICLP'16)}, {M.~Carro} {and} {A.~King}, Eds.
  OpenAccess Series in Informatics (OASIcs), 2:1--2:15.

\bibitem[\protect\citeauthoryear{Gödel}{Gödel}{1932}]{goedel32a}
{\sc Gödel, K.} 1932.
\newblock Zum intuitionistischen {A}ussagenkalkül.
\newblock {\em Anzeiger der Akademie der Wissenschaften in Wien\/}, 65--66.

\bibitem[\protect\citeauthoryear{Harel, Tiuryn, and Kozen}{Harel
  et~al\mbox{.}}{2000}]{hatiko00a}
{\sc Harel, D.}, {\sc Tiuryn, J.}, {\sc and} {\sc Kozen, D.} 2000.
\newblock {\em Dynamic Logic}.
\newblock MIT Press.

\bibitem[\protect\citeauthoryear{Heyting}{Heyting}{1930}]{heyting30a}
{\sc Heyting, A.} 1930.
\newblock Die formalen {R}egeln der intuitionistischen {L}ogik.
\newblock In {\em Sitzungsberichte der Preussischen Akademie der
  Wissenschaften}. Deutsche Akademie der Wissenschaften zu Berlin, 42--56.
\newblock Reprint in Logik-Texte: Kommentierte Auswahl zur Geschichte der
  Modernen Logik, Akademie-Verlag, 1986.

\bibitem[\protect\citeauthoryear{Kowalski and Sergot}{Kowalski and
  Sergot}{1986}]{kowser86a}
{\sc Kowalski, R.} {\sc and} {\sc Sergot, M.} 1986.
\newblock A logic-based calculus of events.
\newblock {\em New Generation Computing\/}~{\em 4,\/}~1, 67--95.

\bibitem[\protect\citeauthoryear{Lifschitz}{Lifschitz}{1999}]{lifschitz99b}
{\sc Lifschitz, V.} 1999.
\newblock Answer set planning.
\newblock In {\em Proceedings of the International Conference on Logic
  Programming (ICLP'99)}, {D.~{de Schreye}}, Ed. MIT Press, 23--37.

\bibitem[\protect\citeauthoryear{Ouaknine and Worrell}{Ouaknine and
  Worrell}{2005}]{ouawor05a}
{\sc Ouaknine, J.} {\sc and} {\sc Worrell, J.} 2005.
\newblock On the decidability of metric temporal logic.
\newblock In {\em Proceedings of the Twentieth Annual Symposium on Logic in
  Computer Science (LICS'10)}. {IEEE} Computer Society Press, 188--197.

\bibitem[\protect\citeauthoryear{Pearce}{Pearce}{1997}]{pearce96a}
{\sc Pearce, D.} 1997.
\newblock A new logical characterisation of stable models and answer sets.
\newblock In {\em Proceedings of the Sixth International Workshop on
  Non-Monotonic Extensions of Logic Programming (NMELP'96)}, {J.~Dix},
  {L.~Pereira}, {and} {T.~Przymusinski}, Eds. Springer, 57--70.

\bibitem[\protect\citeauthoryear{Pnueli}{Pnueli}{1977}]{pnueli77a}
{\sc Pnueli, A.} 1977.
\newblock The temporal logic of programs.
\newblock In {\em Proceedings of the Eight-teenth Symposium on Foundations of
  Computer Science (FOCS'77)}. {IEEE} Computer Society Press, 46--57.

\bibitem[\protect\citeauthoryear{Son, Baral, and Tuan}{Son
  et~al\mbox{.}}{2004}]{sobatu04a}
{\sc Son, T.}, {\sc Baral, C.}, {\sc and} {\sc Tuan, L.} 2004.
\newblock Adding time and intervals to procedural and hierarchical control
  specifications.
\newblock In {\em Proceedings of the Nineteenth National Conference on
  Artificial Intelligence (AAAI'04)}, {D.~McGuinness} {and} {G.~Ferguson}, Eds.
  AAAI Press, 92--97.

\bibitem[\protect\citeauthoryear{Walega, Kaminski, and {Cuenca Grau}}{Walega
  et~al\mbox{.}}{2019}]{wakagr19a}
{\sc Walega, P.}, {\sc Kaminski, M.}, {\sc and} {\sc {Cuenca Grau}, B.} 2019.
\newblock Reasoning over streaming data in metric temporal datalog.
\newblock In {\em Proceedings of the Thirty-third National Conference on
  Artificial Intelligence (AAAI'19)}, {P.~{Van Hentenryck}} {and} {Z.~Zhou},
  Eds. AAAI Press, 3092--3099.

\end{thebibliography}
\end{document}